%% file: main.tex
\documentclass{article}
\usepackage{arxiv}
\usepackage{times}
\usepackage{natbib}
\setcitestyle{authoryear,round,citesep={;},aysep={,},yysep={;}}
\renewcommand{\headeright}{Preprint}
\renewcommand{\undertitle}{Preprint}
\date{}
\input{math_commands.tex}

\input{smac3_numbers.tex}

\newcommand{\UnifiedCFDObserved}{6{,}629}
\newcommand{\CampaignMaxLd}{34.0}

\newcommand{\AgentLowBestAtC}{18.23}

\newcommand{\AgentNominalBestAtC}{33.04}

\newcommand{\AgentHighBestAtC}{30.24}

\newcommand{\OpenDeliveredAtFour}{29}

\newcommand{\PlanningDeliveredAtFour}{28}

\newcommand{\ReflectionDeliveredAtFour}{30}

\newcommand{\AvgCoreHoursPerEvalOverall}{30.4}

\newcommand{\CoreHoursOverall}{200{,}019}
\newcommand{\TimedEvaluationsOverall}{6{,}576}
\usepackage{hyperref}
\hypersetup{hidelinks}
\usepackage{url}
\usepackage{graphicx}
\usepackage{tikz}
\usetikzlibrary{plotmarks}
\definecolor{wbNavy}{HTML}{004E85}
\definecolor{wbNavyLight}{HTML}{7EA4C2}
\definecolor{wbMaroon}{HTML}{8F1F22}
\definecolor{wbTeal}{HTML}{107A83}
\definecolor{wbOchre}{HTML}{DF8932}
\definecolor{wbIndigo}{HTML}{56669E}
\definecolor{wbTerracotta}{HTML}{C36D56}
\definecolor{wbAqua}{HTML}{50B9AE}
\tikzset{
  wbfig/.style={x=0.75pt,y=0.84pt,font=\footnotesize,line width=0.8pt},
  wblabel/.style={font=\small},
  wbgrid/.style={black!12,line width=0.35pt},
  wbframe/.style={black!60,line width=0.5pt},
  wbmarks/.style={only marks,mark size=1.6pt,
    mark options={fill=white,line width=0.8pt,solid}},
  cma no feedback/.style={wbNavyLight,dashed},
  cma no feedback marks/.style={wbNavyLight,wbmarks,mark=square*},
  cma selection/.style={wbNavy,solid},
  cma selection marks/.style={wbNavy,wbmarks,mark=*},
  cma feedback/.style={wbMaroon,solid},
  cma feedback marks/.style={wbMaroon,wbmarks,mark=diamond*},
  de feedback/.style={wbTeal,dashdotted},
  de feedback marks/.style={wbTeal,wbmarks,mark=triangle*},
  smac3/.style={wbOchre,densely dotted},
  smac3 marks/.style={wbOchre,wbmarks,mark=+},
  open research/.style={wbIndigo,solid},
  open research marks/.style={wbIndigo,wbmarks,mark=*},
  planning/.style={wbTerracotta,dashed},
  planning marks/.style={wbTerracotta,wbmarks,mark=square*},
  reflection/.style={wbAqua,densely dotted},
  reflection marks/.style={wbAqua,wbmarks,mark=triangle*},
}
\newcommand{\wbkey}[2]{\tikz[wbfig,baseline=-0.6ex]{%
  \draw[#1] (0,0) -- (18,0);\draw[#1 marks] plot coordinates {(9,0)};}~#2}
\usepackage{booktabs}
\usepackage{placeins}
\usepackage{microtype}
\usepackage{amsmath}
\usepackage{caption}
\graphicspath{{./}}

\title{WingBench: Benchmarking How AI Agents Search for Better Designs}

\author{%
  Yiming Yang\textsuperscript{1,}\thanks{Equal contribution.}\quad
  David Xia\textsuperscript{2,}\footnotemark[1]\quad
  Jisen Li\textsuperscript{2}\quad
  Yuqing Jian\textsuperscript{2}\quad
  Yunjia Yang\textsuperscript{3}\quad
  Koray Aydin\textsuperscript{1}\quad
  Zhaoran Wang\textsuperscript{1}\\[4pt]
  \textsuperscript{1}Northwestern University\quad
  \textsuperscript{2}University of Illinois at Urbana-Champaign\quad
  \textsuperscript{3}Technical University of Munich%
}

\usepackage{longtable}
\usepackage{float}
\usepackage{listings}
\lstdefinestyle{prompt}{basicstyle=\ttfamily\scriptsize,breaklines=true,breakatwhitespace=false,columns=fullflexible,keepspaces=true,frame=single,framerule=0.3pt,xleftmargin=4pt,xrightmargin=4pt,showstringspaces=false}
\begin{document}

\maketitle

\begin{abstract}
Design search often uses cheap learned predictions to choose which candidates
deserve expensive tests, but these tests can fail, especially on the unusual
designs that optimization favors. We introduce WingBench, a multifidelity
benchmark in which optimizers and AI agents design transonic wings with
computational fluid dynamics (CFD) inside the search. They can query a learned
aerodynamic surrogate and spend a fixed budget of sequential CFD evaluations,
each returning a score or a failure type. WingBench grades the design each
method recommends, reporting separately whether it receives a valid score and
how good that score is. Using WingBench, we contribute a study of five baseline
optimizers and a comparison of three AI agent methods, which
produced \UnifiedCFDObserved{} CFD evaluation records. Feedback helps
baseline optimizers deliver valid designs when it changes their later proposals,
but the differential-evolution variant does so with lower aerodynamic quality.
An open-ended research agent finds designs of similar quality to the strongest
baselines over 32 CFD queries, with different prior information and unmatched
total computation. In early search, our reflection and planning implementations
do not consistently improve on it. We release the data, code,
agent prompts, and analysis scripts for reproducibility.
\end{abstract}

\section{Introduction}

Engineering design often requires searching a large space with few expensive
evaluations. Learned models make this practical by screening candidates, but
their predictions can become unreliable on the unusual designs favored by
optimization. A search method must choose both what to test next and which
design the available evidence supports recommending.

AI agents combine prior knowledge, tools and experimental feedback in
computational research \citep{lu2026aiscientist} and laboratory hypothesis
generation \citep{swanson2025virtuallab,gottweis2026coscientist}. Better design
choices do not automatically follow: classical optimizers have outperformed
agents in controlled comparisons \citep{ferreira2026autoresearch}, and agents
can fail to respond to their own results \citep{gupta2025llmbo}. We ask whether
explicit reflection \citep{shinn2023reflexion} or replanning
\citep{liu2024rafa} improves on open-ended research when expensive tests can fail.

We introduce WingBench, a multifidelity benchmark for studying how optimization methods and agents use cheap predictions and limited, failure-prone tests to find designs. Transonic wing design provides a concrete setting. A
candidate wing is judged by its lift-to-drag ratio $L/D$ under computational
fluid dynamics (CFD), a
high-fidelity numerical model. WingBench's recorded mesh and flow timings
average \AvgCoreHoursPerEvalOverall{} processor-core hours per timed record
(Appendix~\ref{app:resources}). AeroTransformer
\citep{yang2026aerotransformer}, a learned aerodynamic model, estimates the same
quantities in about a second. However, AeroTransformer's predictions can still be optimistic for wings
that fail to mesh or converge under CFD. Optimizers must learn both where
$L/D$ is high and where the high-fidelity process fails.

Aerodynamic benchmarks run CFD inside optimization from a given
starting design, with gradients \citep{lyu2015crm,felten2025engibench} or without
them \citep{plaban2026derivativefree}. Others, such as
ShapeBench \citep{fazliani2026shapebench}, search with learned surrogates and
use CFD to verify final designs. WingBench focuses on recording unsuccessful
evaluations during search and separately grading the final recommendation.
An episode is one optimization run, with up to
4{,}096 predictions from the surrogate and 32 sequential CFD evaluations per
episode. At checkpoints, the method recommends one wing for grading without
receiving that grade as feedback. We report the valid-recommendation rate
(\emph{delivery}) and mean $L/D$ among valid recommendations
(\emph{conditional quality}), separating reliability from aerodynamic performance.

We use WingBench for two complementary studies. The first study establishes baseline
performance with five optimizers. These include three CMA-ES variants that
isolate different uses of CFD feedback, a feedback-guided differential-evolution
method, and a Bayesian optimization baseline implemented with SMAC3
\citep{lindauer2022smac3}. The second study tests whether explicit reflection or planning
improves on open-ended agentic research. We compared these three agent methods after four CFD queries and extended open-ended research to
32 queries to contextualize its performance against the baseline optimizers.
These cross-family results describe the tested configurations; prior information
and total computation are not matched. At four queries, the three methods deliver similarly, while open-ended research
leads in aerodynamic quality at nominal load and no method leads elsewhere. Neither
reflection nor planning consistently improves on open-ended research across
both outcomes in the tested implementations.

The paper makes four contributions.
\begin{enumerate}
\item \textbf{A multifidelity benchmark with CFD in the search.} WingBench combines a
cheap learned surrogate with sequential CFD evaluations that can fail with a
recorded failure type. It grades each
recommendation privately and reports delivery and conditional quality
separately.
\item \textbf{A baseline study.} Five baseline optimizers show that feedback
raises delivery when it changes later proposals, and that the proposal
generator still sets the quality of what is delivered.
\item \textbf{An optimization-agent study.} Three agent optimizers test whether
explicit reflection or planning improves on open-ended research.
\item \textbf{Released data and code.} The release holds all
\UnifiedCFDObserved{} evaluation records with failure types, recommendation
scores, benchmark and baseline code, prompts, and analysis scripts.
\end{enumerate}

\section{WingBench}
\label{sec:benchmark}

\subsection{Design space and objective}

WingBench uses the wing parameterization from SuperWing
\citep{yang2026superwing}. Each wing is
described by 38 input variables, including 20 class-shape-transformation
coefficients that control the upper and lower airfoil profiles
\citep{kulfan2008universal} and
18 variables for planform and spanwise shape. We standardize each coordinate
with the released data and use the box $[-3.5,3.5]^{38}$ as our design space. Ninety-three percent
of dataset designs lie inside this box, while the remaining volume still
permits substantial extrapolation. This is compact compared with the 153
degrees of freedom that \citet{yang2026aerotransformer} use for the Common
Research Model wing \citep{vassberg2008crm}.
The objective is $L/D$ at a target lift coefficient $C_L$. Because the required
lift is the operating constraint, angle of attack is adjusted until computed
lift is within 0.02 of the target. We use $C_L\in\{0.15,0.41,0.66\}$, the rounded
lower quartile, median, and upper quartile of successful SuperWing conditions.
We call these low, nominal, and high load. Each wing is evaluated at Mach
0.7853, 0.8195, and 0.8530, the lower quartile, median, and upper quartile of
Mach number among successful SuperWing conditions, with Reynolds number
$2\times10^7$. The score is the
geometric mean of the three $L/D$ values. A design delivers only if all three
conditions return valid results, $L/D\in(0,40]$, and angle of attack lies in
$[-10^\circ,25^\circ]$. The upper limit of 40 leaves headroom above
both the largest successful single-condition $L/D$ in the dataset,
29.4, and the largest three-condition score observed in our studies,
\CampaignMaxLd{}.

\subsection{High-fidelity evaluator and surrogate}

\begin{figure}[ht]
\centering
\includegraphics{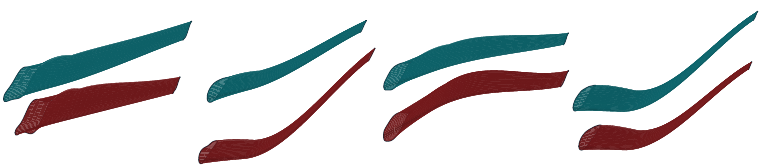}
\caption{Teal wings (top) returned CFD scores; red wings (bottom) failed the
mesh-quality gate under the configured pipeline. These are surface views,
not volume meshes or evidence of physical infeasibility.}
\label{fig:wing-mesh-outcomes}
\end{figure}

We use ATsurf-L, the large surface-flow variant of AeroTransformer \citep{yang2026aerotransformer}.
It predicts surface quantities and integrates them to estimate lift and drag.
In our three-worker CPU deployment, one surrogate evaluation scores one wing in
about a second, including all three Mach conditions and the angle-of-attack
adjustment needed to reach the required lift.
The high-fidelity evaluator checks candidate wings using ADflow
\citep{mader2020adflow}, the SuperWing mesh procedure, and common solver settings.
The released reference setup assigns 16 cores and 64 GB to each CFD job on an
AWS m7a.4xlarge. Historical runs also used cluster workers; identical hardware
across the study is not established.
Meshing has a 30-minute limit and each of three flow conditions a two-hour
limit, both in wall-clock time.
A high-fidelity evaluation can nevertheless fail to produce a usable score.
The geometry may be rejected as unmeshable, meshing may stop without a valid
mesh, or the flow solver may stop before satisfying the convergence and
lift requirements. We call these outcomes nondeliveries and record their types
rather than assign them invented objective values. Across both studies, 4{,}009
of the 6{,}629 requests delivered no score: 557 mesh-gate rejections,
2{,}173 mesh-process stops, 1{,}251 flow timeouts/process errors and 28 other
outcomes. These categories count requests, including reused results.
Timing attached to \TimedEvaluationsOverall{} records sums to
\CoreHoursOverall{} core-hours, including reused outcomes; this is not
unique compute consumed (Appendix~\ref{app:resources}). The failure types give an
optimizer useful evidence about the proposed geometry and where the evaluation
process encounters difficulties. Figure~\ref{fig:wing-mesh-outcomes} shows
recorded wings that returned scores or failed the mesh-quality gate. These
surface views do not establish meshability under other meshing procedures.

Additional checks separate mesh and flow outcomes
(Appendix~\ref{app:evaluator-checks}). Among 3{,}845 audited requests with
accepted meshes, condition failure rates are 19.3--23.3\% (20.8\% pooled);
requiring all three to succeed gives 31.9\% evaluation failure. These include
timeouts and execution errors. Of 1{,}000 successful source geometries,
999 mesh with benchmark spacing and 1{,}000 with source spacing. We also reran
all 1{,}913 distinct failed geometries recoverable from the benchmark records
with the original, unmodified SuperWing/AeroTransformer meshing script;
1{,}912 remained unsuccessful under matched inputs and execution settings.
One additional mesh-failure request lacks an archived geometry and could not
be rerun. These checks support meshing consistency under the tested
configuration, not independent validation of aerodynamic accuracy or
robustness throughout the search space. Failures describe the configured
pipeline and do not establish physical infeasibility.

The surrogate does not solve this problem by itself. It was trained on
converged data and frequently predicts strong scores for regions in which the CFD pipeline is
unreliable. Furthermore, standard failure
classifiers trained on source records reach AUC 0.65--0.71 in cross-validation
but only 0.41--0.75 on optimizer proposals, depending on task and model.
Appendix~\ref{app:classifier} gives details.

\subsection{Episodes, budgets, and private recommendations}

An episode is an independent search, and information is not shared across episodes. An optimizer may spend at most 4{,}096 cheap evaluations and 32
sequential online CFD evaluations. It submits one design, observes the accepted
result, and then chooses again.
After selected CFD query counts, the optimizer nominates one design. Its
CFD performance is assessed outside the search, and the assessment provides no
additional feedback to the optimizer. The benchmark supports checkpoints
$q\in\{0,1,2,4,8,16,32\}$.
We report three quantities. Recommendation delivery is the fraction of checkpoint
recommendations with a valid CFD score. Conditional quality is the mean benchmark
$L/D$ among those valid recommendations. Online delivery is the fraction of requested
experiments that return a score. Appendix~\ref{app:formal} states the problem
in symbols. Appendix~\ref{app:rules} explains how to add a method through
\texttt{propose(ctx)}, \texttt{observe(result)} and \texttt{recommend()},
check integration with a worked example and recorded outcomes, and preserve
comparability through separately versioned evaluator changes.

\section{Related work}

Choosing which candidates to evaluate matters when high-fidelity simulations
are costly. Adjoint methods make gradient-based shape
optimization practical \citep{martins2022aerodynamic}, and benchmarks such as
the Common Research Model wing case \citep{lyu2015crm} and the EngiBench
airfoil problem \citep{felten2025engibench} run CFD inside that optimization.
They start from a given design and report a failed simulation as an error. WingBench
instead treats CFD as a black box with a fixed budget of 32 evaluations that
can fail. Surrogate-assisted search explores more designs with cheap models
\citep{forrester2009recent,jin2011surrogate}, and multifidelity Bayesian
optimization adds lower-cost approximations
\citep{kandasamy2017multifidelity,takeno2020multifidelity}, even when
simulations can fail \citep{he2021variable}. Work on hidden constraints, which a
simulation reveals only by failing \citep{ledigabel2024taxonomy}, learns where
the objective can be computed
\citep{lee2011hidden,gelbart2014unknown,sacher2018classification,bachoc2020failures,chakrabarty2021failure},
including in CFD and aircraft design \citep{tran2019pbo,he2020dual,tfaily2024hidden}, and
benchmark problems with such failures exist outside CFD
\citep{stenger2022benchmark,bussemaker2024hidden}.
ShapeBench \citep{fazliani2026shapebench} is the closest work to ours. It uses
the same SuperWing parameterization and also runs LLM-driven optimizers, one
with a persistent reflection memory, but they search with surrogates and CFD,
if available, only verifies the final designs. WingBench instead integrates CFD evaluations
directly into the search, returning each evaluation's score or failure type to
the optimizer. Furthermore, it privately grades the design each optimizer
explicitly recommends, not just the best design it found during search. Agent
benchmarks make similar distinctions, reporting the best submission apart from
the best attempt \citep{nathani2025mlgym,chan2025mlebench,huang2024mlagentbench} or valid execution apart from success
\citep{chen2025scienceagentbench,somasekharan2026cfdllmbench}, but checking an
attempt there is cheap and can be repeated within a run. A WingBench grade
costs hours of CFD and can fail, so each recommendation is graded once,
privately, at a fixed budget. Together, these
let WingBench test whether an optimizer turns limited, failure-prone CFD
feedback into a design worth recommending.

The optimizer can also be an AI agent. Systems that conduct computational
research \citep{lu2026aiscientist,yamada2025aiscientistv2,schmidgall2025agentlab},
run laboratory chemistry \citep{boiko2023coscientist,bran2024chemcrow}, discover
algorithms \citep{romeraparedes2024funsearch,novikov2025alphaevolve}, or propose
scientific designs \citep{swanson2025virtuallab} show how language models can
participate in research beyond predicting an outcome. Language-model
agents can also run CFD, using solver errors to repair their simulation setups
\citep{yue2026foamagent}, and language-model optimizers can estimate the
chance that an evaluation fails \citep{meindl2026semanticopt}. For design
search, this raises the question of how an agent should organize its reasoning. Reflexion addresses this by turning feedback into
written reflections that are retained to guide subsequent attempts
\citep{shinn2023reflexion}. Reason for Future, Act for Now (RAFA) emphasizes planning ahead from the
interaction history, taking the first planned action, and revising the plan
after observing its outcome \citep{liu2024rafa}. Evidence on when such
structure helps is mixed. Written self-critique improves results mainly when
reliable external feedback is available
\citep{huang2024selfcorrect,kamoi2024selfcorrection}, and searching over
candidate plans pays off only when those plans can be judged accurately
\citep{chen2024treesearch}.
Most of this evidence comes from games, question answering, mathematics, and
coding, where feedback is cheap and an attempt is easy to repeat. It is less
clear whether such structure helps when each piece of feedback is a costly
simulation that may fail. WingBench creates such a setting by combining a
learned aerodynamic surrogate with sequential CFD evaluations, giving
multifidelity feedback on an engineering task.
We test whether reflection or planning improves on open-ended research.

\section{Methods}
\label{sec:methods}
We study two ways of organizing sequential design search. The baseline
optimizers use numerical update rules, while the agents can choose their own
analyses and code.
\subsection{Baseline optimizers}
\label{sec:baseline-methods}

The five baseline methods all receive 32 online CFD evaluations. Four also use
4{,}096 surrogate evaluations. Table~\ref{tab:baseline-methods} summarizes the
controlled differences. At each checkpoint, every method recommends its best
delivered online design once it has one. Before that, it recommends an untested
design chosen by its own selection rule. The four surrogate-assisted methods
also start every episode from the same 62 CFD results recorded during benchmark
development. Those that learn from feedback use them as prior data, none of the
62 is tested again, and the agents did not receive them.

\begin{table}[htbp]
\caption{Baseline optimizers.}
\label{tab:baseline-methods}
\centering
\scriptsize
\setlength{\tabcolsep}{3.5pt}
\begin{tabular*}{\textwidth}{@{\extracolsep{\fill}}lccccl}
\toprule
method & \shortstack{surrogate\\calls} & \shortstack{CFD\\calls} & \shortstack{feedback\\picks next} & \shortstack{changes\\proposals} & generator \\
\midrule
CMA-ES, no feedback & 4{,}096 & 32 & no & no & CMA-ES \\
CMA-ES, selection only & 4{,}096 & 32 & yes & no & CMA-ES \\
Feedback-guided CMA-ES & 4{,}096 & 32 & yes & yes & CMA-ES \\
Feedback-guided DE & 4{,}096 & 32 & yes & yes & DE \\
SMAC3 & 0 & 32 & yes & yes & GP BO \\
\bottomrule
\end{tabular*}
\end{table}

The first three use CMA-ES \citep{hansen2001cmaes,hansen2016tutorial} as their
proposal generator. They share a 64-design archive policy. The \emph{no-feedback} method completes the surrogate search and evaluates the 32
best distinct archive designs in descending surrogate-score order.
The \emph{selection-only} method fits simple
delivery and value-correction models to choose which archived design to evaluate
next, but cannot generate a new design. The \emph{feedback-guided CMA-ES} method alternates
eight surrogate generations with one CFD evaluation. The resulting feedback
updates beliefs that guide both selection and the future proposal distribution.
The \emph{feedback-guided differential evolution} method uses differential
evolution \citep{storn1997differential}. We used DE/rand/1/bin with mutation
factor 0.8 and crossover probability 0.9. It otherwise shares the feedback
design. The \emph{SMAC3} baseline \citep{lindauer2022smac3} uses version
2.4.0 for CFD-only Gaussian-process Bayesian optimization with 8 Sobol initial
points and 24 adaptive queries, with a common penalty for each nondelivery.
Appendix~\ref{app:formal} gives the belief model's equations and every setting.

\subsection{Optimization agents}
\label{sec:agents}

\begin{figure}[t]
\centering
\includegraphics[width=\linewidth]{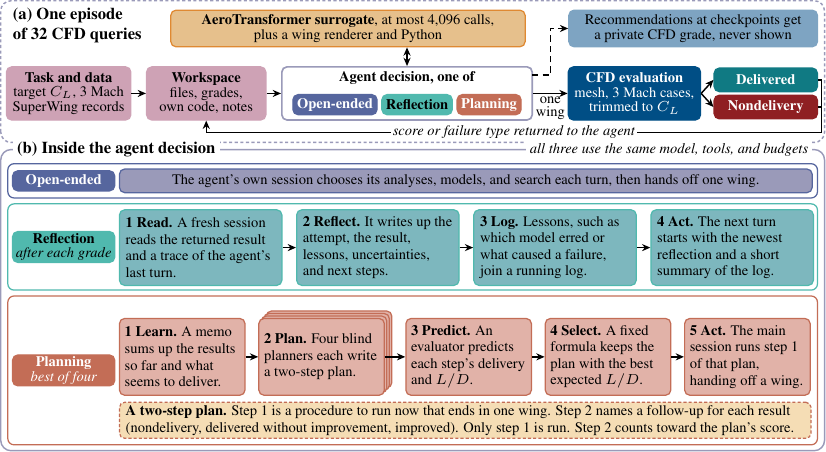}
\caption{Agent episodes and decision procedures. (a) In each episode, the agent
works in its own workspace, may query the AeroTransformer surrogate, and hands
off one wing for each CFD query. The score or failure type is returned, and recommendations nominated at checkpoints are graded without
feedback. (b) The three decision procedures share query caps but can use
different numbers of model calls.}
\label{fig:agent-pipeline}
\end{figure}

The agent study compares three methods with shared task specifications, tools,
CFD and surrogate query caps, separate workspaces, the DeepSeek V4.1 Flash model,
and maximum reasoning-effort setting.
The agent can inspect the supplied SuperWing data,
visualize wing geometries, write and run analysis code,
fit predictive models to the data available within its episode, and run
numerical optimization to propose new designs. Figure~\ref{fig:agent-pipeline}
outlines the steps that every agent episode follows.

\textbf{Open-ended research.} The agent receives the optimization task and
available tools without an imposed planning or reflection template. It may
choose its own analysis, code, and decision process.

\textbf{Explicit reflection.} Following the verbal-feedback approach of
Reflexion \citep{shinn2023reflexion}, the same agent writes a reflection after every
newly returned online grade. The next decision receives the newest full
reflection and a bounded cumulative summary of at most 4{,}000 characters.
Reflection supplements the numerical history rather than replacing it.

\textbf{Best-of-four contingent planning.} A learner first summarizes the
observed environment. Four blind, independent planners are then asked to produce complete
two-step contingent plans. A common value estimator scores the plans by
expected best delivered $L/D$, the first action of the selected plan is
executed, and the process replans after the returned grade. At most three
planning cycles are used per decision. We call this RAFA-motivated because it
learns, plans, acts once, and replans, but its finite horizon and discrete
best-of-four selection differ from RAFA \citep{liu2024rafa}.
Appendix~\ref{app:prompts} maps the method stages to the released prompts.
The implementations include bounded schema/finalization recoveries in nine
episodes by query 4 (Appendix~\ref{app:agent-behavior}); their effects are not
isolated from the prompting methods.

\section{Experimental design and analysis}
\label{sec:analysis}

The \emph{baseline study} contains 150 episodes spanning five methods, three
tasks, and ten seeds. Its 4{,}800 online evaluations and 493 additional recommendation
grades total \TotalCFDEvaluations{} CFD evaluations. For delivery, seed-level method differences are
averaged over the three tasks. For conditional quality, each seed contributes
an average over the tasks where both methods deliver. Seeds without such a
pair are excluded from that contrast. We used exact sign-flip tests and
10{,}000 seed-block bootstrap resamples. Holm correction covers six delivery and
paired-quality tests. These three-task contrasts are exploratory: the frozen
plan designated low and nominal load as core tasks, with high load separate
(Appendix~\ref{app:formal}).

The \emph{agent comparison} contains 90 episodes spanning three methods and
three tasks, with the same ten seed labels for every method. We compared their
recommendations after one, two, and four CFD queries. The four-query point
is an intermediate assessment within the original 32-query task; the agents
were not instructed to optimize specifically for a four-query horizon.
These runs contain
360 online evaluations and 136 additional private recommendation evaluations.
We continued the open-ended agent's 30 episodes to 32 queries for comparison
with the baselines, choosing it for its highest conditional quality on two of the three tasks after four queries.
The extension added 840 online evaluations, bringing both studies to \UnifiedCFDObserved{} evaluation records.
This count follows from the episode design, not a dataset-saturation target.
Requests within episodes are correlated and can reuse results; episodes or
seed blocks, rather than individual requests, are the units of replication.
Ten runs per method/task limit precision: even 10/10 valid recommendations
has an exact 95\% interval of approximately $[0.69,1]$. More independent
episodes would improve precision for smaller differences; fewer would
weaken it. The evaluation count therefore reflects replicated search
trajectories and their cost, not a claim that the design space is exhausted.
For the agent comparison, we report delivery and conditional mean $L/D$ for
each task. At query 4, we estimated differences between open-ended research and
each alternative. Because a seed label does not pair agent episodes, the two
methods' episodes are resampled independently, 10{,}000 times, and quality
differences compare delivered recommendations only. Delivery rates carry
exact binomial intervals and agent delivery differences Newcombe score
intervals (Appendix~\ref{app:formal}). Appendix~\ref{app:fulltables}
lists every result table, Appendix~\ref{app:datacard} the released records, and
Appendix~\ref{app:cfdprotocol} the protocol.

\paragraph{Scope of the comparisons.}
Table~\ref{tab:resource-conditions} states the initial information, surrogate
allowances and online CFD assessment checkpoints.
The four surrogate-assisted baselines share prior information and final query
budgets; how they allocate surrogate calls and use CFD feedback is part of the
comparison. SMAC3 is a CFD-only reference. The three agents share model, source
data, tools and query caps at query 4, but planning and reflection add model
calls. Neither comparison matches total computation or wall time. Cross-family
results are descriptive, not rankings under equal resources; the open-ended
extension was also selected after the four-query comparison.
Appendices~\ref{app:resources} and~\ref{app:resource-followup} report original and follow-up resource measurements.

\begin{table}[htbp]
\centering\scriptsize
\caption{Initial information, surrogate allowances and online CFD assessment
checkpoints. All methods were assigned a 32-query task; the common agent
comparison uses its first four queries. Private recommendation grades are
additional and are not returned to the method.}
\label{tab:resource-conditions}
\setlength{\tabcolsep}{3pt}
\begin{tabular*}{\textwidth}{@{\extracolsep{\fill}}llrr}
\toprule
Method group & Prior CFD context & Surrogate cap & \shortstack{Online CFD\\checkpoint} \\
\midrule
Four surrogate baselines & Same 62 development records & 4{,}096 & 32 \\
SMAC3 & No warm start & 0 & 32 \\
Three agents & SuperWing source; not the 62 records & 4{,}096 & 4 \\
Open-ended extension & Continued episode history & Same episode cap & 32 \\
\bottomrule
\end{tabular*}
\end{table}

\section{Results and discussion}
\label{sec:results}
\label{sec:discussion}

\subsection{Baseline optimizer performance}

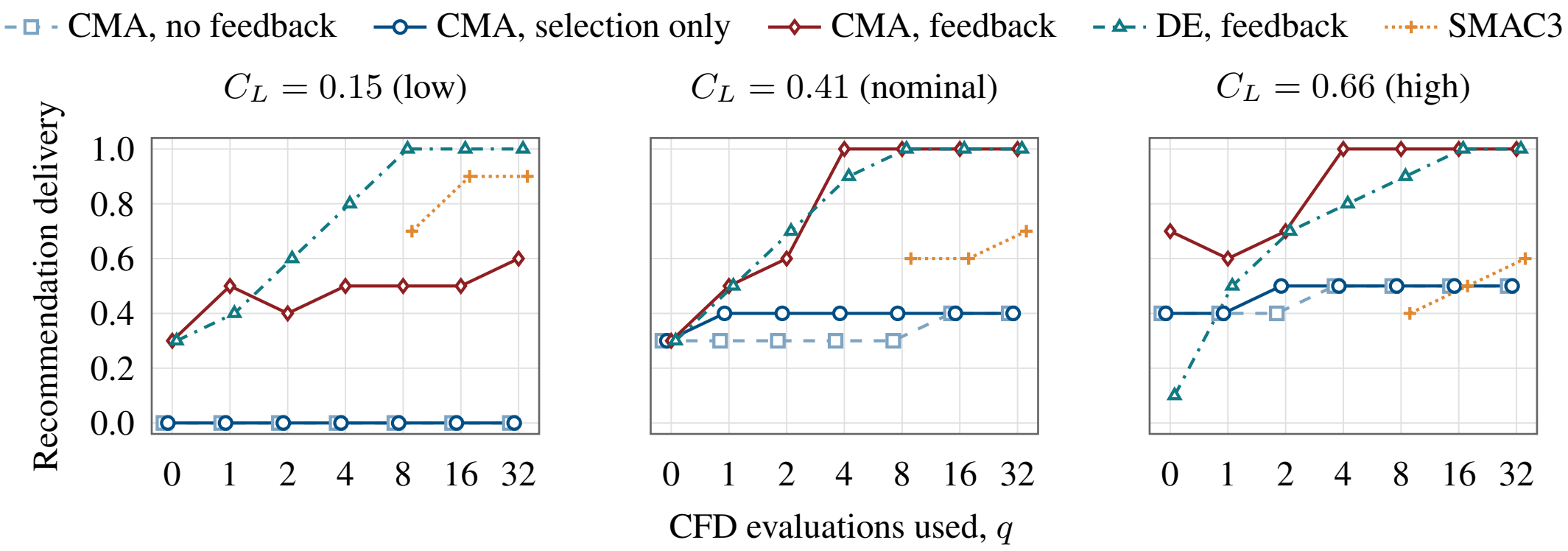
\begin{figure}[t]
\centering
{\footnotesize \wbkey{cma no feedback}{CMA, no feedback}\hspace{1em}\wbkey{cma selection}{CMA, selection only}\hspace{1em}\wbkey{cma feedback}{CMA, feedback}\hspace{1em}\wbkey{de feedback}{DE, feedback}\hspace{1em}\wbkey{smac3}{SMAC3}}\par\smallskip
\begin{tikzpicture}[wbfig]
\node[wblabel,rotate=90] at (0,45) {Recommendation delivery};
\node[wblabel] at (270,-29) {CFD evaluations used, $q$};
\begin{scope}[shift={(35,0)}]
\node[wblabel] at (66,105) {$C_L=0.15$ (low)};
\draw[wbgrid] (0,3.333) -- (132,3.333);
\node[anchor=east,inner sep=1pt] at (-4,3.333) {0.0};
\draw[wbgrid] (0,20.000) -- (132,20.000);
\node[anchor=east,inner sep=1pt] at (-4,20.000) {0.2};
\draw[wbgrid] (0,36.667) -- (132,36.667);
\node[anchor=east,inner sep=1pt] at (-4,36.667) {0.4};
\draw[wbgrid] (0,53.333) -- (132,53.333);
\node[anchor=east,inner sep=1pt] at (-4,53.333) {0.6};
\draw[wbgrid] (0,70.000) -- (132,70.000);
\node[anchor=east,inner sep=1pt] at (-4,70.000) {0.8};
\draw[wbgrid] (0,86.667) -- (132,86.667);
\node[anchor=east,inner sep=1pt] at (-4,86.667) {1.0};
\draw[wbgrid] (7.000,0) -- (7.000,90);
\node at (7.000,-12) {0};
\draw[wbgrid] (26.667,0) -- (26.667,90);
\node at (26.667,-12) {1};
\draw[wbgrid] (46.333,0) -- (46.333,90);
\node at (46.333,-12) {2};
\draw[wbgrid] (66.000,0) -- (66.000,90);
\node at (66.000,-12) {4};
\draw[wbgrid] (85.667,0) -- (85.667,90);
\node at (85.667,-12) {8};
\draw[wbgrid] (105.333,0) -- (105.333,90);
\node at (105.333,-12) {16};
\draw[wbgrid] (125.000,0) -- (125.000,90);
\node at (125.000,-12) {32};
\draw[wbframe] (0,0) rectangle (132,90);
\draw[cma no feedback] plot coordinates {(4.000,3.333) (23.667,3.333) (43.333,3.333) (63.000,3.333) (82.667,3.333) (102.333,3.333) (122.000,3.333)};
\draw[cma selection] plot coordinates {(5.500,3.333) (25.167,3.333) (44.833,3.333) (64.500,3.333) (84.167,3.333) (103.833,3.333) (123.500,3.333)};
\draw[cma feedback] plot coordinates {(7.000,28.333) (26.667,45.000) (46.333,36.667) (66.000,45.000) (85.667,45.000) (105.333,45.000) (125.000,53.333)};
\draw[de feedback] plot coordinates {(8.500,28.333) (28.167,36.667) (47.833,53.333) (67.500,70.000) (87.167,86.667) (106.833,86.667) (126.500,86.667)};
\draw[smac3] plot coordinates {(88.667,61.667) (108.333,78.333) (128.000,78.333)};
\draw[cma no feedback marks] plot coordinates {(4.000,3.333) (23.667,3.333) (43.333,3.333) (63.000,3.333) (82.667,3.333) (102.333,3.333) (122.000,3.333)};
\draw[cma selection marks] plot coordinates {(5.500,3.333) (25.167,3.333) (44.833,3.333) (64.500,3.333) (84.167,3.333) (103.833,3.333) (123.500,3.333)};
\draw[cma feedback marks] plot coordinates {(7.000,28.333) (26.667,45.000) (46.333,36.667) (66.000,45.000) (85.667,45.000) (105.333,45.000) (125.000,53.333)};
\draw[de feedback marks] plot coordinates {(8.500,28.333) (28.167,36.667) (47.833,53.333) (67.500,70.000) (87.167,86.667) (106.833,86.667) (126.500,86.667)};
\draw[smac3 marks] plot coordinates {(88.667,61.667) (108.333,78.333) (128.000,78.333)};
\end{scope}
\begin{scope}[shift={(205,0)}]
\node[wblabel] at (66,105) {$C_L=0.41$ (nominal)};
\draw[wbgrid] (0,3.333) -- (132,3.333);
\draw[wbgrid] (0,20.000) -- (132,20.000);
\draw[wbgrid] (0,36.667) -- (132,36.667);
\draw[wbgrid] (0,53.333) -- (132,53.333);
\draw[wbgrid] (0,70.000) -- (132,70.000);
\draw[wbgrid] (0,86.667) -- (132,86.667);
\draw[wbgrid] (7.000,0) -- (7.000,90);
\node at (7.000,-12) {0};
\draw[wbgrid] (26.667,0) -- (26.667,90);
\node at (26.667,-12) {1};
\draw[wbgrid] (46.333,0) -- (46.333,90);
\node at (46.333,-12) {2};
\draw[wbgrid] (66.000,0) -- (66.000,90);
\node at (66.000,-12) {4};
\draw[wbgrid] (85.667,0) -- (85.667,90);
\node at (85.667,-12) {8};
\draw[wbgrid] (105.333,0) -- (105.333,90);
\node at (105.333,-12) {16};
\draw[wbgrid] (125.000,0) -- (125.000,90);
\node at (125.000,-12) {32};
\draw[wbframe] (0,0) rectangle (132,90);
\draw[cma no feedback] plot coordinates {(4.000,28.333) (23.667,28.333) (43.333,28.333) (63.000,28.333) (82.667,28.333) (102.333,36.667) (122.000,36.667)};
\draw[cma selection] plot coordinates {(5.500,28.333) (25.167,36.667) (44.833,36.667) (64.500,36.667) (84.167,36.667) (103.833,36.667) (123.500,36.667)};
\draw[cma feedback] plot coordinates {(7.000,28.333) (26.667,45.000) (46.333,53.333) (66.000,86.667) (85.667,86.667) (105.333,86.667) (125.000,86.667)};
\draw[de feedback] plot coordinates {(8.500,28.333) (28.167,45.000) (47.833,61.667) (67.500,78.333) (87.167,86.667) (106.833,86.667) (126.500,86.667)};
\draw[smac3] plot coordinates {(88.667,53.333) (108.333,53.333) (128.000,61.667)};
\draw[cma no feedback marks] plot coordinates {(4.000,28.333) (23.667,28.333) (43.333,28.333) (63.000,28.333) (82.667,28.333) (102.333,36.667) (122.000,36.667)};
\draw[cma selection marks] plot coordinates {(5.500,28.333) (25.167,36.667) (44.833,36.667) (64.500,36.667) (84.167,36.667) (103.833,36.667) (123.500,36.667)};
\draw[cma feedback marks] plot coordinates {(7.000,28.333) (26.667,45.000) (46.333,53.333) (66.000,86.667) (85.667,86.667) (105.333,86.667) (125.000,86.667)};
\draw[de feedback marks] plot coordinates {(8.500,28.333) (28.167,45.000) (47.833,61.667) (67.500,78.333) (87.167,86.667) (106.833,86.667) (126.500,86.667)};
\draw[smac3 marks] plot coordinates {(88.667,53.333) (108.333,53.333) (128.000,61.667)};
\end{scope}
\begin{scope}[shift={(375,0)}]
\node[wblabel] at (66,105) {$C_L=0.66$ (high)};
\draw[wbgrid] (0,3.333) -- (132,3.333);
\draw[wbgrid] (0,20.000) -- (132,20.000);
\draw[wbgrid] (0,36.667) -- (132,36.667);
\draw[wbgrid] (0,53.333) -- (132,53.333);
\draw[wbgrid] (0,70.000) -- (132,70.000);
\draw[wbgrid] (0,86.667) -- (132,86.667);
\draw[wbgrid] (7.000,0) -- (7.000,90);
\node at (7.000,-12) {0};
\draw[wbgrid] (26.667,0) -- (26.667,90);
\node at (26.667,-12) {1};
\draw[wbgrid] (46.333,0) -- (46.333,90);
\node at (46.333,-12) {2};
\draw[wbgrid] (66.000,0) -- (66.000,90);
\node at (66.000,-12) {4};
\draw[wbgrid] (85.667,0) -- (85.667,90);
\node at (85.667,-12) {8};
\draw[wbgrid] (105.333,0) -- (105.333,90);
\node at (105.333,-12) {16};
\draw[wbgrid] (125.000,0) -- (125.000,90);
\node at (125.000,-12) {32};
\draw[wbframe] (0,0) rectangle (132,90);
\draw[cma no feedback] plot coordinates {(4.000,36.667) (23.667,36.667) (43.333,36.667) (63.000,45.000) (82.667,45.000) (102.333,45.000) (122.000,45.000)};
\draw[cma selection] plot coordinates {(5.500,36.667) (25.167,36.667) (44.833,45.000) (64.500,45.000) (84.167,45.000) (103.833,45.000) (123.500,45.000)};
\draw[cma feedback] plot coordinates {(7.000,61.667) (26.667,53.333) (46.333,61.667) (66.000,86.667) (85.667,86.667) (105.333,86.667) (125.000,86.667)};
\draw[de feedback] plot coordinates {(8.500,11.667) (28.167,45.000) (47.833,61.667) (67.500,70.000) (87.167,78.333) (106.833,86.667) (126.500,86.667)};
\draw[smac3] plot coordinates {(88.667,36.667) (108.333,45.000) (128.000,53.333)};
\draw[cma no feedback marks] plot coordinates {(4.000,36.667) (23.667,36.667) (43.333,36.667) (63.000,45.000) (82.667,45.000) (102.333,45.000) (122.000,45.000)};
\draw[cma selection marks] plot coordinates {(5.500,36.667) (25.167,36.667) (44.833,45.000) (64.500,45.000) (84.167,45.000) (103.833,45.000) (123.500,45.000)};
\draw[cma feedback marks] plot coordinates {(7.000,61.667) (26.667,53.333) (46.333,61.667) (66.000,86.667) (85.667,86.667) (105.333,86.667) (125.000,86.667)};
\draw[de feedback marks] plot coordinates {(8.500,11.667) (28.167,45.000) (47.833,61.667) (67.500,70.000) (87.167,78.333) (106.833,86.667) (126.500,86.667)};
\draw[smac3 marks] plot coordinates {(88.667,36.667) (108.333,45.000) (128.000,53.333)};
\end{scope}
\end{tikzpicture}
\caption{Recommendation delivery for the five baseline optimizers. A method
keeps its best delivered design once it has one. Before that, it recommends
untested designs, so a curve can fall. SMAC3 is
shown from query 8 onward, omitting the early checkpoints within its Sobol
initialization.}
\label{fig:baseline-curves}
\end{figure}

At 32 queries, feedback-guided CMA-ES combines reliable delivery with strong
aerodynamic quality at nominal and high load. It delivers in all ten episodes
on both tasks, with conditional mean $L/D$ of 33.04 and 31.64. The archive-only CMA-ES methods find slightly stronger wings among their
successful runs, averaging about 33.2 and 32.1, but each delivers in only four
nominal-load and five high-load episodes (Figure~\ref{fig:baseline-curves}).
Both archive-only methods fail to deliver at low load.
The nominal-load comparison illustrates why agent evaluation must report
recommendation reliability alongside quality: the archive-only conditional
mean is slightly higher, but excludes six unsuccessful recommendations per
method. Improving quality among valid recommendations alone can therefore
leave the reliability of the returned design unresolved.

The three CMA-ES variants suggest that feedback is most useful when it changes
future proposals. Reordering an existing archive cannot escape shared meshing
difficulties or surrogate errors, whereas feedback-guided CMA-ES can redirect
its search. Its higher online delivery rates show that adaptation changes
which designs are tested, not just which recommendation is retained
(Table~\ref{tab:baseline-yield}). The variants also allocate cheap evaluations
differently, completing them up front or interleaving them with CFD. Their
comparison therefore tests whole search strategies, not failure modeling alone.
For agent design, this motivates using CFD evidence to revise candidate
generation, with reranking a fixed candidate set as a distinct comparator.
\begin{table}[t]
\caption{Online CFD delivery rate in the baseline study.}
\label{tab:baseline-yield}
\centering
\scriptsize
\setlength{\tabcolsep}{4pt}
\begin{tabular*}{\textwidth}{@{\extracolsep{\fill}}lccc}
\toprule
method & $C_L=0.15$ & $C_L=0.41$ & $C_L=0.66$ \\
\midrule
CMA-ES, no feedback & 0.00 & 0.26 & 0.37 \\
CMA-ES, selection only & 0.00 & 0.27 & 0.35 \\
Feedback-guided CMA-ES & 0.04 & 0.77 & 0.77 \\
Feedback-guided DE & 0.29 & 0.62 & 0.62 \\
SMAC3 & \SMACLowOnlineRate & \SMACNomOnlineRate & \SMACHighOnlineRate \\
\bottomrule
\end{tabular*}
\end{table}
Feedback-guided DE also delivers in every nominal- and high-load episode,
but mean quality is lower at 31.09 and 27.05. At low load, DE delivers in all ten episodes with mean $L/D$ of 17.52,
compared with six deliveries and 16.76 for feedback-guided CMA-ES. Since both
use the same feedback model, the generator remains important. DE
may preserve useful alternatives where CMA-ES concentrates too narrowly,
while CMA-ES may refine productive regions more effectively on other
tasks.
SMAC3 delivers in \SMACLowFinalDelivered, \SMACNominalFinalDelivered, and
\SMACHighFinalDelivered{} episodes at low, nominal, and high load, but its
conditional mean $L/D$ is much lower at \SMACLowFinalMean,
\SMACNominalFinalMean, and \SMACHighFinalMean. Unlike the other baselines,
it learns from expensive feedback without a pretrained aerodynamic model.
With only 32 evaluations in 38 dimensions, with many failing,
its resources are limited. This is
a plausible explanation for its low quality.

The exploratory three-task final-recommendation contrasts quantify the
reliability-quality tradeoff.
Relative to both archive-only CMA-ES variants, feedback-guided CMA-ES improves
delivery by 0.57 averaged across tasks (95\% bootstrap interval [0.37, 0.77],
Holm $p=0.039$). Its paired conditional-quality differences are about $-0.35$,
with bootstrap intervals of [$-0.79$, $-0.01$] and [$-0.80$, $-0.01$], but seven
pairs leave the exploratory sign-flip tests underpowered (Holm $p=0.28$). Relative to the
selection-only method, feedback-guided DE improves delivery by 0.70
[0.57, 0.83] (Holm $p=0.012$) but loses 3.50 $L/D$ [2.28, 4.94] among
paired delivered recommendations (Holm $p=0.047$).
At nominal load, feedback-guided CMA-ES largely resolves the tradeoff with complete delivery at similar quality.
This leaves less room for improvement at query 32, although differences
at other loads rule out uniform saturation. At query 4, the nominal-load
archive-only methods already average about 33.2 but deliver in three or
four episodes, while feedback-guided CMA-ES delivers in all ten at 28.73.

\subsection{Agent performance over 32 queries}

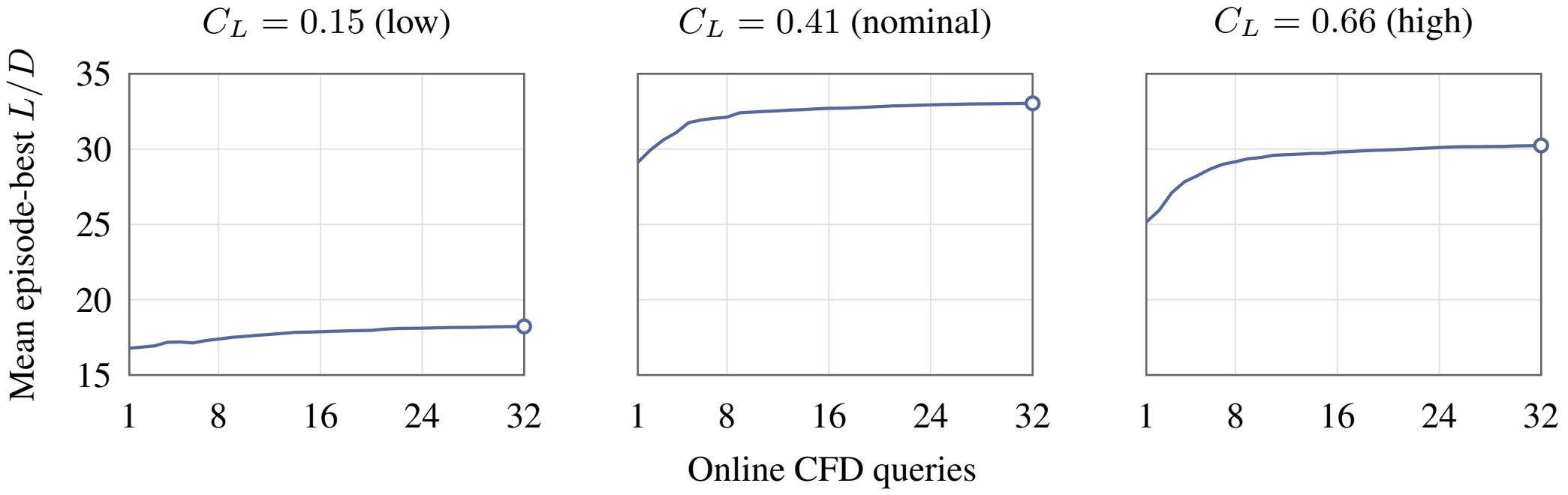
\begin{figure}[t]
\centering
\begin{tikzpicture}[wbfig]
\node[wblabel,rotate=90] at (0,45) {Mean episode-best $L/D$};
\node[wblabel] at (270,-29) {Online CFD queries};
\begin{scope}[shift={(35,0)}]
\draw[wbgrid] (0,0.000) -- (132,0.000);
\node[anchor=east,inner sep=1pt] at (-4,0.000) {15};
\draw[wbgrid] (0,22.500) -- (132,22.500);
\node[anchor=east,inner sep=1pt] at (-4,22.500) {20};
\draw[wbgrid] (0,45.000) -- (132,45.000);
\node[anchor=east,inner sep=1pt] at (-4,45.000) {25};
\draw[wbgrid] (0,67.500) -- (132,67.500);
\node[anchor=east,inner sep=1pt] at (-4,67.500) {30};
\draw[wbgrid] (0,90.000) -- (132,90.000);
\node[anchor=east,inner sep=1pt] at (-4,90.000) {35};
\node[wblabel] at (66,105) {$C_L=0.15$ (low)};
\draw[wbgrid] (0.000,0) -- (0.000,90);
\node at (0.000,-12) {1};
\draw[wbgrid] (29.806,0) -- (29.806,90);
\node at (29.806,-12) {8};
\draw[wbgrid] (63.871,0) -- (63.871,90);
\node at (63.871,-12) {16};
\draw[wbgrid] (97.935,0) -- (97.935,90);
\node at (97.935,-12) {24};
\draw[wbgrid] (132.000,0) -- (132.000,90);
\node at (132.000,-12) {32};
\draw[wbframe] (0,0) rectangle (132,90);
\draw[open research] plot coordinates {(0.000,7.999) (4.258,8.354) (8.516,8.721) (12.774,9.791) (17.032,9.873) (21.290,9.596) (25.548,10.291) (29.806,10.737) (34.065,11.243) (38.323,11.519) (42.581,11.867) (46.839,12.122) (51.097,12.432) (55.355,12.762) (59.613,12.822) (63.871,12.927) (68.129,13.060) (72.387,13.170) (76.645,13.275) (80.903,13.343) (85.161,13.701) (89.419,13.898) (93.677,13.953) (97.935,13.993) (102.194,14.096) (106.452,14.180) (110.710,14.228) (114.968,14.250) (119.226,14.342) (123.484,14.421) (127.742,14.474) (132.000,14.525)};
\draw[open research marks] plot coordinates {(132.000,14.525)};
\end{scope}
\begin{scope}[shift={(205,0)}]
\draw[wbgrid] (0,0.000) -- (132,0.000);
\draw[wbgrid] (0,22.500) -- (132,22.500);
\draw[wbgrid] (0,45.000) -- (132,45.000);
\draw[wbgrid] (0,67.500) -- (132,67.500);
\draw[wbgrid] (0,90.000) -- (132,90.000);
\node[wblabel] at (66,105) {$C_L=0.41$ (nominal)};
\draw[wbgrid] (0.000,0) -- (0.000,90);
\node at (0.000,-12) {1};
\draw[wbgrid] (29.806,0) -- (29.806,90);
\node at (29.806,-12) {8};
\draw[wbgrid] (63.871,0) -- (63.871,90);
\node at (63.871,-12) {16};
\draw[wbgrid] (97.935,0) -- (97.935,90);
\node at (97.935,-12) {24};
\draw[wbgrid] (132.000,0) -- (132.000,90);
\node at (132.000,-12) {32};
\draw[wbframe] (0,0) rectangle (132,90);
\draw[open research] plot coordinates {(0.000,63.526) (4.258,67.266) (8.516,70.216) (12.774,72.385) (17.032,75.408) (21.290,76.214) (25.548,76.688) (29.806,77.034) (34.065,78.333) (38.323,78.557) (42.581,78.752) (46.839,78.928) (51.097,79.162) (55.355,79.270) (59.613,79.512) (63.871,79.680) (68.129,79.719) (72.387,79.859) (76.645,80.020) (80.903,80.184) (85.161,80.388) (89.419,80.464) (93.677,80.594) (97.935,80.706) (102.194,80.815) (106.452,80.870) (110.710,80.965) (114.968,80.999) (119.226,81.034) (123.484,81.078) (127.742,81.122) (132.000,81.172)};
\draw[open research marks] plot coordinates {(132.000,81.172)};
\end{scope}
\begin{scope}[shift={(375,0)}]
\draw[wbgrid] (0,0.000) -- (132,0.000);
\draw[wbgrid] (0,22.500) -- (132,22.500);
\draw[wbgrid] (0,45.000) -- (132,45.000);
\draw[wbgrid] (0,67.500) -- (132,67.500);
\draw[wbgrid] (0,90.000) -- (132,90.000);
\node[wblabel] at (66,105) {$C_L=0.66$ (high)};
\draw[wbgrid] (0.000,0) -- (0.000,90);
\node at (0.000,-12) {1};
\draw[wbgrid] (29.806,0) -- (29.806,90);
\node at (29.806,-12) {8};
\draw[wbgrid] (63.871,0) -- (63.871,90);
\node at (63.871,-12) {16};
\draw[wbgrid] (97.935,0) -- (97.935,90);
\node at (97.935,-12) {24};
\draw[wbgrid] (132.000,0) -- (132.000,90);
\node at (132.000,-12) {32};
\draw[wbframe] (0,0) rectangle (132,90);
\draw[open research] plot coordinates {(0.000,45.694) (4.258,49.163) (8.516,54.458) (12.774,57.729) (17.032,59.533) (21.290,61.520) (25.548,62.952) (29.806,63.728) (34.065,64.625) (38.323,64.976) (42.581,65.650) (46.839,65.825) (51.097,65.991) (55.355,66.196) (59.613,66.219) (63.871,66.631) (68.129,66.751) (72.387,66.980) (76.645,67.130) (80.903,67.263) (85.161,67.403) (89.419,67.598) (93.677,67.772) (97.935,67.975) (102.194,68.142) (106.452,68.209) (110.710,68.211) (114.968,68.275) (119.226,68.291) (123.484,68.446) (127.742,68.487) (132.000,68.583)};
\draw[open research marks] plot coordinates {(132.000,68.583)};
\end{scope}
\end{tikzpicture}
\caption{Mean episode-best $L/D$ during the open-ended search. Each curve
conditions on episodes with at least one delivered design. A retained best
design cannot deteriorate within an episode, but the population contributing
to the mean can change when another episode first delivers.}
\label{fig:agent-search-quality}
\end{figure}

At 32 queries, open-ended research finds designs of similar quality to the
strongest baselines under the different resource conditions in
Table~\ref{tab:resource-conditions}. For the baselines,
recommendation outcomes at query 32 match their best delivered online designs.
Every delivered recommendation is that design, and no episode without an online
delivery delivers its recommendation. We report the agent's best delivered
online design separately from its standing recommendation. Its
low-load mean of \AgentLowBestAtC{} exceeds DE's 17.52, while its nominal-load
mean of \AgentNominalBestAtC{} is essentially identical to feedback-guided
CMA-ES (33.038 and 33.040). At high load, its \AgentHighBestAtC{} exceeds DE's
27.05 but remains below CMA-ES's 31.64. All ten agent episodes find a delivered
online design on every task. The extension requested no further recommendations,
so we also score each episode's last nomination as its standing recommendation.
It is a delivered wing in 28 of 30 episodes, nine each at low and high load and
ten at nominal load, with mean $L/D$ of 18.21, 33.03, and 30.43, and the best
delivered wing in 20 (Appendix~\ref{app:agent-comparison}).
Table~\ref{tab:agent-extension} and Figure~\ref{fig:agent-search-quality}
show how this performance develops. The scores are episode maxima among
delivered online designs, not private recommendation assessments.
Ten of the 960 online evaluations failed in the grading service before CFD
began and were charged to the agent against the released rules
(Appendix~\ref{app:agent-comparison}). Regraded after the study, the high-load wing was unmeshable, three of the other four delivered, one above its episode's best online score, and the fourth reached the time limit at one of its three conditions.
Appendix~\ref{app:gallery} shows the
best delivered wing of every method and task, and Appendix~\ref{app:figures}
the outcome composition of the online evaluations.

\begin{table}[t]
\caption{Open-ended research over a longer search. Best $L/D$ is the mean
episode maximum over delivered online designs (nine low-load episodes have one
at query 4, ten in every other cell).}
\label{tab:agent-extension}
\centering
\scriptsize
\setlength{\tabcolsep}{4pt}
\begin{tabular*}{\textwidth}{@{\extracolsep{\fill}}lrrrr}
\toprule
$C_L$ & online delivery through query 32 & best $L/D$ at 4 & at 16 & at 32 \\
\midrule
$0.15$ & 0.62 & 17.18 & 17.87 & 18.23 \\
$0.41$ & 0.70 & 31.09 & 32.71 & 33.04 \\
$0.66$ & 0.62 & 27.83 & 29.81 & 30.24 \\
\bottomrule
\end{tabular*}
\end{table}

Better designs need not come with higher delivery rates.
At high load, online delivery falls from 0.85 over the first four queries to
0.625 over the first 16 queries, even as mean episode-best $L/D$ improves.
This is consistent with exploring less reliable regions while retaining
strong designs.

The largest three-condition score across the studies is \CampaignMaxLd{},
compared with a maximum single-condition $L/D$ of 29.4 in the SuperWing
source records. This exceeds the dataset's recorded range but is not a
matched estimate of improvement over its best wing. SuperWing samples
operating conditions, whereas WingBench trims lift and aggregates three
Mach conditions.

\subsection{Agent recommendations at four queries}

\begin{figure}[t]
\centering
{\footnotesize \wbkey{open research}{Open-ended research}\hspace{2em}\wbkey{planning}{Planning}\hspace{2em}\wbkey{reflection}{Reflection}}\par\smallskip
\begin{tikzpicture}[wbfig]
\node[wblabel,rotate=90] at (0,167) {Recommendation delivery};
\node[wblabel,rotate=90] at (0,42) {Mean delivered $L/D$};
\node[wblabel] at (270,-29) {CFD queries};
\node[wblabel] at (101,216) {$C_L=0.15$ (low)};
\begin{scope}[shift={(35,120)}]
\draw[wbgrid] (0,3.864) -- (132,3.864);
\node[anchor=east,inner sep=1pt] at (-4,3.864) {0.0};
\draw[wbgrid] (0,19.318) -- (132,19.318);
\node[anchor=east,inner sep=1pt] at (-4,19.318) {0.2};
\draw[wbgrid] (0,34.773) -- (132,34.773);
\node[anchor=east,inner sep=1pt] at (-4,34.773) {0.4};
\draw[wbgrid] (0,50.227) -- (132,50.227);
\node[anchor=east,inner sep=1pt] at (-4,50.227) {0.6};
\draw[wbgrid] (0,65.682) -- (132,65.682);
\node[anchor=east,inner sep=1pt] at (-4,65.682) {0.8};
\draw[wbgrid] (0,81.136) -- (132,81.136);
\node[anchor=east,inner sep=1pt] at (-4,81.136) {1.0};
\draw[wbgrid] (10.000,0) -- (10.000,85);
\node at (10.000,-12) {1};
\draw[wbgrid] (47.333,0) -- (47.333,85);
\node at (47.333,-12) {2};
\draw[wbgrid] (122.000,0) -- (122.000,85);
\node at (122.000,-12) {4};
\draw[wbframe] (0,0) rectangle (132,85);
\draw[open research] plot coordinates {(3.500,57.955) (40.833,65.682) (115.500,73.409)};
\draw[planning] plot coordinates {(10.000,81.136) (47.333,81.136) (122.000,81.136)};
\draw[reflection] plot coordinates {(16.500,65.682) (53.833,81.136) (128.500,81.136)};
\draw[open research marks] plot coordinates {(3.500,57.955) (40.833,65.682) (115.500,73.409)};
\draw[planning marks] plot coordinates {(10.000,81.136) (47.333,81.136) (122.000,81.136)};
\draw[reflection marks] plot coordinates {(16.500,65.682) (53.833,81.136) (128.500,81.136)};
\end{scope}
\begin{scope}[shift={(35,0)}]
\draw[wbgrid] (0,0.000) -- (132,0.000);
\node[anchor=east,inner sep=1pt] at (-4,0.000) {15};
\draw[wbgrid] (0,21.250) -- (132,21.250);
\node[anchor=east,inner sep=1pt] at (-4,21.250) {20};
\draw[wbgrid] (0,42.500) -- (132,42.500);
\node[anchor=east,inner sep=1pt] at (-4,42.500) {25};
\draw[wbgrid] (0,63.750) -- (132,63.750);
\node[anchor=east,inner sep=1pt] at (-4,63.750) {30};
\draw[wbgrid] (0,85.000) -- (132,85.000);
\node[anchor=east,inner sep=1pt] at (-4,85.000) {35};
\draw[wbgrid] (10.000,0) -- (10.000,85);
\node at (10.000,-12) {1};
\draw[wbgrid] (47.333,0) -- (47.333,85);
\node at (47.333,-12) {2};
\draw[wbgrid] (122.000,0) -- (122.000,85);
\node at (122.000,-12) {4};
\draw[wbframe] (0,0) rectangle (132,85);
\draw[open research] plot coordinates {(3.500,6.106) (40.833,8.283) (115.500,9.118)};
\draw[planning] plot coordinates {(10.000,6.977) (47.333,8.526) (122.000,10.240)};
\draw[reflection] plot coordinates {(16.500,6.763) (53.833,8.426) (128.500,9.922)};
\draw[open research marks] plot coordinates {(3.500,6.106) (40.833,8.283) (115.500,9.118)};
\draw[planning marks] plot coordinates {(10.000,6.977) (47.333,8.526) (122.000,10.240)};
\draw[reflection marks] plot coordinates {(16.500,6.763) (53.833,8.426) (128.500,9.922)};
\end{scope}
\node[wblabel] at (271,216) {$C_L=0.41$ (nominal)};
\begin{scope}[shift={(205,120)}]
\draw[wbgrid] (0,3.864) -- (132,3.864);
\draw[wbgrid] (0,19.318) -- (132,19.318);
\draw[wbgrid] (0,34.773) -- (132,34.773);
\draw[wbgrid] (0,50.227) -- (132,50.227);
\draw[wbgrid] (0,65.682) -- (132,65.682);
\draw[wbgrid] (0,81.136) -- (132,81.136);
\draw[wbgrid] (10.000,0) -- (10.000,85);
\node at (10.000,-12) {1};
\draw[wbgrid] (47.333,0) -- (47.333,85);
\node at (47.333,-12) {2};
\draw[wbgrid] (122.000,0) -- (122.000,85);
\node at (122.000,-12) {4};
\draw[wbframe] (0,0) rectangle (132,85);
\draw[open research] plot coordinates {(3.500,73.409) (40.833,73.409) (115.500,81.136)};
\draw[planning] plot coordinates {(10.000,73.409) (47.333,73.409) (122.000,73.409)};
\draw[reflection] plot coordinates {(16.500,81.136) (53.833,81.136) (128.500,81.136)};
\draw[open research marks] plot coordinates {(3.500,73.409) (40.833,73.409) (115.500,81.136)};
\draw[planning marks] plot coordinates {(10.000,73.409) (47.333,73.409) (122.000,73.409)};
\draw[reflection marks] plot coordinates {(16.500,81.136) (53.833,81.136) (128.500,81.136)};
\end{scope}
\begin{scope}[shift={(205,0)}]
\draw[wbgrid] (0,0.000) -- (132,0.000);
\draw[wbgrid] (0,21.250) -- (132,21.250);
\draw[wbgrid] (0,42.500) -- (132,42.500);
\draw[wbgrid] (0,63.750) -- (132,63.750);
\draw[wbgrid] (0,85.000) -- (132,85.000);
\draw[wbgrid] (10.000,0) -- (10.000,85);
\node at (10.000,-12) {1};
\draw[wbgrid] (47.333,0) -- (47.333,85);
\node at (47.333,-12) {2};
\draw[wbgrid] (122.000,0) -- (122.000,85);
\node at (122.000,-12) {4};
\draw[wbframe] (0,0) rectangle (132,85);
\draw[open research] plot coordinates {(3.500,58.058) (40.833,62.854) (115.500,68.364)};
\draw[planning] plot coordinates {(10.000,56.426) (47.333,57.096) (122.000,60.829)};
\draw[reflection] plot coordinates {(16.500,57.021) (53.833,59.714) (128.500,63.014)};
\draw[open research marks] plot coordinates {(3.500,58.058) (40.833,62.854) (115.500,68.364)};
\draw[planning marks] plot coordinates {(10.000,56.426) (47.333,57.096) (122.000,60.829)};
\draw[reflection marks] plot coordinates {(16.500,57.021) (53.833,59.714) (128.500,63.014)};
\end{scope}
\node[wblabel] at (441,216) {$C_L=0.66$ (high)};
\begin{scope}[shift={(375,120)}]
\draw[wbgrid] (0,3.864) -- (132,3.864);
\draw[wbgrid] (0,19.318) -- (132,19.318);
\draw[wbgrid] (0,34.773) -- (132,34.773);
\draw[wbgrid] (0,50.227) -- (132,50.227);
\draw[wbgrid] (0,65.682) -- (132,65.682);
\draw[wbgrid] (0,81.136) -- (132,81.136);
\draw[wbgrid] (10.000,0) -- (10.000,85);
\node at (10.000,-12) {1};
\draw[wbgrid] (47.333,0) -- (47.333,85);
\node at (47.333,-12) {2};
\draw[wbgrid] (122.000,0) -- (122.000,85);
\node at (122.000,-12) {4};
\draw[wbframe] (0,0) rectangle (132,85);
\draw[open research] plot coordinates {(3.500,81.136) (40.833,81.136) (115.500,81.136)};
\draw[planning] plot coordinates {(10.000,65.682) (47.333,73.409) (122.000,73.409)};
\draw[reflection] plot coordinates {(16.500,81.136) (53.833,81.136) (128.500,81.136)};
\draw[open research marks] plot coordinates {(3.500,81.136) (40.833,81.136) (115.500,81.136)};
\draw[planning marks] plot coordinates {(10.000,65.682) (47.333,73.409) (122.000,73.409)};
\draw[reflection marks] plot coordinates {(16.500,81.136) (53.833,81.136) (128.500,81.136)};
\end{scope}
\begin{scope}[shift={(375,0)}]
\draw[wbgrid] (0,0.000) -- (132,0.000);
\draw[wbgrid] (0,21.250) -- (132,21.250);
\draw[wbgrid] (0,42.500) -- (132,42.500);
\draw[wbgrid] (0,63.750) -- (132,63.750);
\draw[wbgrid] (0,85.000) -- (132,85.000);
\draw[wbgrid] (10.000,0) -- (10.000,85);
\node at (10.000,-12) {1};
\draw[wbgrid] (47.333,0) -- (47.333,85);
\node at (47.333,-12) {2};
\draw[wbgrid] (122.000,0) -- (122.000,85);
\node at (122.000,-12) {4};
\draw[wbframe] (0,0) rectangle (132,85);
\draw[open research] plot coordinates {(3.500,42.148) (40.833,47.602) (115.500,54.521)};
\draw[planning] plot coordinates {(10.000,47.146) (47.333,49.321) (122.000,53.515)};
\draw[reflection] plot coordinates {(16.500,40.508) (53.833,45.670) (128.500,50.091)};
\draw[open research marks] plot coordinates {(3.500,42.148) (40.833,47.602) (115.500,54.521)};
\draw[planning marks] plot coordinates {(10.000,47.146) (47.333,49.321) (122.000,53.515)};
\draw[reflection marks] plot coordinates {(16.500,40.508) (53.833,45.670) (128.500,50.091)};
\end{scope}
\end{tikzpicture}
\caption{Agent recommendation delivery and conditional mean $L/D$ after one,
two, and four CFD queries. Quality
is averaged over the delivered subset at each checkpoint, which can change.}
\label{fig:agent-recommendations}
\end{figure}
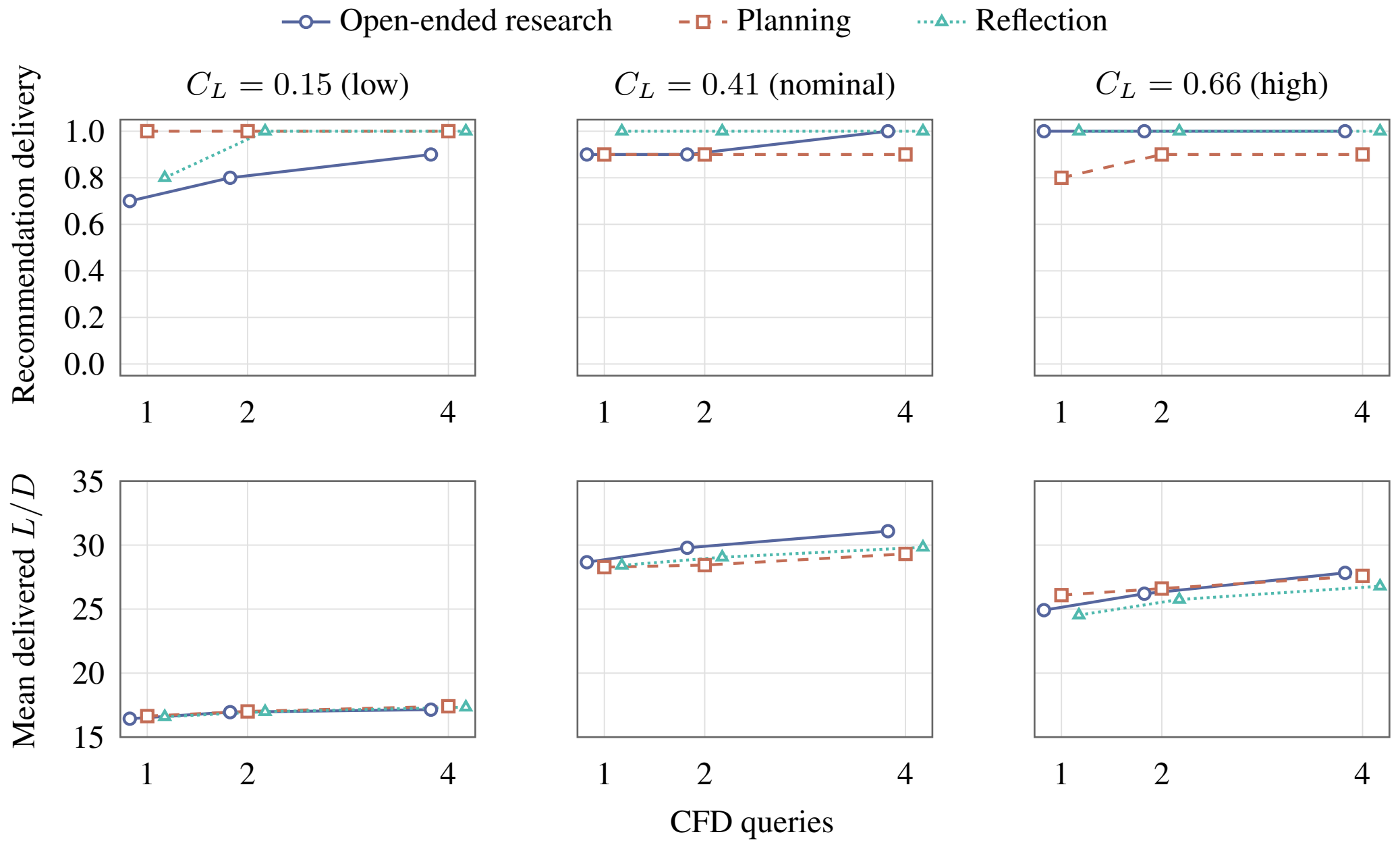

The early checkpoint allows all three agent methods to be compared.
Reflection delivers \ReflectionDeliveredAtFour{} of 30 recommendations,
open-ended research \OpenDeliveredAtFour{}, and planning
\PlanningDeliveredAtFour{} (Table~\ref{tab:agent-recommendations}).
Planning has the highest low-load conditional mean $L/D$, while open-ended
research leads at nominal and high load. No method leads across all tasks
and both outcomes (Figure~\ref{fig:agent-recommendations}).

\begin{table}[t]
\caption{Agent recommendations after four CFD queries. Mean $L/D$ includes only delivered recommendations,
so the quality columns can describe different subsets of episodes.}
\label{tab:agent-recommendations}
\centering
\scriptsize
\setlength{\tabcolsep}{3.2pt}
\begin{tabular*}{\textwidth}{@{\extracolsep{\fill}}lrrrrrr}
\toprule
method & \multicolumn{2}{c}{$C_L=0.15$} & \multicolumn{2}{c}{$C_L=0.41$} & \multicolumn{2}{c}{$C_L=0.66$} \\
 & delivered & mean $L/D$ & delivered & mean $L/D$ & delivered & mean $L/D$ \\
\midrule
Open-ended research & 9/10 & 17.15 & 10/10 & 31.09 & 10/10 & 27.83 \\
Planning & 10/10 & 17.41 & 9/10 & 29.31 & 9/10 & 27.59 \\
Reflection & 10/10 & 17.33 & 10/10 & 29.83 & 10/10 & 26.79 \\
\bottomrule
\end{tabular*}
\end{table}

The quality comparisons reinforce this task dependence. At nominal load, the
delivered recommendations of open-ended research exceed planning's by 1.77
$L/D$ with a descriptive 95\% bootstrap interval of [0.32, 3.21]. At low load,
the difference is $-0.26$ with interval [$-0.85$, 0.31], so planning's higher
mean is not distinguishable from chance. The intervals against reflection
include zero on all three tasks. Full comparisons are given in
Appendix~\ref{app:agent-comparison}. At query 4, mean charged surrogate usage
per episode was 913 for open-ended research, 704 for planning, and 828 for
reflection. Recorded session time per CFD query was 31, 25, and 31 minutes,
respectively, including tools and local code, not just model inference.
Appendix~\ref{app:resources} separates these measurements from the 32-query
extension, model charges, and CFD timing. In a descriptive cross-family
comparison at this checkpoint, open-ended research attains
higher nominal- and high-load quality than feedback-guided CMA-ES while
matching its complete delivery. Prior information and surrogate usage differ:
the adaptive baselines have used 512 surrogate evaluations at this point.
These are intermediate results within the
32-query task, not a test of strategies built to stop after four evaluations.

\subsection{Agent strategies and failure modes}

Across all 90 four-query episodes, 64 of the 270 adjacent query pairs start
after an unsuccessful evaluation. The next query succeeds in 19/27 open-ended,
10/20 planning and 15/17 reflection transitions. These descriptive counts
condition on different failures and do not rank causal recovery ability.
Appendix~\ref{app:agent-behavior} gives the complete audit and selected examples.

The baseline comparison motivates using feedback to change candidate
generation; the traces show executable changes of this kind. In nominal-load
A080, the agent runs CMA-ES, observes optimistic predictions and adds a
source-distance penalty; CFD improves from 33.15 to 33.40, then plateaus.
In low-load A076, reducing extreme sweep and adding a mesh-quality screen
after a mesh failure yields $L/D=16.04$, but the next, higher surrogate
prediction yields only 15.74. In high-load A006, a planning executor adds a
surface-triangle quality screen after two mesh failures; the next wing
scores 28.55 versus the initial 28.52. This recovers evaluability with little
quality gain. Its handoffs replace placeholder plans, so the recovery
demonstrates executor adaptation, not planning alone. Twenty of 120 planning
handoffs mention placeholders.

Additional modeling should be checked against subsequent CFD evidence.
In nominal-load reflection episode A054, a supplementary predictor becomes
about 12\% optimistic while AeroTransformer remains within 0.5\%.
In high-load A035, disagreement with a local ensemble helps defer a wing
family; the later tested candidate scores 24.89, close to AeroTransformer's
24.85 and above the ensemble's 22.91. Appendix~\ref{app:agent-behavior}
connects these predictions to the recorded decisions. At query 4, neither
the tested planning nor reflection implementation consistently improves both
recommendation delivery and conditional quality across tasks
(Figure~\ref{fig:agent-recommendations}). Extra reasoning should therefore
be assessed through the experiments it changes and the recommendations it
produces.

Together, the comparisons motivate evaluating reliability alongside quality,
using diagnostics to revise proposals, and testing supplementary predictions
against CFD. The traces establish which adaptations occurred and what
followed them; prospective ablations are needed to isolate their individual
benefits. Preserving a verified fallback and checking plan executability
remain additional hypotheses supported by the recorded failures.

\section{Conclusion}
\label{sec:conclusion}

WingBench evaluates design search with limited, failure-prone CFD and grades
the recommendation a method would return to a user. Among the tested numerical
methods, feedback improved delivery when it changed later proposals; reranking
a fixed archive did not produce that improvement. In the four-query agent
comparison, explicit reflection and planning did not consistently improve on
open-ended research under shared information, tools and query caps. The
open-ended extension found high-quality designs over 32 queries, but different
prior information, surrogate usage and unmeasured computation prevent a
resource-matched ranking against the baselines. Best-found designs and standing
recommendations also give different outcomes. We release the 6{,}629 graded
records, resource-accounting tables, rules, runner, verifier, prompts and
analysis. The original distributed agent controller is not included. Matching
initial information, compute limits and endpoints remains necessary to measure
cross-family rankings under equal resources.

\subsection*{AI use statement}

Generative AI tools assisted with writing and editing the manuscript, with
implementing and debugging code for the analysis, the experiment pipeline, and
the agent harness, and with organizing the CFD records. The authors conceived
the research, designed the benchmark and the experiments, oversaw their
execution, checked every reported number against the accepted CFD records, and
take responsibility for all analyses, claims, and text.

\subsection*{Reproducibility statement}

Sections~\ref{sec:benchmark}, \ref{sec:methods}, and \ref{sec:analysis} describe
the benchmark, optimization methods, experimental budgets, and statistical
analysis. Appendix~\ref{app:agent-comparison}
gives the agent comparisons, while
Appendices~\ref{app:classifier}, \ref{app:weight}, and \ref{app:outcomes}
provide the failure-classifier analysis, sensitivity checks, and CFD outcomes.
The accompanying materials, a public repository at
\url{https://github.com/davidxia3/wingbench}, support three levels of
reproduction. First, every
result table can be recomputed from the accompanying records with analysis
scripts that need no CFD. The revision adds portable extracts for resource
usage, evaluator checks, decision handoffs and bounded recoveries; qualitative
case interpretations remain distinguished from recorded outcomes. Second, every evaluated wing but five is released with its 38 coordinates, and
every evaluation with a result document, 6{,}241 of the 6{,}629, with its per-condition lift, drag,
angle of attack, and convergence values, so a design can be regraded, and
the CFD evaluator can be rebuilt from a container recipe, a fetch script for the
SuperWing simulation files, and the public surrogate weights, all pinned by
fingerprint. Third, the benchmark folder holds a runner that lets a new
optimizer play episodes under the released episode machine, a verifier that
replays its trace, and a scorer that compares it with the six reference methods.
The runner was tested with recorded outcomes, and its CFD path was run
once in a rebuilt environment on a fresh machine of the reference type, where
it reproduced a recorded evaluation exactly, to the last digit of lift, drag,
and angle of attack at every condition (Appendix~\ref{app:outcomes}). The
SuperWing run records are redistributed with the authors' permission. The code
that ran jobs on our cloud and cluster machines, the agent session controller,
and the full agent traces are not included. The reference hardware requirement
does not establish a common machine type for the historical study.


\clearpage
\appendix
\raggedbottom
\input{app_limitations.tex}
\input{app_resources.tex}
\input{app_evaluator_checks.tex}

\input{app_agent_behavior.tex}

\section{Agent comparisons}
\label{app:agent-comparison}

All agents used DeepSeek V4.1 Flash on a dedicated deployment, through an
OpenAI-compatible chat interface with reasoning effort at its maximum. The
requests set no temperature or other sampling option, so the provider's
defaults applied. Each model response could use up to 262{,}144 output tokens
within the model's context of 1{,}048{,}576 tokens, and each turn could make
up to 64 rounds of tool calls. The agent episodes ran in September 2026. For
the agents, a seed is a label. The same ten labels identify episodes within each method;
they do not set model sampling or create paired random draws across methods.

The planning value rule is fixed. For a candidate first action, the evaluator
predicts the probability $p$ that the evaluation completes and, given
completion, the 10th, 50th, and 90th percentiles of its $L/D$. The rule turns
this into four scenarios, incomplete with weight $1 - p$ and complete with
weights $0.3p$, $0.4p$, and $0.3p$ at the three percentiles, the Pearson--Tukey
three-point approximation. With $\beta_t$ the best complete $L/D$
measured after step $t$ of the plan, zero if none, and $b_0$ that best before
the plan, the value of a two-step plan is $\mathbb{E}[(\beta_1 + \beta_2)/2]$
over the scenario tree, where the second step's predictions are conditional on
whether the first step was incomplete, complete without improving on $b_0$, or
improved on it. Plans are ranked by this value, then by the first step's
completion probability.

The agent recommendation data contain 270 assessments at queries 1, 2, and 4,
across three methods, three tasks, and ten seeds. Each assessment identifies its
episode, checkpoint, geometry, delivery outcome, and conditional $L/D$.
Table~\ref{tab:agent-contrasts} compares open-ended research with each alternative
at query 4. Because a seed label does not pair agent episodes, the two methods'
ten episodes are resampled independently. A quality difference is the difference
between the two conditional means in Table~\ref{tab:agent-recommendations}, and
its interval comes from the delivered recommendations of each resample. Delivery
differences carry Newcombe score intervals, because a bootstrap of two arms that
both deliver every recommendation has no spread. The intervals describe uncertainty in these individual comparisons
and do not account for multiple comparisons or the choice of which agent to
continue.

The data also contain 90 recommendations made before the first online query.
Before any CFD feedback, 64 of the 90 delivered, including 18 of 30 for
open-ended research and 23 of 30 for both reflection and planning. The four
surrogate-assisted baselines delivered 7 to 13 of 30 at this checkpoint. Among
the delivered agent recommendations, mean $L/D$ is 16.5 to 16.8 at low load,
28.2 to 29.1 at nominal load, and 24.4 to 26.1 at high load, below every value
at query 4 in Table~\ref{tab:agent-recommendations}. At this point, agents
have only the surrogate and the SuperWing records to guide them.

The 32-query continuation asked for no further checkpoint recommendations, but
agents could record one at any time, and all 30 open-ended episodes did.
Taking the last recommendation recorded by 32 results as the standing
recommendation, 28 of 30 name a wing that delivered
online, 2 name a wing that failed, and 20 name the best delivered wing. The
delivered ones average 18.21, 33.03, and 30.43 $L/D$ at low, nominal, and high
load, within 0.2 of the best online values in Table~\ref{tab:agent-extension}.
At 8 and 16 results, 19 and 14 of the standing recommendations were untested
wings with unknown delivery.

Ten of the extension's 960 online evaluations, in five episodes and covering
five distinct wings, ended before CFD began, with the grader's exit code and a
reason code saying that the grading job exited without a result. Nine fell within
twelve minutes on one day, and one agent resubmitted the same wing five times
into that outage. They were recorded as inconclusive outcomes and consumed the
agent's queries, which the released rules do not allow. Excluding them raises
online delivery in Table~\ref{tab:agent-extension} from 0.62 to 0.63 at low
load and from 0.70 to 0.71 at nominal load, leaves 0.62 at high load, and
changes no episode maximum. We regraded the five wings after the study on
fresh machines of the reference type. The high-load wing was rejected as
unmeshable in three minutes, so that outcome was the wing's. Three of the other four delivered, at 18.39, 30.59, and 33.97, and the fourth was valid at two conditions and hit the two-hour limit at the third. The low-load wing of episode A070 exceeded its episode's best online score of 18.35. The linked repository's regrade table lists them.

The whole agent study reports 71.5 million input tokens, 29.1 million output
tokens, and 6.2 billion cache-read tokens, with reference model charges of about
\$118. These totals exclude CFD and hosting; eleven sessions lack token counts.
Appendix~\ref{app:resources} separates the common four-query comparison from
the open-ended extension and reports session elapsed time, which includes
tools and local code rather than measuring pure model inference.
The released tables list every session and every recorded recommendation.

\begin{table}[h]
\caption{Open-ended research minus each comparator after four CFD queries,
with the two methods' episodes resampled independently. Quality differences
compare delivered recommendations, counted in the delivered column for each
method. Positive differences favor open-ended research. Quality intervals use
10{,}000 bootstrap resamples and delivery intervals are Newcombe's.}
\label{tab:agent-contrasts}
\centering
\scriptsize
\setlength{\tabcolsep}{3pt}
\begin{tabular*}{\textwidth}{@{\extracolsep{\fill}}llrrrrr}
\toprule
comparator & $C_L$ & $\Delta$ delivery & 95\% CI & delivered & $\Delta L/D$ & 95\% CI \\
\midrule
Planning & $0.15$ & -0.10 & [-0.40, 0.19] & 9/10 & -0.26 & [-0.85, 0.31] \\
Planning & $0.41$ & 0.10 & [-0.19, 0.40] & 10/9 & 1.77 & [0.32, 3.21] \\
Planning & $0.66$ & 0.10 & [-0.19, 0.40] & 10/9 & 0.24 & [-1.76, 2.15] \\
Reflection & $0.15$ & -0.10 & [-0.40, 0.19] & 9/10 & -0.19 & [-0.67, 0.31] \\
Reflection & $0.41$ & 0.00 & [-0.28, 0.28] & 10/10 & 1.26 & [-0.03, 2.55] \\
Reflection & $0.66$ & 0.00 & [-0.28, 0.28] & 10/10 & 1.04 & [-0.33, 2.38] \\
\bottomrule
\end{tabular*}
\end{table}

Table~\ref{tab:marginal} gives each reported mean its own interval, an exact
binomial interval for the delivery rate and a bootstrap interval for the
conditional mean from 10{,}000 resamples of the ten episodes of each method and task.
At nominal load, the delivered mean of open-ended research, 31.09, has interval
[30.04, 32.11], and planning's 29.31 has [28.28, 30.30]. Among the baselines,
the archive-only methods' delivered means rest on four or five episodes, and
SMAC3's rest on scores that vary widely.

\begin{table}[h]
\caption{95\% intervals for the recommendation delivery rate (exact binomial)
and for the mean $L/D$ of delivered recommendations (percentile bootstrap from
10{,}000 resamples of each method's ten episodes per task). Agents at query 4,
baselines at query 32. A dash marks a cell with no delivery.}
\label{tab:marginal}
\centering
\scriptsize
\setlength{\tabcolsep}{4pt}
\begin{tabular*}{\textwidth}{@{\extracolsep{\fill}}llrlrl}
\toprule
method & $C_L$ & delivered & 95\% CI & mean $L/D$ & 95\% CI \\
\midrule
\multicolumn{6}{l}{\emph{Agents after four queries}} \\
Open-ended research & $0.15$ & 9/10 & [0.56, 1.00] & 17.15 & [16.73, 17.56] \\
 & $0.41$ & 10/10 & [0.69, 1.00] & 31.09 & [30.04, 32.11] \\
 & $0.66$ & 10/10 & [0.69, 1.00] & 27.83 & [26.58, 29.02] \\
Planning & $0.15$ & 10/10 & [0.69, 1.00] & 17.41 & [16.99, 17.80] \\
 & $0.41$ & 9/10 & [0.56, 1.00] & 29.31 & [28.28, 30.30] \\
 & $0.66$ & 9/10 & [0.56, 1.00] & 27.59 & [26.03, 29.20] \\
Reflection & $0.15$ & 10/10 & [0.69, 1.00] & 17.33 & [17.09, 17.58] \\
 & $0.41$ & 10/10 & [0.69, 1.00] & 29.83 & [29.05, 30.58] \\
 & $0.66$ & 10/10 & [0.69, 1.00] & 26.79 & [26.15, 27.41] \\
\midrule
\multicolumn{6}{l}{\emph{Baselines after 32 queries}} \\
CMA-ES, no feedback & $0.15$ & 0/10 & [0.00, 0.31] & -- & -- \\
 & $0.41$ & 4/10 & [0.12, 0.74] & 33.22 & [32.85, 33.63] \\
 & $0.66$ & 5/10 & [0.19, 0.81] & 32.12 & [32.00, 32.23] \\
CMA-ES, selection only & $0.15$ & 0/10 & [0.00, 0.31] & -- & -- \\
 & $0.41$ & 4/10 & [0.12, 0.74] & 33.22 & [32.87, 33.63] \\
 & $0.66$ & 5/10 & [0.19, 0.81] & 32.12 & [31.99, 32.23] \\
Feedback-guided CMA-ES & $0.15$ & 6/10 & [0.26, 0.88] & 16.76 & [15.60, 18.01] \\
 & $0.41$ & 10/10 & [0.69, 1.00] & 33.04 & [32.68, 33.39] \\
 & $0.66$ & 10/10 & [0.69, 1.00] & 31.64 & [31.25, 31.98] \\
Feedback-guided DE & $0.15$ & 10/10 & [0.69, 1.00] & 17.52 & [17.04, 17.97] \\
 & $0.41$ & 10/10 & [0.69, 1.00] & 31.09 & [30.61, 31.53] \\
 & $0.66$ & 10/10 & [0.69, 1.00] & 27.05 & [25.83, 28.04] \\
SMAC3 & $0.15$ & 9/10 & [0.56, 1.00] & 10.56 & [7.52, 13.63] \\
 & $0.41$ & 7/10 & [0.35, 0.93] & 17.00 & [13.77, 20.12] \\
 & $0.66$ & 6/10 & [0.26, 0.88] & 12.82 & [8.74, 16.93] \\
\bottomrule
\end{tabular*}
\end{table}

\section{Offline failure-classifier details}
\label{app:classifier}

The source records provide outcomes for 4{,}544 designs, including the 4{,}239
released geometries and attempted designs with no valid condition. We
standardized the 38 variables with the benchmark normalization and predicted
whether a design fails any SuperWing condition. Logistic regression,
15-nearest-neighbors, and histogram gradient boosting reach five-fold AUC
0.69, 0.65, and 0.71. Training on the central 85\% and testing on
the extreme 15\% reduces these to 0.62, 0.52, and 0.63.

We then applied the source-trained classifiers to the 3{,}840 designs proposed by
the four surrogate-assisted baseline methods and labeled them with their actual
WingBench outcomes. AUC is 0.41, 0.52, and 0.50 at low load and 0.51--0.75 at
nominal and high load. These values are per task because online
nondelivery differs sharply across tasks. The experiment does not prove that
failure prediction is impossible. It shows that the tested source models do not
remove the need for online learning.

\section{Sensitivity to the inconclusive-outcome weight}
\label{app:weight}

The two feedback-guided baseline methods give inconclusive CFD outcomes weight
0.5 in their delivery model. Without changing any completed search, we refitted the
belief for the 103 checkpoint recommendations that could depend on this weight.
Using weight 1 changes 21 recommendations, and using weight 0 changes 33.
At query 32, only four of 60 episodes lack a delivered online design and so
use this merit-based path, and two
recommendations change between weights 0.5 and 1. Both alternative designs and
both designs they replace fail at the mesh stage. Consequently no reported
final endpoint changes. This is a recommendation sensitivity check, not a full
rerun, since changing the weight during search could change the proposal trajectory.

\section{CFD outcomes}
\label{app:outcomes}

The two studies contain \UnifiedCFDObserved{} graded CFD requests. Of these,
2{,}620 returned a score and 4{,}009 did not. Table~\ref{tab:all-outcomes}
accounts for the online evaluations and private recommendation evaluations in
both studies. These are counts of graded requests, including accepted results
reused for repeated designs, rather than counts of distinct solver executions.
They cover at least 5{,}205 distinct wings. The timing aggregates cover
\TimedEvaluationsOverall{} records and sum to \CoreHoursOverall{}
processor-core hours. This is timing attached to records, not unique compute
consumed: it includes reused outcomes and two superseded grades
(Appendix~\ref{app:resources}).

\begin{table}[htbp]
\caption{CFD score delivery across both studies. Private evaluations score
recommendations separately from the online search when a prior online grade
cannot be reused.}
\label{tab:all-outcomes}
\centering
\scriptsize
\setlength{\tabcolsep}{5pt}
\begin{tabular*}{\textwidth}{@{\extracolsep{\fill}}llrrr}
\toprule
study & evaluation & delivered a score & no score & total \\
\midrule
Five baseline optimizers & online & 1{,}648 & 3{,}152 & 4{,}800 \\
 & private & 62 & 431 & 493 \\
Three agent methods & online & 806 & 394 & 1{,}200 \\
 & private & 104 & 32 & 136 \\
\midrule
Both studies & all & 2{,}620 & 4{,}009 & 6{,}629 \\
\bottomrule
\end{tabular*}
\end{table}
\FloatBarrier

The baseline study contributes \TotalCFDEvaluations{} graded requests,
including 3{,}583 nondeliveries. Table~\ref{tab:outcomes} assigns each
baseline outcome to its first failing stage. Of 954 evaluations that build a
mesh and attempt all three conditions but do not deliver, 375 produce a valid
result at two of three conditions. This group differs slightly from the 955
time-limit and solver-error outcomes in Table~\ref{tab:outcomes}. It adds five
other errors that attempted all three conditions and leaves out six outcomes
that have no condition records. A nondelivery is therefore often a near miss,
not a geometry from which the pipeline obtains nothing.

\begin{table}[htbp]
\caption{CFD outcomes in the five-method baseline study. Private counts include
only recommendations requiring a separate grade. Counts are evaluations, not
distinct wings.}
\label{tab:outcomes}
\centering
\scriptsize
\setlength{\tabcolsep}{5pt}
\begin{tabular*}{\textwidth}{@{\extracolsep{\fill}}lrrr}
\toprule
outcome & online & checkpoint & total \\
\midrule
delivered a score & 1{,}648 & 62 & 1{,}710 \\
\midrule
mesh generator rejected geometry & 438 & 65 & 503 \\
mesh generator stopped without a mesh & 1{,}825 & 282 & 2{,}107 \\
flow condition reached time limit & 552 & 47 & 599 \\
solver process stopped with an error & 319 & 37 & 356 \\
other errors & 18 & 0 & 18 \\
all nondeliveries & 3{,}152 & 431 & 3{,}583 \\
\midrule
evaluations & 4{,}800 & 493 & 5{,}293 \\
\bottomrule
\end{tabular*}
\end{table}

The agent study contributes 1{,}336 graded requests. Its 136 private grades
include 90 before the first online query and 46 at later recommendation
checkpoints. Another 224 recommendation assessments reused online grades and
add no new request to this count. Table~\ref{tab:agent-outcomes} reports the
recorded agent outcome types for both online and private grades. A partial
outcome has usable results for only some of the required flow conditions and
does not receive a complete wing score.

Both tables come from the same grader. Table~\ref{tab:agent-outcomes} uses the
four outcome types that the feedback-guided baselines weight in their delivery
model, and agents receive each result together with the files the evaluation
produced. Table~\ref{tab:outcomes} groups the grader's reason codes by
the first failing stage. In baseline records with both labels, every mesh
rejection is a conclusive failure and every mesh stop undetermined.
Of the 955 time-limit and solver-error outcomes, the 641 with a type label are partial in 389 cases and undetermined in 252. The other 314 are SMAC3 records, which carry no type label.

\begin{table}[htbp]
\caption{Recorded CFD outcome types in the three-method agent study. A pass
delivers a complete score. The other outcome types do not. Private counts are
the 136 separate grades.}
\label{tab:agent-outcomes}
\centering
\scriptsize
\setlength{\tabcolsep}{5pt}
\begin{tabular*}{\textwidth}{@{\extracolsep{\fill}}lrrr}
\toprule
outcome & online & private & total \\
\midrule
pass & 806 & 104 & 910 \\
partial & 212 & 6 & 218 \\
undetermined & 137 & 17 & 154 \\
conclusive failure & 45 & 9 & 54 \\
\midrule
evaluations & 1{,}200 & 136 & 1{,}336 \\
\bottomrule
\end{tabular*}
\end{table}

Mesh rejection in the baseline study is a geometry failure under the chosen
meshing procedure. Stopping without a mesh is recorded as inconclusive because
the flow solver never tests the wing. Most of these stops are crashes, not time
limits. Of the 2{,}100 mesh stops with a recorded attempt, 1{,}582 ended in a
process error after a median of 125 seconds, 1{,}526 of them with a
segmentation fault, and 518 reached the 30-minute limit. A flow timeout consumes
the full condition budget without producing a valid result. Outcomes set by a
time limit, these 518 mesh stops and the 599 flow timeouts, make up 31\% of
baseline nondeliveries and could change on faster hardware. If every
clock-limited nondelivery were instead counted as unknown, baseline online
delivery would lie between 0.34 and 0.54 overall, between 0.53 and 0.66 for
feedback-guided CMA-ES, 0.51 and 0.67 for feedback-guided DE, 0.21 and 0.43
for no-feedback CMA-ES, 0.21 and 0.41 for selection-only CMA-ES, and 0.26 and
0.55 for SMAC3. Mesh stops are
concentrated at low load (1{,}212 of 2{,}107) and solver-process errors at high
load (276 of 356). The records also show how repeatable outcomes are. 809
task and wing pairs were evaluated more than once in separate runs, with
different mesh times. Delivery agreed in 800 of them (98.9\%). All nine
disagreements were high-load time limits or solver-process errors.
The raw release preserves condition-level evidence and keeps infrastructure
failures separate from these scientific outcomes.

The rebuilt environment was tested once. On 2026-09-25 we built the CFD
evaluator from the released container recipe on a new machine of the reference
type. The container built from the recipe, the fetch script matched all four
pinned SuperWing files, and the local grading adapter regraded a recorded
nominal-load wing. It reproduced the recorded score of 33.4018 exactly, with
lift, drag, and angle of attack identical to the last digit at all three
conditions, in 7{,}973 seconds. Only the runtime fingerprint differed, as
documented, because it binds the original machines. The
regrades of the five outage wings in Appendix~\ref{app:agent-comparison} used
the same rebuilt evaluator on machines of the same type.

\input{app_formal.tex}
\input{app_rules.tex}
\input{app_data_cfd.tex}
\input{app_gallery.tex}
\input{app_results.tex}
\input{app_figures.tex}
\input{app_prompts.tex}
\clearpage
\input{app_resource_followup.tex}

\end{document}

%% file: math_commands.tex
\usepackage{amsmath,amsfonts,bm}

\def\eqref#1{equation~\ref{#1}}

\def\1{\bm{1}}

\DeclareMathAlphabet{\mathsfit}{\encodingdefault}{\sfdefault}{m}{sl}
\SetMathAlphabet{\mathsfit}{bold}{\encodingdefault}{\sfdefault}{bx}{n}



%% file: smac3_numbers.tex
\newcommand{\TotalCFDEvaluations}{5{,}293}

\newcommand{\SMACLowOnlineRate}{0.36}
\newcommand{\SMACNomOnlineRate}{0.28}
\newcommand{\SMACHighOnlineRate}{0.15}

\newcommand{\SMACLowFinalMean}{10.56}

\newcommand{\SMACLowFinalDelivered}{9}

\newcommand{\SMACNominalFinalMean}{17.00}

\newcommand{\SMACNominalFinalDelivered}{7}

\newcommand{\SMACHighFinalMean}{12.82}

\newcommand{\SMACHighFinalDelivered}{6}


%% file: app_limitations.tex
\section{Limitations}
\label{app:limitations}

WingBench focuses on transonic wing design under a fixed simulation protocol
and computational budget. Ten seeds per method and task provide limited power
to detect small performance differences. Delivery rates also depend on the
computing environment: time limits account for 31\% of baseline nondeliveries,
with their impact bounded in Appendix~\ref{app:outcomes}. Planning results
capture the combined contributions of the selected plans and the executor's
reasoning. Future work could extend the benchmark to other design domains,
increase the number of independent runs, and evaluate additional search
strategies across different computational budgets. The present study does not
include an adjoint comparison and does not establish competitiveness with
aerodynamic optimization practice.

The evaluator records success under a particular mesh recipe, solver and
execution limits. Upstream consistency and source-geometry checks do not
establish mesh independence or universal feasibility. The layer sensitivity
in Appendix~\ref{app:evaluator-checks} shows that some recorded mesh failures
are recoverable under different numerical settings. Agent transition counts
and selected cases motivate hypotheses, but do not isolate prompt effects;
bounded recoveries and executable-plan substitutions also affect the tested
implementations (Appendix~\ref{app:agent-behavior}).

The cross-family comparison uses different prior CFD information and surrogate
access. Equal online query counts do not imply equal resources: surrogate
usage differs at intermediate checkpoints, and model calls and elapsed session
time differ among the agents. Optimizer CPU/GPU time, model-server compute and
consistent end-to-end wall time were not recorded across all methods
(Appendix~\ref{app:resources}). The open-ended extension was selected after the
four-query comparison and its best-found design need not be its recommendation.
The records therefore support performance under the stated configurations,
not a ranking under fully matched information and compute budgets.

%% file: app_resources.tex
\section{Resource conditions and recorded usage}
\label{app:resources}
The following tables account for recorded usage; they do not estimate rankings under equal resources. Each method has 30 episodes (ten per task). The query-4 rows describe prefixes of a 32-query task, not runs instructed to stop at four. The open-ended query-32 row continues its query-4 episodes and must not be added to that row.
\subsection{Surrogate and CFD requests}
Table~\ref{tab:resource-queries} separates prior information, online requests, private recommendation grades, and actual surrogate usage. The four surrogate-assisted baselines receive the same 62 development records; agents receive SuperWing source records but not this 62-record context. SMAC3 has neither the warm start nor surrogate access. Private grades are not returned to the method. They are additional recorded requests, not necessarily new solver executions.

\begin{table}[htbp]\centering\scriptsize
\caption{Recorded requests through each checkpoint. Surrogate entries are mean [minimum, maximum] charged candidate evaluations per episode, with a 4,096 episode cap except for SMAC3. CFD columns are totals over 30 episodes.}
\label{tab:resource-queries}
\setlength{\tabcolsep}{3pt}
\begin{tabular*}{\textwidth}{@{\extracolsep{\fill}}lrrrr}\toprule
Method & Checkpoint & Surrogate / episode & Online CFD & Private CFD \\ \midrule
CMA-ES, no feedback & 4 & 4{,}096.0 [4{,}096, 4{,}096] & 120 & 99 \\
CMA-ES, selection only & 4 & 4{,}096.0 [4{,}096, 4{,}096] & 120 & 96 \\
Feedback-guided CMA-ES & 4 & 512.0 [512, 512] & 120 & 75 \\
Feedback-guided DE & 4 & 512.0 [512, 512] & 120 & 73 \\
SMAC3 & 4 & 0.0 [0, 0] & 120 & 0 \\
Open-ended research & 4 & 912.8 [108, 2{,}486] & 120 & 47 \\
Planning & 4 & 704.1 [57, 1{,}562] & 120 & 42 \\
Reflection & 4 & 828.3 [148, 2{,}009] & 120 & 47 \\
\midrule
CMA-ES, no feedback & 32 & 4{,}096.0 [4{,}096, 4{,}096] & 960 & 163 \\
CMA-ES, selection only & 32 & 4{,}096.0 [4{,}096, 4{,}096] & 960 & 159 \\
Feedback-guided CMA-ES & 32 & 4{,}096.0 [4{,}096, 4{,}096] & 960 & 89 \\
Feedback-guided DE & 32 & 4{,}096.0 [4{,}096, 4{,}096] & 960 & 74 \\
SMAC3 & 32 & 0.0 [0, 0] & 960 & 8 \\
Open-ended research & 32 & 1{,}988.9 [116, 4{,}091] & 960 & 47 \\
\bottomrule\end{tabular*}\end{table}

The archive methods spend all 4,096 surrogate evaluations before online CFD; the adaptive baseline methods have spent 512 at query 4 and 4,096 at query 32. Agents choose their own usage. Agent counts sum committed ledger charges; exact repeats served from the episode cache are free and are excluded. Thus charged evaluations are not the number of tool invocations. Baseline counts come from checkpoint budget counters; SMAC3's zero follows from its CFD-only implementation. These accounting rules and the different information sets remain limitations of cross-family comparisons.
\subsection{Agent sessions and model usage}
Table~\ref{tab:resource-model} counts all launched sessions assigned to decisions through the checkpoint, including checkpoint-0 reviews, later reviews, reflection, errors and retries. Planning's nested model calls are included in its research sessions. Summed session elapsed time includes tool execution, local code, and waiting within sessions; it is neither pure model inference time nor end-to-end campaign wall time, and overlapping sessions need not be additive elapsed time. Token counts are provider-reported and incomplete for eleven sessions across the whole study. Cache-read tokens are reported separately. Dollar values are reference model charges, excluding CFD and hosting.

\begin{table}[htbp]\centering\scriptsize
\caption{Recorded agent usage at a common four-query checkpoint and for the open-ended extension. Tokens are totals in millions across 30 episodes. Time is summed session minutes divided by online CFD requests. Missing-token sessions still contribute their recorded time and cost.}
\label{tab:resource-model}
\setlength{\tabcolsep}{2pt}
\begin{tabular*}{\textwidth}{@{\extracolsep{\fill}}lrrrrrrrr}\toprule
Method & $q$ & Input & Output & Cache & \shortstack{Model\\calls} & \shortstack{Minutes\\/ CFD} & \shortstack{Cost\\(\$)} & \shortstack{Missing\\sessions} \\ \midrule
Open-ended research & 4 & 3.72 & 5.64 & 918.94 & 5{,}946 & 30.97 & 17.07 & 0 \\
Planning & 4 & 5.47 & 6.17 & 899.16 & 7{,}096 & 25.01 & 18.12 & 5 \\
Reflection & 4 & 15.00 & 10.18 & 1{,}112.29 & 8{,}824 & 31.40 & 27.93 & 3 \\
Open-ended research & 32 & 51.01 & 12.79 & 4{,}150.75 & 13{,}188 & 8.60 & 72.31 & 3 \\
\bottomrule\end{tabular*}\end{table}

\subsection{CFD timing and unmeasured resources}
Table~\ref{tab:resource-cfd-time} uses the existing episode timing aggregates: summed mesh and flow-attempt wall seconds multiplied by 16 cores. These include online and private grading. They describe timing attached to evaluation records, not unique physical compute or a cloud bill: outcomes can be reused, some records have no timing, and the SMAC3 timing aggregate retains two superseded grades. The aggregates do not provide a separate query-4 timing total for the open-ended extension or the baselines. They also do not establish identical hardware across campaigns.

\begin{table}[htbp]\centering\scriptsize
\caption{CFD timing at each method's recorded final horizon. These rows cover different query horizons. No matched-runtime comparison is inferred.}
\label{tab:resource-cfd-time}
\setlength{\tabcolsep}{4pt}
\begin{tabular*}{\textwidth}{@{\extracolsep{\fill}}lrrr}\toprule
Method & Final query & Timed records & Recorded core-hours \\ \midrule
CMA-ES, no feedback & 32 & 1{,}123 & 16{,}689.1 \\
CMA-ES, selection only & 32 & 1{,}119 & 16{,}583.6 \\
Feedback-guided CMA-ES & 32 & 1{,}049 & 38{,}759.4 \\
Feedback-guided DE & 32 & 1{,}034 & 38{,}063.9 \\
SMAC3 & 32 & 954 & 35{,}306.8 \\
Open-ended research & 32 & 968 & 42{,}112.9 \\
Planning & 4 & 162 & 6{,}112.2 \\
Reflection & 4 & 167 & 6{,}391.6 \\
\bottomrule\end{tabular*}\end{table}

Optimizer CPU/GPU time, model-server compute, and consistent end-to-end episode wall time were not recorded across all methods. Missing values are not treated as zero. Consequently, neither equal CFD caps nor this accounting supplies a ranking under equal total computation or wall-clock limits. Such a ranking would require a new experiment with a common initial information set, resource rules and recommendation endpoint; it cannot be reconstructed by subtracting prior records or truncating costs after the fact.
\paragraph{Reproduction.} \texttt{analysis/20\_resource\_accounting.py} reads the released episode, evaluation and agent-session tables plus \texttt{data/surrogate\_usage.csv}, and writes per-episode and method-level tables and this appendix. The surrogate extract includes checkpoint counts for all 240 episodes and source fingerprints in \texttt{data/surrogate\_usage\_provenance.json}. The optional archive extractor is supplied separately; the original campaign archives are not required to reproduce the accounting tables.

%% file: app_evaluator_checks.tex
\section{Additional evaluator checks}
\label{app:evaluator-checks}

These checks distinguish failure to obtain a benchmark score from numerical
nonconvergence, and implementation consistency from robustness of a mesh
recipe. They do not change any original grade or optimizer result.

\subsection{Condition-level outcomes after an accepted mesh}

We audited 3{,}845 evaluation requests with accepted meshes and recoverable,
hash-verified raw result documents. Each contains three condition attempts.
This is a request-level audit: repeated requests and reused results remain
included, and the subset is not a new sample of independent solver runs.
It is separate from the aggregate classification of all 6{,}629 requests.

\begin{table}[htbp]
\centering\small
\caption{Failures after mesh acceptance in the audited subset. The final row
counts evaluations with at least one failed condition, not condition attempts.}
\label{tab:flow-audit}
\begin{tabular}{lrrr}
\toprule
Unit & Failed & Total & Rate \\
\midrule
Mach 0.7853 & 742 & 3{,}845 & 19.3\% \\
Mach 0.8195 & 765 & 3{,}845 & 19.9\% \\
Mach 0.8530 & 896 & 3{,}845 & 23.3\% \\
All condition attempts & 2{,}403 & 11{,}535 & 20.8\% \\
Three-condition evaluations & 1{,}226 & 3{,}845 & 31.9\% \\
\bottomrule
\end{tabular}
\end{table}

The 2{,}403 failed condition attempts comprise 1{,}222 timeouts and 1{,}181
execution/infrastructure outcomes. They cannot be described as numerical
nonconvergence alone. Requiring all three conditions to succeed raises the
observed evaluation failure rate to 31.9\%; this is measured directly, without
assuming independence between conditions. The condition rates remain material
and do not establish adequacy of the solver settings or time limits.

\subsection{Source-geometry controls and upstream meshing}

We remeshed 1{,}000 distinct previously successful SuperWing geometries with
the benchmark spacing and the original source spacing. The two cohorts of
500 were separately frozen random selections from eligible source geometries;
the second was selected after the first cohort's results. We therefore give
descriptive combined counts, without a pooled confirmatory interval. We also
tested all 1{,}913 distinct failed geometries recoverable from the benchmark
records with the original, unmodified meshing script released with SuperWing
in the AeroTransformer repository. The script is byte-for-byte identical to
the published version pinned in Appendix~\ref{app:cfdprotocol}, and was run
directly with its original tip-template path, without the WingBench case
preparer or grading orchestration, under matched inputs, MPI ranks, runtime,
and timeout settings.

\begin{table}[htbp]
\centering\small
\caption{Mesh acceptance in source controls and the complete recoverable
failed-geometry census. Acceptance requires
positive minimum cell volume and quality; marginal positive quality is accepted.
These are mesh checks, not completed CFD evaluations.}
\label{tab:mesh-controls}
\begin{tabular}{llrr}
\toprule
Cohort & Mesh setting & Accepted & Tested \\
\midrule
Successful source geometries & Benchmark spacing & 999 & 1{,}000 \\
Same source geometries & Original source spacing & 1{,}000 & 1{,}000 \\
Previously failed geometries & Upstream default (81 layers) & 1 & 1{,}913 \\
Same failed geometries & Diagnostic 129 layers & 605 & 1{,}913 \\
\bottomrule
\end{tabular}
\end{table}

The failed-geometry census is complete: all 30 prespecified batches from the
randomized ordering of 1{,}913 distinct geometries finished by 9 November 2026.
It includes every recoverable geometry identified as mesh-blocked in either
the released request categories or the raw evaluation records. One additional
mesh-failure request has no archived geometry and could not be rerun.
Under the unchanged upstream default, 408 were rejected by mesh gates,
1{,}504 stopped inconclusively, and one produced an accepted mesh.
Thus 1{,}912/1{,}913 remained unsuccessful.
This supports consistency with the published procedure; it does not establish
that the procedure is robust throughout the search box or that failed shapes
are physically infeasible. Rebuilt environments also do not establish exact
historical binary and hardware equivalence.

The paired 129-layer diagnostic is an important qualification: it accepted
605 meshes, including the one accepted by the default, so 604 default failures
became mesh successes. Failures are therefore sensitive to numerical mesh
settings. These meshes have not been used to replace benchmark grades, and
mesh acceptance alone says nothing about subsequent convergence or $L/D$.
Improving the recipe requires separate flow validation and a new protocol
version. The current evidence narrows an implementation-mismatch explanation
but does not establish a universal ground-truth feasibility evaluator.

The portable tables \texttt{accepted\_mesh\_flow\_audit.csv} and
\texttt{mesh\_validation\_checks.csv}, with source fingerprints in
\texttt{revision\_evidence\_provenance.json}, preserve the counting units and
cohort membership. \texttt{analysis/21\_revision\_evidence.py} reproduces the
tables above without new CFD runs.

\subsection{Surface appearance and automatic volume meshing}
\label{app:automatic-meshing}

A wing surface and a valid surrounding volume mesh are different objects.
The published SuperWing script constructs surface blocks and uses pyHyp to
march them toward the far field under fixed settings. Local curvature can
make marching lines approach or cross; pyHyp documents smoothing controls
for preventing such crossings in concave regions \citep{mdolabpyhyp}.
This is a numerical limitation that can arise with a correctly implemented
meshing method. A different topology, tuned settings or manual mesh construction
could succeed on a surface for which the configured automatic recipe fails.
Such alternatives must be validated and specified as a separate protocol;
the benchmark does not perform unrecorded, wing-specific manual repairs.

%% file: app_agent_behavior.tex
\section{Agent behavior: complete transition audit and selected examples}
\label{app:agent-behavior}

We separate a descriptive audit of every common-horizon episode from selected
examples of decision-making. The evidence is the 360 recorded handoffs in all
90 episodes through query 4, joined by geometry hash to their CFD grades.
\texttt{data/agent\_handoffs.csv} contains these decision rationales and the
additional steps used below; it does not contain full model reasoning streams.
A rationale describes the agent's stated basis for a decision, not an
independently validated physical explanation.

\subsection{Outcomes following an unsuccessful evaluation}

Every four-query episode has three adjacent query pairs. Of the 270 pairs,
64 begin with a request that returned no complete score. Table~\ref{tab:agent-transitions}
counts what happens immediately afterward. A changed geometry is identified
by its content hash; it does not, by itself, demonstrate that feedback caused
the change. Query-4 failures have no successor within the common horizon.

\begin{table}[htbp]
\centering\small
\caption{All transitions following a failed evaluation within queries 1--4.
Counts condition on different preceding failures and are correlated within
episodes; they are not a causal comparison of recovery strategies.}
\label{tab:agent-transitions}
\begin{tabular}{lrrrr}
\toprule
Method & Transitions & Episodes & New geometry & Next valid \\
\midrule
Open-ended research & 27 & 21 & 27 & 19 \\
Planning & 20 & 14 & 18 & 10 \\
Reflection & 17 & 15 & 17 & 15 \\
\bottomrule
\end{tabular}
\end{table}

This shows that the traces contain both successful and unsuccessful next
steps, rather than only successful anecdotes. It cannot attribute recovery
to a particular heuristic or prompt. The accompanying script emits every
transition so that alternative descriptive summaries can be checked.

\subsection{Selected decisions and their observed outcomes}

\textbf{A076, open-ended, low load: constrain extrapolation, then isolate a
change.} Query 1 tested a 47.73-degree sweep wing with predicted $L/D$ 20.64
and failed the mesh gate. Query 2 reduced sweep to 37.10 degrees and used a
dataset-trained mesh-quality estimate; CFD returned 16.04 against a surrogate
prediction of 18.30. Query 3 had a higher surrogate prediction, 19.08, but a
lower CFD score, 15.74. The next steps demonstrate both a failed adjustment
and a more controlled test (Table~\ref{tab:agent-case-a076}).

\begin{table}[htbp]
\centering\small
\caption{A076's subsequent decisions. Queries 5--8 belong to the selected
open-ended extension, not the common four-query comparison.}
\label{tab:agent-case-a076}
\begin{tabular}{cp{0.68\linewidth}r}
\toprule
Query & Recorded change & CFD $L/D$ \\
\midrule
4 & Increase q2 sweep to 38.3 degrees and reduce root thickness & timeout \\
5 & Near-repeat of q4, with root adjustment changed by 0.002 & timeout \\
6 & Return to q2 and change only sweep to 38.3 degrees & 16.35 \\
7 & Add a smaller root-thickness change & 16.64 \\
8 & Increase sweep to 39.0 degrees & 16.77 \\
\bottomrule
\end{tabular}
\end{table}

The q6 rationale explicitly tests whether sweep or thickness contributed to
the timeout. This is an example of using a controlled change after failure,
not proof of the timeout's physical cause or of the heuristic's general value.

\textbf{A080, open-ended, nominal load: revise a numerical optimizer's
objective.} The agent used CMA-ES on AeroTransformer. Its first two predictions
were 32.76 and 35.70, while CFD returned 32.06 and 33.15. It then used a
surrogate-bias correction based on distance to the source data: the next two
CFD scores were 33.4018 and 33.4014. The q4 rationale specifies the penalized
objective $y-5.083d$, where $d$ is nearest-source distance in the agent's
normalized coordinates. This illustrates optimizer orchestration and a
feedback-dependent change, but fitting a correction from two observations is
weak evidence for calibration, and the fourth step did not improve the score.

\textbf{A006, planning, high load: add a geometric screen.} The first wing
scored 28.52. A local refinement and then a smaller-radius refinement both
failed meshing. The executor subsequently computed a local surface-triangle
quality measure and screened candidates at the first wing's quality level.
Its fourth wing scored 28.55. The tool trace records the screening computation,
and the handoff names the threshold (0.0114), the candidate value (0.01232),
and the earlier failed candidates' values (at most 0.0106). The gain over q1
is only 0.03. Moreover, this episode's handoffs explicitly report substituting
an executable procedure for placeholder plans; it demonstrates executor
adaptation, not successful planning in isolation.

\textbf{A054, reflection, nominal load: distinguish predictors.} Four valid
scores increased from 28.79 to 31.14. On q2--q4, a supplementary predictor
reported 34.11, 34.35 and 34.84, about 12\% above CFD, while AeroTransformer
stayed within 0.5\%. The later rationales label the local predictor diagnostic
only. Building another model need not provide better candidate selection;
its disagreement should itself be tested against CFD.

\textbf{A035, reflection, high load: test predictor disagreement.} The q2
handoff prefers a candidate from the 2110 family over the higher-surrogate
5590 family because the latter conflicts with the parent's observed
high-Mach performance. At q4, the agent selects a 5590-family candidate as
an explicit test of disagreement between AeroTransformer (24.85) and its
local ensemble (22.91). CFD returns 24.89. The sequence shows how additional
modeling can discourage a productive direction, and how a subsequent
experiment can challenge that judgment; it does not isolate the ensemble's
effect from the other evidence used in selection.

\subsection{Implementation fidelity and limits of the lessons}

Twenty of the 120 planning handoff rationales, spanning seven episodes,
explicitly contain the word ``placeholder.'' This reproducible lexical count
is a lower bound on explicit mentions, not a classification of all plans.
Together with A006, it shows that valid structured output does not ensure an
actionable plan. Execution can supply substantive reasoning omitted by planning.

The campaign also used bounded schema or finalization recoveries. By query 4,
there were ten successful recovery decisions in nine episodes (six planning,
three reflection); the full extension adds two decisions in two open-ended
episodes. A006 and A054 each had a q1 recovery. All episodes remain included
in the reported outcomes. \texttt{data/agent\_recoveries.csv} identifies these
events; recovery assistance, additional stages and differing model usage are
part of the tested implementations, not effects isolated by randomization.

The actionable hypotheses are to preserve a verified fallback, make failure
diagnostics change candidate generation or screening, test surrogate
disagreement with observed CFD, and validate plans for executability. These
are motivated by the records and baseline comparisons. Demonstrating that any
one improves agent performance requires a prospective ablation; this audit
does not supply that causal evidence.

%% file: app_formal.tex
\section{Formal problem statement and method equations}
\label{app:formal}

This appendix states the benchmark and the reference methods in symbols. It
adds no new claim. Every quantity is the one the released code computes, and
Table~\ref{tab:app-settings} lists the settings.

\subsection{Design space, evaluator, and score}

A design is a vector $z \in \mathcal{B} = [-3.5, 3.5]^{38}$ of standardized
coordinates. Its physical parameters are $x = \mu + \sigma \odot z$, where
$\mu$ and $\sigma$ hold the per-coordinate mean and standard deviation of the
SuperWing designs and $\odot$ is the coordinatewise product. A task is a target
lift coefficient $C_L^{*} \in \{0.15, 0.41, 0.66\}$. Every design is evaluated
at three flow conditions $c \in \{1, 2, 3\}$ with Mach numbers $0.7853$,
$0.8195$, and $0.8530$ and Reynolds number $2 \times 10^{7}$, as in
Section~\ref{sec:benchmark}.

For one condition the evaluator meshes the wing, adjusts the angle of attack
$\alpha$ until the computed lift coefficient matches the target, and returns
the lift-to-drag ratio at that angle,
\begin{equation}
r_c(x) = \frac{C_L(x, c, \alpha_c^{*})}{C_D(x, c, \alpha_c^{*})},
\qquad
\bigl|C_L(x, c, \alpha_c^{*}) - C_L^{*}\bigr| \le 0.02 .
\label{eq:condition}
\end{equation}
The condition is valid only if a mesh exists, the solver process exits
normally, the solver's lift oscillation measure is below $5 \times 10^{-4}$, the
trimmed angle lies in $[-10^{\circ}, 25^{\circ}]$, and $0 < r_c(x) \le 40$.
Otherwise the condition returns a failure type instead of a number. The score
is the geometric mean over the three conditions,
\begin{equation}
y(x) = \Bigl( \prod_{c=1}^{3} r_c(x) \Bigr)^{1/3},
\label{eq:score}
\end{equation}
and it exists only when all three conditions are valid. We write
$d(x) \in \{0, 1\}$ for this event, delivery. When $d(x) = 0$ the
evaluation returns one of three outcome types. A conclusive failure means the
geometry was rejected or a condition was conclusively invalid. Partial means
some conditions returned values and none was invalid. Undetermined means no
condition returned one. In the released records every conclusive failure is a
rejected geometry.

The surrogate $s(x)$ is the same geometric mean computed from AeroTransformer
predictions with the same trim procedure. One surrogate evaluation takes about
a second and returns a number for every finite $x$, including designs with
$d(x) = 0$.

\subsection{An episode}

An episode is one search on one task. Let $\mathcal{H}_k$ denote everything the
optimizer has observed after $k$ online evaluations, that is, its surrogate
queries, the designs it submitted, and for each submitted design either the
score or the failure type. A policy $\pi$ maps $\mathcal{H}_k$ either to a
batch of at most 16 surrogate queries or to the next online design $x_{k+1}$.
The budgets are
\begin{equation}
\text{surrogate evaluations} \le 4096, \qquad k \le 32,
\label{eq:budgets}
\end{equation}
and online evaluations are strictly sequential, so $x_{k+1}$ may depend on
$d(x_k)$ and $y(x_k)$. At each checkpoint $q \in \mathcal{Q} = \{0, 1, 2, 4, 8,
16, 32\}$ the policy also names one recommendation $\rho_q \in \mathcal{B}$,
the design it would hand to a user. The recommendation is graded by the same
evaluator, the grade is never shown to the policy, and it costs no budget. If
$\rho_q$ was already evaluated online, the grade reuses that result. A design
submitted twice in one episode returns its recorded result and still costs a
query.

\subsection{Endpoints}

For one method and one task with episodes $e = 1, \dots, n$, where $n = 10$
seeds, write $d_q^{e} = d(\rho_q^{e})$ and $y_q^{e} = y(\rho_q^{e})$. Then
\begin{equation}
D_q = \frac{1}{n} \sum_{e=1}^{n} d_q^{e},
\qquad
Q_q = \frac{\sum_{e} d_q^{e}\, y_q^{e}}{\sum_{e} d_q^{e}},
\qquad
O = \frac{1}{32\, n} \sum_{e=1}^{n} \sum_{k=1}^{32} d(x_k^{e}).
\label{eq:endpoints}
\end{equation}
$D_q$ is recommendation delivery, $Q_q$ is conditional quality, defined only
when some recommendation delivers, and $O$ is online delivery. The paper
reports $D_q$ and $Q_q$ at $q = 32$ for the baselines and at $q = 4$ for the
agents. The delivery averaged across tasks in Section~\ref{sec:analysis} is the
mean of $D_{32}$ over tasks.

\subsection{The belief model of the surrogate-assisted baselines}

The four surrogate-assisted baselines share one belief model, fitted anew in
every episode from the 62 warm-start results and the episode's own online
outcomes. Its feature map is
\begin{equation}
\phi(x) = \bigl( 1,\; z_1/3.5,\; \dots,\; z_{38}/3.5,\; \log s(x) - \log 30
\bigr) \in \mathbb{R}^{40}.
\label{eq:features}
\end{equation}
The delivery model is a weighted logistic regression with an $\ell_2$ penalty,
\begin{equation}
\begin{aligned}
p(x) &= \operatorname{clip}\bigl( \sigma(w^{\top} \phi(x)),\, 0.05,\, 0.95 \bigr), \\
w &= \arg\min_{w} \sum_{i} \omega_i\, \ell\bigl( d_i, \sigma(w^{\top} \phi(x_i)) \bigr)
+ 4\, \| w_{1:39} \|^{2} + 0.1\, w_0^{2},
\end{aligned}
\label{eq:delivery-model}
\end{equation}
where $\ell$ is the logistic loss, the label $d_i$ is 1 for a pass and 0
otherwise, and the weight $\omega_i$ is 1 for a pass or a conclusive failure
and $0.5$ for a partial or undetermined outcome. Appendix~\ref{app:weight}
varies this weight. The fit runs 40 iterations. The discrepancy model is a
ridge regression, on the delivered results, of the log ratio between the CFD
score and the surrogate score,
\begin{equation}
\begin{aligned}
\delta_i &= \log y_i - \log s(x_i), \qquad \hat{\delta}(x) = v^{\top} \phi(x), \\
v &= \arg\min_{v} \sum_{i:\, d_i = 1} \bigl( \delta_i - v^{\top} \phi(x_i) \bigr)^{2}
+ 10\, \| v_{1:39} \|^{2} + 0.1\, v_0^{2},
\end{aligned}
\label{eq:discrepancy-model}
\end{equation}
with a predictive standard deviation $\hat{\sigma}(x)$ floored at $0.05$ and
set to $0.25$ while no delivered result exists. The private merit of a design
combines the two models,
\begin{equation}
m(x) = \log p(x) + \log s(x) + \hat{\delta}(x) + \tfrac{1}{2}\, \hat{\sigma}(x)
+ \tfrac{1}{10}\, H\bigl( p(x) \bigr),
\qquad
H(p) = -p \log p - (1 - p) \log (1 - p).
\label{eq:merit}
\end{equation}
The first three terms are the log of the expected delivered score, the
delivery probability times the corrected surrogate score. The last two terms
reward uncertainty in value and delivery.

\subsection{The five baseline variants}

All four surrogate-assisted variants run their generator in unit coordinates
on the box with a population of 16 and 256 generations, which is the
4{,}096-evaluation budget, and keep an active archive of the 64 best
cheap-valid designs. A design whose surrogate evaluation is not finite receives
a loss of $10^{12}$.

\textbf{CMA-ES, no feedback.} The objective is $s(x)$. All 256 generations run
before the first online query. The 32 best distinct archive designs are
queued in descending $s$, and online results change nothing.

\textbf{CMA-ES, selection only.} The same search and the same archive. At online
step $k$ the method submits $\arg\max m(x)$ over the unqueried archive designs,
refitting $p$ and $\hat{\delta}$ after every result. The archive never
changes, so feedback picks the next design but cannot create a new one.

\textbf{Feedback-guided CMA-ES.} One continuing CMA-ES state with initial mean
$0.5$ in unit coordinates and initial step size $\sigma_0 = 0.25$. Before each
online query it runs eight generations, 128 surrogate evaluations, with
objective $m(x)$, then submits the archive $\arg\max$ of $m$. Both models
are refitted after every result, so feedback changes the selection and the
proposal distribution.

\textbf{Feedback-guided DE.} DE/rand/1/bin with $F = 0.8$ and
$\mathit{CR} = 0.9$ replaces the CMA-ES state. A
trial replaces its parent when its merit is at least the parent's, and a
mutant coordinate outside the box is redrawn uniformly. Schedule and selector
are those of feedback-guided CMA-ES.

\textbf{SMAC3.} The black-box facade of SMAC 2.4.0 with its Gaussian-process
model, 8 Sobol initial designs, 24 model-guided queries, one evaluation per
configuration, and a common penalty value for every nondelivery. It uses no
surrogate and no warm-start results.

\textbf{Recommendation rule.} For every baseline, if any online evaluation has
delivered by checkpoint $q$,
\begin{equation}
\rho_q = \arg\max_{k \le q:\; d(x_k) = 1} y(x_k),
\label{eq:recommendation}
\end{equation}
which is the best delivered online design. Otherwise the recommendation is an
untested design chosen by the method's own rule, the unqueried archive design
with the highest $s$ for the no-feedback method, with the highest selector
value for the selection-only method, and with the highest $m$ for the two
feedback-guided methods, and the next proposal for SMAC3. Ties break by a
stable candidate identifier. The checkpoint-32 recommendation is the terminal
one.

\subsection{Statistical procedures}

\textbf{Seed-block contrasts for the baselines.} For two methods $A$ and $B$
and seed $j$, the delivery effect at checkpoint 32 is the mean over the three
tasks $t$ of $d_{A,t,j} - d_{B,t,j}$, and the quality effect is the mean of
$y_{A,t,j} - y_{B,t,j}$ over the tasks where both recommendations deliver,
undefined when there is none. With $n_b$ available blocks and effects
$\Delta_1, \dots, \Delta_{n_b}$, the estimate is $\bar{\Delta}$ and the exact
two-sided sign-flip $p$ value is
\begin{equation}
p = 2^{-n_b} \,\bigl| \{ \varepsilon \in \{-1, +1\}^{n_b} :
\bigl| \tfrac{1}{n_b} \textstyle\sum_j \varepsilon_j \Delta_j \bigr| \ge
| \bar{\Delta} | \} \bigr| .
\label{eq:signflip}
\end{equation}
The 95\% interval is the 2.5th and 97.5th percentile of $\bar{\Delta}$ over
10{,}000 resamples of the blocks with replacement, drawn from a PCG64
generator with seed 73001. The six tests are corrected by Holm's step-down procedure. With
ordered $p$ values $p_{(1)} \le \dots \le p_{(6)}$ the adjusted values are
$\tilde{p}_{(i)} = \max_{j \le i} \min\{1, (7 - j)\, p_{(j)}\}$, and a
contrast is significant when $\tilde{p} \le 0.05$. With seven blocks the
smallest attainable $p$ in Equation~\ref{eq:signflip} is $2/128 = 0.016$,
which is why Section~\ref{sec:results} reports the quality contrasts as
underpowered, not null.

\textbf{Analysis-plan scope.} The frozen development plan designated low and
nominal load as the core comparison and high load as a separate stress task.
The reported implementation pools all three. No pre-results amendment
establishing that broader scope was recovered, so the three-task estimates,
intervals and adjusted tests are exploratory. Their calculations are unchanged;
the historical script filename \texttt{15\_confirmatory\_contrasts.py} does not
confer confirmatory status. The unchanged plan is included as
\texttt{configs/benchmark/\allowbreak goldilocks\_v2\_phase4\_analysis\_plan\_v1.json}.

\textbf{Agent contrasts.} A seed label does not pair agent episodes, so the two
methods' ten episodes are resampled independently, 10{,}000 times with the same
generator. The quality difference is the difference between the two
conditional means of Equation~\ref{eq:endpoints}, and each resample
recomputes both means from its delivered recommendations.
Table~\ref{tab:agent-contrasts} lists them.

\textbf{Intervals for a single rate.} For $k$ delivered recommendations out of
$n$, Table~\ref{tab:marginal} reports the exact Clopper--Pearson interval
\begin{equation}
\bigl[ B_{\alpha/2}(k,\, n - k + 1),\; B_{1 - \alpha/2}(k + 1,\, n - k) \bigr],
\label{eq:clopper}
\end{equation}
where $B_{u}(a, b)$ is the $u$ quantile of the beta distribution, the ends
are 0 when $k = 0$ and 1 when $k = n$, and $\alpha = 0.05$. Its coverage is at
least the nominal 95\% for every $k$.

\textbf{Intervals for a difference of rates.} A bootstrap of two arms that both
deliver every recommendation has no spread, so Table~\ref{tab:agent-contrasts}
uses Newcombe's score interval. With $\hat{p} = k / n$, $z = 1.96$, and the
Wilson bounds
\begin{equation}
l, u = \frac{\hat{p} + z^{2}/2n \;\mp\; z \sqrt{\hat{p}(1 - \hat{p})/n + z^{2}/4n^{2}}}
{1 + z^{2}/n},
\label{eq:wilson}
\end{equation}
the interval for $\hat{p}_1 - \hat{p}_2$ is
\begin{equation}
\Bigl[ \hat{p}_1 - \hat{p}_2 - \sqrt{(\hat{p}_1 - l_1)^{2} + (u_2 - \hat{p}_2)^{2}},\;
\hat{p}_1 - \hat{p}_2 + \sqrt{(u_1 - \hat{p}_1)^{2} + (\hat{p}_2 - l_2)^{2}} \Bigr].
\label{eq:newcombe}
\end{equation}

\subsection{Two remarks}

\textbf{Why the surrogate is optimistic where the search goes.} Write the
surrogate score of candidate $i$ as $s_i = y_i + \epsilon_i$ with an error
$\epsilon_i$ that has mean zero for a randomly chosen design. A search that
screens many candidates and keeps $i^{*} = \arg\max_i s_i$ selects for large
positive errors as much as for large true scores. This is the optimizer's
curse of \citet{smith2006curse}. Under their assumption that the error has mean
zero given the true score, selecting the maximum gives
$\mathbb{E}[\, s_{i^{*}} - y_{i^{*}} \,] \ge 0$, and the gap grows with the
number of candidates screened and with the error variance. In WingBench the error variance is
largest away from the SuperWing data, which is where optimization pushes, and
the same selection applies to delivery, since a design whose failure the
surrogate does not see is favored exactly because that failure is not priced.
Appendix~\ref{app:classifier} measures the consequence, and the agents that
fitted their own models in Section~\ref{sec:results} met the same effect.

\textbf{Why the recommendation rule is the best delivered design.} Repeated
evaluations of the same wing agreed in 800 of 809 cases
(Appendix~\ref{app:outcomes}), so a private regrade of an already delivered
design returns its online score. Among the designs an optimizer has evaluated,
Equation~\ref{eq:recommendation} therefore maximizes the delivered score with
certainty. An untested design $x$ can only beat it in expectation when
$p(x)\, \mathbb{E}[\, y(x) \mid d(x) = 1 \,]$ exceeds the best delivered score,
and the baselines never take that gamble once a design has delivered. This is
a conservative rule, chosen because the benchmark asks for a design worth
handing to a user, not for the design with the best expected score.

\begin{table}[H]
\caption{Settings of the reference methods and of the analysis, as fixed in
the released configurations.}
\label{tab:app-settings}
\centering
\scriptsize
\begin{tabular*}{\textwidth}{@{\extracolsep{\fill}}ll}
\toprule
setting & value \\
\midrule
design box & $[-3.5, 3.5]^{38}$ standardized, edge tolerance $10^{-9}$ \\
flow conditions & Mach $0.7853$, $0.8195$, $0.8530$ at $\mathrm{Re} = 2 \times 10^{7}$ \\
lift match & $|C_L - C_L^{*}| \le 0.02$, lift oscillation $< 5 \times 10^{-4}$, $\alpha \in [-10^{\circ}, 25^{\circ}]$ \\
score envelope & $0 < r_c \le 40$ for every condition \\
time limits & 30 min per mesh, 2 h per flow condition, wall clock \\
reference machine & one job at a time on an AWS m7a.4xlarge, 16 cores, 64 GB \\
surrogate budget & 4{,}096 evaluations in batches of at most 16 \\
online budget & 32 sequential CFD evaluations \\
checkpoints & 0, 1, 2, 4, 8, 16, 32 \\
warm start & the same 62 CFD results for the four surrogate-assisted baselines \\
population, generations & 16, 256 (eight generations per online query when interleaved) \\
active archive & 64 designs \\
CMA-ES & initial mean $0.5$ in unit coordinates, $\sigma_0 = 0.25$, no mirrors \\
DE & DE/rand/1/bin, $F = 0.8$, $\mathit{CR} = 0.9$ \\
invalid surrogate value & loss $10^{12}$ \\
delivery model & logistic, $\ell_2$ penalty 4 (intercept 0.1), 40 iterations, clip to $[0.05, 0.95]$ \\
outcome weights & pass 1, conclusive failure 1, partial 0.5, undetermined 0.5 \\
discrepancy model & ridge in $\log y - \log s$, penalty 10 (intercept 0.1), $\hat{\sigma} \ge 0.05$, 0.25 with no label \\
merit & $\log p + \log s + \hat{\delta} + 0.5\, \hat{\sigma} + 0.1\, H(p)$ \\
SMAC3 & version 2.4.0, black-box facade, 8 Sobol points, 24 model-guided queries \\
agents & DeepSeek V4.1 Flash, maximum reasoning effort, 64 tool rounds per turn \\
reflection & summary capped at 4{,}000 characters \\
planning & four plans of two steps, at most three cycles per decision \\
bootstrap & 10{,}000 resamples, PCG64 seed 73001, 95\% percentile intervals \\
multiple comparisons & Holm over six exploratory baseline contrasts at $\alpha = 0.05$ \\
\bottomrule
\end{tabular*}
\end{table}

%% file: app_rules.tex
\section{Adding methods and preserving comparable versions}
\label{app:rules}

The complete rules are in \texttt{benchmark/RULES.md}. The public runner
supports a 38-parameter wing, three tasks, a 4{,}096-surrogate cap and a
32-query CFD horizon. It records proposals, feedback and private nominations;
\texttt{verify.py} checks the trace and \texttt{score.py} summarizes outcomes.
The historical studies' actual access conditions and exceptions are reported
in Section~\ref{sec:analysis} and Appendix~\ref{app:resources}; they must not be
inferred from the public interface's maximum allowances.

\subsection{Integration contract}

A new method subclasses \texttt{Optimizer} in
\texttt{benchmark/optimizer\_api.py}. Its three decision methods are
\texttt{propose(ctx)}, returning the next 38-coordinate wing;
\texttt{observe(result)}, receiving the score or failure codes; and
\texttt{recommend()}, returning its current recommendation. Optional
\texttt{start(ctx)} initializes state. The context supplies the task, design
box, history, remaining budgets and the surrogate query function. A method
may maintain local state and call a language model inside this interface;
the evaluator and private grading contract need not change.

The worked optimizer and recorded-result evaluator in
\texttt{benchmark/README.md} exercise integration without CFD. The recorded
backend can answer only already graded wings and cannot validate a new
optimizer's aerodynamic performance. Fresh evaluations use the CFD backend
and the documented container. The original distributed agent campaign
controller is not included; reproducing that campaign's orchestration
requires additional engineering. The released interface, prompts and tools
support new implementations, not one-command reproduction of that controller.

\subsection{Versioning and maintenance scope}

The preservation policy is to retain the original records and content-pinned
protocol, distinguish documentation corrections from evaluator changes, and
report results under different mesh, solver, budget or outcome-policy versions
separately. A rebuilt runtime receives its own fingerprint and protocol
revision; it is not silently substituted for the original environment.
The container recipe, source/weight hashes, recorded example and analysis
scripts provide checks against dependency drift. Issues and integration
instructions are maintained with the repository, but the release offers no
hosted grading service or guaranteed maintenance duration. These mechanisms
reduce dependence on a live service; they do not guarantee that dependencies
will remain available or that future maintainers will have CFD resources.

For future submissions, the runner's closed track permits the provided
source data and warm-start records; an open track also permits public
benchmark records. The historical agents did not receive the 62-record
baseline context. Agent seed labels identify episodes and do not create
paired random draws across methods. Report actual resource use and protocol
version alongside recommendation validity and mean $L/D$ among valid
recommendations. A failed computation is a failure under that protocol,
not a certificate of physical infeasibility.

%% file: app_data_cfd.tex
\section{Released data}
\label{app:datacard}

The field-level data dictionary is \texttt{data/README.md}. Tables join on
\texttt{episode}, \texttt{evaluation} and the geometry content hash
\texttt{wing}. The principal files are summarized below. Missing measurements
remain missing; a reused grade is not a new independent solver execution.

\begin{table}[htbp]
\centering\small
\caption{Data map, relative to \texttt{data/}. Full schemas and coverage notes
are in the repository data dictionary.}
\label{tab:data-overview}
\begin{tabular}{p{0.42\linewidth}p{0.50\linewidth}}
\toprule
File & Content \\
\midrule
\texttt{episodes.csv} & 240 episodes, methods, tasks and seed labels \\
\texttt{evaluations.csv} & 6{,}629 online/private requests, scores and failure codes \\
\texttt{recommendations.csv}, \texttt{wings.csv} & Recommendation endpoints and recoverable 38-coordinate geometries \\
\texttt{cfd\_attempts.csv}, \texttt{cfd\_conditions.csv} & Available attempt timings and condition-level CFD evidence \\
\texttt{agent\_nominations.csv}, \texttt{agent\_turns.csv} & Standing recommendations and recorded model-session usage \\
\texttt{surrogate\_usage.csv} & Charged surrogate use at the reported checkpoints \\
\texttt{agent\_handoffs.csv}, \texttt{agent\_recoveries.csv} & Decision rationales, joined outcomes and bounded recovery events \\
\texttt{accepted\_mesh\_flow\_audit.csv} & 11{,}535 audited condition records after accepted meshes \\
\texttt{mesh\_validation\_checks.csv} & Source controls and paired upstream/default-layer diagnostics \\
\texttt{regrades\_after\_outage.csv} & Separate post-study grades of the five outage wings \\
\texttt{superwing/} & Source geometry and run records, shared with permission \\
\bottomrule
\end{tabular}
\end{table}

There are 150 baseline episodes and 90 agent episodes. Online requests number
4{,}800 for baselines, 360 at the common agent horizon, and 840 in the open-ended
extension; 629 additional private grades complete the total. Recommendations
that reuse an online grade do not add a request. The 6{,}629 records are thus
neither a training-dataset target nor independent statistical replicates.
Five evaluated geometries are unavailable. The original condition table covers
6{,}241 result documents; the later accepted-mesh audit is a separately
recovered subset with its own provenance, not an assertion of complete raw
coverage. Agent seed labels do not pair model sampling across methods.

\section{CFD protocol and environment}
\label{app:cfdprotocol}

The protocol is
\texttt{configs/cfd/\allowbreak final\_protocol\_candidate\_v2.json};
\texttt{scripts/cfd/grading\_protocol.py} verifies its source bindings.
The local grading wrapper builds one mesh, runs three conditions serially,
and records both aggregate eligibility and per-condition evidence. The
configuration retains its development identity (revision 3,
\texttt{qualification\_required}, \texttt{official\_grading\_enabled=false});
the public local wrapper is not a hosted official grading authority.

\subsection{Mesh construction}

The geometry and mesher come from SuperWing's published AeroTransformer
simulation files at commit
\texttt{3dc350ff69e354d1451eae468c686a368bd6151f}. The case preparer copies the
geometry inputs, adjusts the local tip-template path and MPI invocation, and
uses the default mesh profile with \texttt{robust\_mesh=false}. The default
volume march uses 81 layers; the optional robustness patch is not used in
reported benchmark grades. First-cell spacing targets $y^+=1$ using the mean
of the input Mach conditions, at 300 K and Reynolds number $2\times10^7$.
Benchmark and source Mach conditions can therefore give different spacing for
the same geometry. An accepted mesh must have positive minimum cell
volume and minimum quality. Quality between 0 and 0.05 is recorded as marginal
but accepted. Missing or malformed evidence and mesh-process stops are
inconclusive, distinct from a completed quality-gate rejection. Meshing has
a 1{,}800-second wall-time limit. Appendix~\ref{app:evaluator-checks}
reports spacing controls, upstream-script checks and a separate 129-layer
sensitivity test; those diagnostic meshes do not replace original grades.

\subsection{Operating conditions, solver and acceptance}

All tasks use Mach 0.7853, 0.8195 and 0.8530, Reynolds number $2\times10^7$,
Reynolds length 1.0, temperature 300 K, generated wing surface area as the
reference area, reference chord 1.0 and reference position 0.25. Lift targets
are 0.15, 0.41 and 0.66. Initial angles of attack are respectively
4.4259755152, 6.6401110468 and 8.924059800550001 degrees, fixed from source-data
medians within 0.02 of each lift target. They are not reselected per wing.

\begin{table}[htbp]
\centering\small
\caption{Principal fixed solver and trim settings. The configuration and
patched runner specify all remaining options explicitly.}
\label{tab:cfd-solver}
\begin{tabular}{p{0.43\linewidth}p{0.49\linewidth}}
\toprule
Setting & Value \\
\midrule
Flow model and multigrid & ADflow RANS; \texttt{3w} cycle \\
Nonlinear solvers & Approximate Newton--Krylov and Newton--Krylov enabled \\
NK switching tolerance & $10^{-8}$ \\
Residual convergence target & $10^{-10}$ \\
Iteration limits & 500 coarse-grid cycles; 3{,}000 solve cycles \\
Lift matching (\texttt{solveCL}) & Step 0.5 degrees; tolerance 0.001; maximum 20 iterations \\
Trim relaxation & Both lift-slope and angle relaxation 1.0; no supplied lift-slope guess \\
Stall handling & No early stall abort; \texttt{stop\_on\_stall=false} \\
Parallelism & 16 MPI ranks, one OpenMP/BLAS thread per rank \\
Time limit & 7{,}200 wall seconds per condition attempt \\
Condition order & Serial; remaining conditions run even after failure \\
\bottomrule
\end{tabular}
\end{table}

A condition must exit with code zero, supply a converged \texttt{solveCL}
result, leave both ADflow failure flags false, and have lift oscillation below
0.0005. Lift must be within 0.02 of target, angle of attack in $[-10,25]$
degrees, and lift and drag positive. The benchmark additionally enforces
$0<L/D\leq40$. A valid aggregate score requires all three conditions and is
the geometric mean of their $L/D$. Partial results remain diagnostics.
Timeouts and process errors are inconclusive computational outcomes;
completed trim nonconvergence is a distinct invalid-solver outcome. None
certifies physical impossibility of the design.

The grader has no internal infrastructure retries. The campaign controller
handles external worker loss separately; ten pre-CFD requests were nevertheless
charged during the agent extension, a documented exception
(Appendix~\ref{app:agent-comparison}). Repeated within-episode wings reuse
recorded results while consuming a query. These accounting rules must be
separated from counts of independent executions.

\subsection{Environment and scope of reproducibility}

The reference deployment assigns one wing 16 vCPUs and 64 GB on an AWS
\texttt{m7a.4xlarge}. Historical records also identify cluster workers, so
identical hardware is not established. Because limits are wall-clock based,
hardware and contention affect acceptance. A wing can spend up to 6.5 hours
in meshing and the three conditions. The grader sets the solver/thread
variables itself, rejects conflicting parent overrides, requires a non-root
account and serializes jobs on the reference worker.

The Dockerfile and Apptainer definition in \texttt{benchmark/environment/}
rebuild from the digest-pinned \texttt{mdolab/public} Ubuntu 20.04 image with
Open MPI 4.1.6, Python 3.9.12 and the MDO Lab solver/mesher libraries. The
geometry library \texttt{cst-modeling3d}, SuperWing scripts and runtime probe
are content-pinned. \texttt{benchmark/environment/ENVIRONMENT.md} gives the
full commits, image digests and hashes, avoiding duplicated opaque identifiers
here. AeroTransformer ATsurf-L runs separately on CPU in float32; its weight
fingerprints and retrieval instructions are in \texttt{benchmark/README.md}.

A fresh reference-machine rebuild reproduced one nominal wing's three
conditions and score (33.401823323619524) exactly on 25 September 2026.
This checks one case, not mesh independence or general evaluator validity.
The rebuild does not recreate the exact historical image digest or runtime
fingerprint. A new probe hash requires a separate protocol revision. The
original authorization, attestation and remote scheduling services are not
distributed; the released local grader runs independently. Likewise, the
agent prompts and tools are available, but the original distributed session
controller and full reasoning streams are not. The additional checks in
Appendix~\ref{app:evaluator-checks} address implementation consistency and
settings sensitivity without claiming equivalence of every historical runtime.

%% file: app_gallery.tex
\section{The best delivered wings}
\label{app:gallery}

\begin{figure}[H]
\centering
\includegraphics{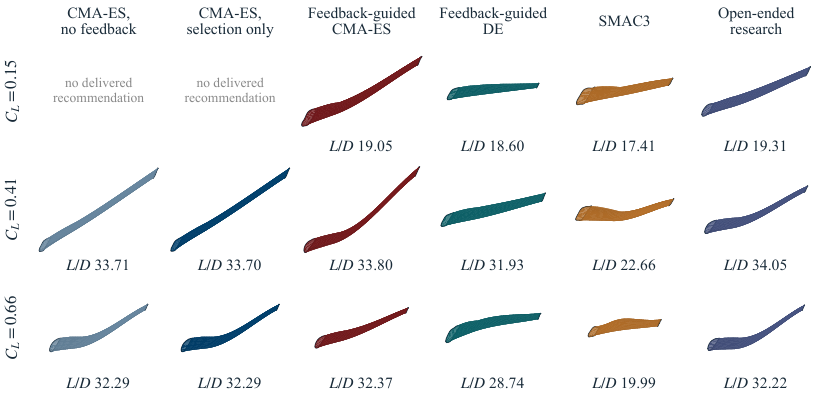}
\caption{The best delivered wing of each method and task, drawn as in Figure~\ref{fig:wing-mesh-outcomes} at one scale. For the five baselines it is the delivered recommendation after 32 queries with the highest $L/D$ over the ten seeds. For open-ended research it is the delivered online design with the highest $L/D$ among queries 5 to 32 of its 30 extension episodes. Each number is the wing's $L/D$. The two archive-only CMA-ES methods delivered nothing after 32 queries at $C_L=0.15$, so those cells are empty.}
\label{fig:wing-gallery}
\end{figure}

This appendix shows the best wing each method delivered. For the five baseline optimizers, we took the recommendation each method made after 32 queries in each of its ten episodes per task and kept the delivered one with the highest $L/D$. The open-ended research agent made no further requested checkpoint recommendation after query 4, so for it we took the delivered online design with the highest $L/D$ among queries 5 to 32 of its 30 extension episodes. Figure~\ref{fig:wing-gallery} draws the 16 wings at one scale, and Table~\ref{tab:wing-gallery} lists their planform values. CMA-ES with no feedback and CMA-ES with selection only each delivered no recommendation after 32 queries at $C_L=0.15$, so those two cells are empty. CMA-ES with no feedback and CMA-ES with selection only delivered the same wing at $C_L=0.66$. The best wings of the different methods are all thin and swept, with a high aspect ratio. Their leading-edge sweep ranges from 26 to 44 degrees, their aspect ratio from 8.9 to 12.5, their taper ratio from 0.26 to 0.52, and their root thickness ratio from 0.126 to 0.138. The highest score at each load is 19.31 for open-ended research at $C_L=0.15$, 34.05 for open-ended research at $C_L=0.41$ and 32.37 for feedback-guided CMA-ES at $C_L=0.66$. Every wing shown was delivered by the CFD evaluator, so each score is a graded result rather than a surrogate prediction.

\begin{table}[H]
\caption{The wings of Figure~\ref{fig:wing-gallery}. Seed is the episode seed, and for open-ended research the episode label. Sweep is the leading-edge sweep in degrees, taper ratio is the tip chord over the root chord, and root $t/c$ is the thickness of the root section over its chord. The four are columns 0, 3, 4 and 37 of the 38 design variables, in the order of the released tables and the guide, and they are the planform values the guide defines without an interpolation rule. A dash marks a cell with no delivery.}
\label{tab:wing-gallery}
\centering
\scriptsize
\setlength{\tabcolsep}{4pt}
\begin{tabular*}{\textwidth}{@{\extracolsep{\fill}}llrrrrrr}
\toprule
method & $C_L$ & seed & $L/D$ & sweep ($^\circ$) & aspect ratio & taper ratio & root $t/c$ \\
\midrule
CMA-ES, no feedback & $0.15$ & -- & -- & -- & -- & -- & -- \\
 & $0.41$ & 6206 & 33.71 & 39.9 & 12.49 & 0.524 & 0.126 \\
 & $0.66$ & 6208 & 32.29 & 37.7 & 12.49 & 0.311 & 0.127 \\
CMA-ES, selection only & $0.15$ & -- & -- & -- & -- & -- & -- \\
 & $0.41$ & 6206 & 33.70 & 39.8 & 12.49 & 0.524 & 0.127 \\
 & $0.66$ & 6208 & 32.29 & 37.7 & 12.49 & 0.311 & 0.127 \\
Feedback-guided CMA-ES & $0.15$ & 6202 & 19.05 & 43.6 & 12.27 & 0.524 & 0.138 \\
 & $0.41$ & 6201 & 33.80 & 37.1 & 12.49 & 0.524 & 0.126 \\
 & $0.66$ & 6209 & 32.37 & 36.1 & 12.49 & 0.272 & 0.126 \\
Feedback-guided DE & $0.15$ & 6207 & 18.60 & 40.9 & 11.87 & 0.257 & 0.130 \\
 & $0.41$ & 6203 & 31.93 & 37.8 & 12.32 & 0.407 & 0.131 \\
 & $0.66$ & 6204 & 28.74 & 38.9 & 12.15 & 0.292 & 0.128 \\
SMAC3 & $0.15$ & 6202 & 17.41 & 39.8 & 10.54 & 0.475 & 0.129 \\
 & $0.41$ & 6202 & 22.66 & 26.1 & 11.74 & 0.524 & 0.126 \\
 & $0.66$ & 6204 & 19.99 & 36.7 & 8.90 & 0.356 & 0.135 \\
Open-ended research & $0.15$ & 9531 & 19.31 & 39.5 & 11.52 & 0.524 & 0.128 \\
 & $0.41$ & 9551 & 34.05 & 39.3 & 12.45 & 0.357 & 0.131 \\
 & $0.66$ & 9531 & 32.22 & 39.0 & 12.46 & 0.283 & 0.133 \\
\bottomrule
\end{tabular*}
\end{table}

%% file: app_results.tex
\section{Full results tables}
\label{app:fulltables}

The tables in this appendix give results per method and task; individual episode records remain in the accompanying CSV tables. They are computed from the released data tables by \texttt{analysis/17\_appendix\_tables.py}.

Table~\ref{tab:app-baseline-recommendations} gives, for every baseline method, task and checkpoint, how many of the ten recommended wings delivered a CFD score and the mean $L/D$ of those that did.

\begin{table}[H]
\caption{Baseline recommendations at every checkpoint. A checkpoint is the number of online CFD queries the method had made when it recommended the wing, so checkpoint 0 is the recommendation made before any CFD feedback. For each method and task, $n$ is the number of the ten episodes whose recommended wing delivered a CFD score, and $L/D$ is their mean score.}
\label{tab:app-baseline-recommendations}
\centering
\scriptsize
\setlength{\tabcolsep}{3pt}
\begin{tabular*}{\textwidth}{@{\extracolsep{\fill}}lrrrrrrrrrrrrrr}
\toprule
method & \multicolumn{14}{c}{online CFD queries before the recommendation} \\
\cmidrule(lr){2-15}
 & \multicolumn{2}{c}{0} & \multicolumn{2}{c}{1} & \multicolumn{2}{c}{2} & \multicolumn{2}{c}{4} & \multicolumn{2}{c}{8} & \multicolumn{2}{c}{16} & \multicolumn{2}{c}{32} \\
\cmidrule(lr){2-3}\cmidrule(lr){4-5}\cmidrule(lr){6-7}\cmidrule(lr){8-9}\cmidrule(lr){10-11}\cmidrule(lr){12-13}\cmidrule(lr){14-15}
 & $n$ & $L/D$ & $n$ & $L/D$ & $n$ & $L/D$ & $n$ & $L/D$ & $n$ & $L/D$ & $n$ & $L/D$ & $n$ & $L/D$ \\
\midrule
\multicolumn{15}{l}{$C_L=0.15$} \\
CMA-ES, no feedback & 0 & -- & 0 & -- & 0 & -- & 0 & -- & 0 & -- & 0 & -- & 0 & -- \\
CMA-ES, selection only & 0 & -- & 0 & -- & 0 & -- & 0 & -- & 0 & -- & 0 & -- & 0 & -- \\
Feedback-guided CMA-ES & 3 & 16.17 & 5 & 15.42 & 4 & 16.47 & 5 & 16.25 & 5 & 16.25 & 5 & 16.87 & 6 & 16.76 \\
Feedback-guided DE & 3 & 13.13 & 4 & 12.38 & 6 & 14.02 & 8 & 14.78 & 10 & 16.10 & 10 & 17.18 & 10 & 17.52 \\
SMAC3 & 1 & 6.33 & 3 & 8.87 & 3 & 8.87 & 6 & 7.48 & 7 & 9.51 & 9 & 9.87 & 9 & 10.56 \\
\addlinespace
\multicolumn{15}{l}{$C_L=0.41$} \\
CMA-ES, no feedback & 3 & 33.19 & 3 & 33.19 & 3 & 33.19 & 3 & 33.22 & 3 & 33.22 & 4 & 33.19 & 4 & 33.22 \\
CMA-ES, selection only & 3 & 33.19 & 4 & 33.14 & 4 & 33.14 & 4 & 33.18 & 4 & 33.19 & 4 & 33.21 & 4 & 33.22 \\
Feedback-guided CMA-ES & 3 & 23.58 & 5 & 22.42 & 6 & 25.82 & 10 & 28.73 & 10 & 32.06 & 10 & 32.91 & 10 & 33.04 \\
Feedback-guided DE & 3 & 19.47 & 5 & 18.79 & 7 & 20.19 & 9 & 22.35 & 10 & 25.40 & 10 & 28.94 & 10 & 31.09 \\
SMAC3 & 1 & 10.24 & 3 & 13.58 & 3 & 13.58 & 5 & 12.03 & 6 & 15.03 & 6 & 15.91 & 7 & 17.00 \\
\addlinespace
\multicolumn{15}{l}{$C_L=0.66$} \\
CMA-ES, no feedback & 4 & 32.03 & 4 & 32.03 & 4 & 32.03 & 5 & 32.04 & 5 & 32.08 & 5 & 32.10 & 5 & 32.12 \\
CMA-ES, selection only & 4 & 32.03 & 4 & 32.03 & 5 & 32.04 & 5 & 32.07 & 5 & 32.10 & 5 & 32.12 & 5 & 32.12 \\
Feedback-guided CMA-ES & 7 & 16.65 & 6 & 17.89 & 7 & 18.85 & 10 & 23.99 & 10 & 28.14 & 10 & 31.00 & 10 & 31.64 \\
Feedback-guided DE & 1 & 17.23 & 5 & 15.20 & 7 & 15.44 & 8 & 18.15 & 9 & 19.94 & 10 & 23.59 & 10 & 27.05 \\
SMAC3 & 1 & 5.66 & 2 & 9.17 & 2 & 9.17 & 3 & 8.45 & 4 & 10.43 & 5 & 12.19 & 6 & 12.82 \\
\bottomrule
\end{tabular*}
\end{table}

Table~\ref{tab:app-agent-recommendations} gives the same for the three agent methods at the checkpoints of the four-query comparison.

\begin{table}[H]
\caption{Agent recommendations at checkpoints 0, 1, 2 and 4, in the format of Table~\ref{tab:app-baseline-recommendations}.}
\label{tab:app-agent-recommendations}
\centering
\scriptsize
\setlength{\tabcolsep}{3pt}
\begin{tabular*}{\textwidth}{@{\extracolsep{\fill}}lrrrrrrrr}
\toprule
method & \multicolumn{8}{c}{online CFD queries before the recommendation} \\
\cmidrule(lr){2-9}
 & \multicolumn{2}{c}{0} & \multicolumn{2}{c}{1} & \multicolumn{2}{c}{2} & \multicolumn{2}{c}{4} \\
\cmidrule(lr){2-3}\cmidrule(lr){4-5}\cmidrule(lr){6-7}\cmidrule(lr){8-9}
 & $n$ & $L/D$ & $n$ & $L/D$ & $n$ & $L/D$ & $n$ & $L/D$ \\
\midrule
\multicolumn{9}{l}{$C_L=0.15$} \\
Open-ended research & 2 & 16.79 & 7 & 16.44 & 8 & 16.95 & 9 & 17.15 \\
Planning & 8 & 16.64 & 10 & 16.64 & 10 & 17.01 & 10 & 17.41 \\
Reflection & 6 & 16.47 & 8 & 16.59 & 10 & 16.98 & 10 & 17.33 \\
\addlinespace
\multicolumn{9}{l}{$C_L=0.41$} \\
Open-ended research & 7 & 29.12 & 9 & 28.66 & 9 & 29.79 & 10 & 31.09 \\
Planning & 7 & 28.19 & 9 & 28.28 & 9 & 28.43 & 9 & 29.31 \\
Reflection & 8 & 28.54 & 10 & 28.42 & 10 & 29.05 & 10 & 29.83 \\
\addlinespace
\multicolumn{9}{l}{$C_L=0.66$} \\
Open-ended research & 9 & 25.58 & 10 & 24.92 & 10 & 26.20 & 10 & 27.83 \\
Planning & 8 & 26.09 & 8 & 26.09 & 9 & 26.61 & 9 & 27.59 \\
Reflection & 9 & 24.42 & 10 & 24.53 & 10 & 25.75 & 10 & 26.79 \\
\bottomrule
\end{tabular*}
\end{table}

Table~\ref{tab:app-online-delivery} counts the online CFD queries of each method and task that delivered a score.

\begin{table}[H]
\caption{Online CFD delivery per method and task. Delivered is the number of online queries that returned a valid score, total is the number of online queries, and rate is their ratio. The agent rows cover queries 1 to 4 of all 90 agent episodes. The extension row covers all 32 queries of the 30 open-ended research episodes that continued, so it counts their first four queries a second time.}
\label{tab:app-online-delivery}
\centering
\scriptsize
\setlength{\tabcolsep}{4pt}
\begin{tabular*}{\textwidth}{@{\extracolsep{\fill}}lrrrrrrrrr}
\toprule
method & \multicolumn{3}{c}{$C_L=0.15$} & \multicolumn{3}{c}{$C_L=0.41$} & \multicolumn{3}{c}{$C_L=0.66$} \\
\cmidrule(lr){2-4}\cmidrule(lr){5-7}\cmidrule(lr){8-10}
 & delivered & total & rate & delivered & total & rate & delivered & total & rate \\
\midrule
\multicolumn{10}{l}{\emph{Baselines, 32 queries in each of 150 episodes}} \\
CMA-ES, no feedback & 0 & 320 & 0.00 & 83 & 320 & 0.26 & 118 & 320 & 0.37 \\
CMA-ES, selection only & 0 & 320 & 0.00 & 85 & 320 & 0.27 & 112 & 320 & 0.35 \\
Feedback-guided CMA-ES & 14 & 320 & 0.04 & 246 & 320 & 0.77 & 247 & 320 & 0.77 \\
Feedback-guided DE & 93 & 320 & 0.29 & 200 & 320 & 0.62 & 199 & 320 & 0.62 \\
SMAC3 & 114 & 320 & 0.36 & 88 & 320 & 0.28 & 49 & 320 & 0.15 \\
\addlinespace
\multicolumn{10}{l}{\emph{Agents, queries 1 to 4 of all 90 episodes}} \\
Open-ended research & 21 & 40 & 0.53 & 28 & 40 & 0.70 & 34 & 40 & 0.85 \\
Planning & 31 & 40 & 0.78 & 30 & 40 & 0.75 & 29 & 40 & 0.72 \\
Reflection & 30 & 40 & 0.75 & 29 & 40 & 0.72 & 37 & 40 & 0.93 \\
\addlinespace
\multicolumn{10}{l}{\emph{Extension, all 32 queries of the 30 continued episodes}} \\
Open-ended research & 198 & 320 & 0.62 & 225 & 320 & 0.70 & 197 & 320 & 0.62 \\
\bottomrule
\end{tabular*}
\end{table}

Table~\ref{tab:app-compute} sums the CFD timing attached to each method's records. These totals include reused outcomes and do not measure unique compute consumed (Appendix~\ref{app:resources}).

\begin{table}[H]
\caption{Recorded CFD timing per method. A core-hour is one hour of one processor core; mesh and flow-attempt durations are multiplied by 16 cores. The totals cover records with timing, including reused outcomes. The SMAC3 total includes two checkpoint-32 grades that were replaced when two SMAC3 episodes were run again.}
\label{tab:app-compute}
\centering
\scriptsize
\setlength{\tabcolsep}{4pt}
\begin{tabular*}{\textwidth}{@{\extracolsep{\fill}}lrrrr}
\toprule
method & episodes & timed evaluations & core-hours & core-hours per evaluation \\
\midrule
\multicolumn{5}{l}{\emph{Baselines}} \\
CMA-ES, no feedback & 30 & 1{,}123 & 16{,}689 & 14.9 \\
CMA-ES, selection only & 30 & 1{,}119 & 16{,}584 & 14.8 \\
Feedback-guided CMA-ES & 30 & 1{,}049 & 38{,}759 & 36.9 \\
Feedback-guided DE & 30 & 1{,}034 & 38{,}064 & 36.8 \\
SMAC3 & 30 & 954 & 35{,}307 & 37.0 \\
All baselines & 150 & 5{,}279 & 145{,}403 & 27.5 \\
\addlinespace
\multicolumn{5}{l}{\emph{Agents}} \\
Open-ended research & 30 & 968 & 42{,}113 & 43.5 \\
Planning & 30 & 162 & 6{,}112 & 37.7 \\
Reflection & 30 & 167 & 6{,}392 & 38.3 \\
All agents & 90 & 1{,}297 & 54{,}617 & 42.1 \\
\midrule
Both studies & 240 & 6{,}576 & 200{,}019 & 30.4 \\
\bottomrule
\end{tabular*}
\end{table}

Table~\ref{tab:app-online-outcomes} splits the same online queries by the grader's outcome type.

\begin{table}[H]
\caption{Outcome types of the online CFD queries, in the groups of Table~\ref{tab:app-online-delivery}. A pass delivered a score. A partial outcome has a result at some but not all flight conditions. An undetermined outcome has no result at any condition. A conclusive failure is a rejected geometry. The SMAC3 records name the reason for each nondelivery but not the outcome type, so its type counts are empty.}
\label{tab:app-online-outcomes}
\centering
\scriptsize
\setlength{\tabcolsep}{4pt}
\begin{tabular*}{\textwidth}{@{\extracolsep{\fill}}llrrrrr}
\toprule
method & $C_L$ & pass & partial & undetermined & conclusive failure & total \\
\midrule
\multicolumn{7}{l}{\emph{Baselines, 32 queries in each of 150 episodes}} \\
CMA-ES, no feedback & $0.15$ & 0 & 0 & 308 & 12 & 320 \\
 & $0.41$ & 83 & 20 & 178 & 39 & 320 \\
 & $0.66$ & 118 & 36 & 119 & 47 & 320 \\
CMA-ES, selection only & $0.15$ & 0 & 0 & 308 & 12 & 320 \\
 & $0.41$ & 85 & 21 & 177 & 37 & 320 \\
 & $0.66$ & 112 & 43 & 120 & 45 & 320 \\
Feedback-guided CMA-ES & $0.15$ & 14 & 15 & 230 & 61 & 320 \\
 & $0.41$ & 246 & 44 & 17 & 13 & 320 \\
 & $0.66$ & 247 & 37 & 35 & 1 & 320 \\
Feedback-guided DE & $0.15$ & 93 & 33 & 135 & 59 & 320 \\
 & $0.41$ & 200 & 47 & 53 & 20 & 320 \\
 & $0.66$ & 199 & 49 & 61 & 11 & 320 \\
SMAC3 & $0.15$ & -- & -- & -- & -- & 320 \\
 & $0.41$ & -- & -- & -- & -- & 320 \\
 & $0.66$ & -- & -- & -- & -- & 320 \\
\addlinespace
\multicolumn{7}{l}{\emph{Agents, queries 1 to 4 of all 90 episodes}} \\
Open-ended research & $0.15$ & 21 & 9 & 5 & 5 & 40 \\
 & $0.41$ & 28 & 6 & 5 & 1 & 40 \\
 & $0.66$ & 34 & 0 & 3 & 3 & 40 \\
Planning & $0.15$ & 31 & 6 & 2 & 1 & 40 \\
 & $0.41$ & 30 & 6 & 4 & 0 & 40 \\
 & $0.66$ & 29 & 0 & 10 & 1 & 40 \\
Reflection & $0.15$ & 30 & 5 & 3 & 2 & 40 \\
 & $0.41$ & 29 & 5 & 5 & 1 & 40 \\
 & $0.66$ & 37 & 0 & 1 & 2 & 40 \\
\addlinespace
\multicolumn{7}{l}{\emph{Extension, all 32 queries of the 30 continued episodes}} \\
Open-ended research & $0.15$ & 198 & 81 & 27 & 14 & 320 \\
 & $0.41$ & 225 & 53 & 34 & 8 & 320 \\
 & $0.66$ & 197 & 56 & 51 & 16 & 320 \\
\bottomrule
\end{tabular*}
\end{table}

All 240 episode summaries remain in \texttt{analysis/out/tables/\allowbreak tab\_appendix\_episodes.csv}, including query counts, valid-query counts, best online $L/D$, and the final recommendation. This table is generated alongside the summaries above rather than duplicated row by row in the PDF.

%% file: app_figures.tex
\section{Additional figures}
\label{app:figures}

The four figures in this appendix show the baseline study in more detail.
They are computed from the released tables \texttt{data/evaluations.csv} and
\texttt{data/recommendations.csv} by the script
\texttt{analysis/19\_appendix\_figures.py}.

\begin{figure}[H]
\centering
\includegraphics[width=\textwidth]{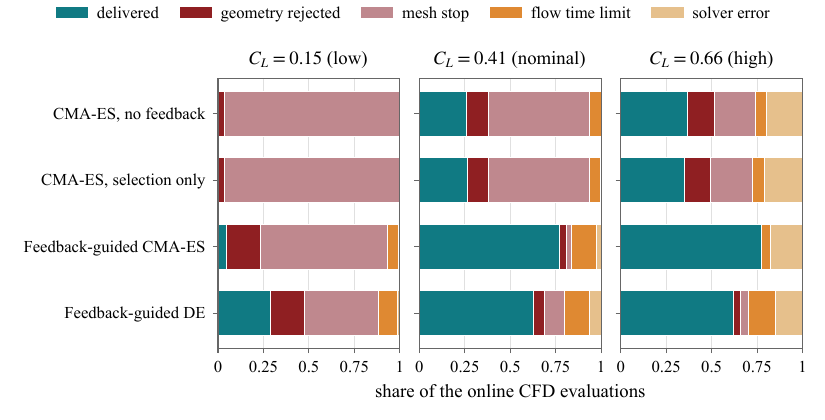}
\caption{Outcomes of the online CFD evaluations of the four surrogate-assisted
baselines, by method and task. Each bar splits the 320 online evaluations of
one method on one task, 32 queries in each of 10 runs, by outcome. The outcome
is read from the reason codes in \texttt{data/evaluations.csv} and named as in
Table~\ref{tab:outcomes}. Delivered means CFD returned a complete score.
Geometry rejected means the mesh generator rejected the wing. Mesh stop means
the mesh generator stopped without a mesh. Flow time limit means a flow
condition reached its time limit. Solver error means the solver process
stopped with an error. An evaluation with both a time limit and a solver error
counts as a time limit. None of these evaluations falls in the other-errors
group of Table~\ref{tab:outcomes}, so that group is not shown.}
\label{fig:app-yield-by-cell}
\end{figure}

\begin{figure}[H]
\centering
\includegraphics[width=\textwidth]{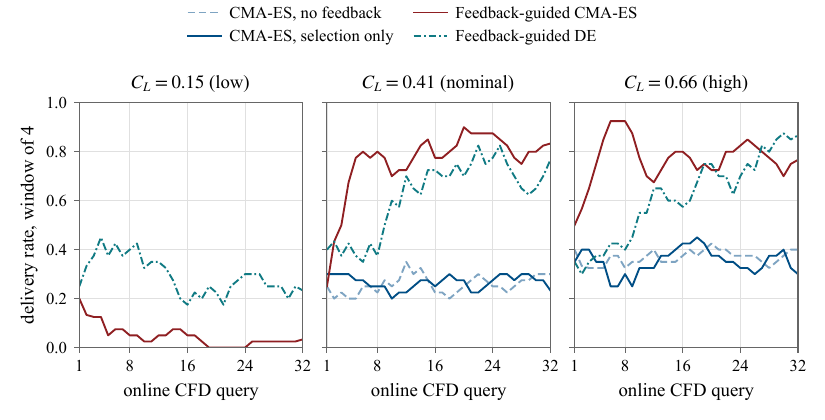}
\caption{Online delivery rate along the run for the four surrogate-assisted
baselines, by task. For each method, task and query number from 1 to 32, the
delivery rate is the share of the 10 runs whose online evaluation at that
query returned a score, from the online rows of
\texttt{data/evaluations.csv}. Each curve is the moving average of these rates
over four neighboring queries, the query itself, the one before it and the two
after it, with fewer at the ends of the run.}
\label{fig:app-online-yield-vs-index}
\end{figure}

\begin{figure}[H]
\centering
\includegraphics[width=\textwidth]{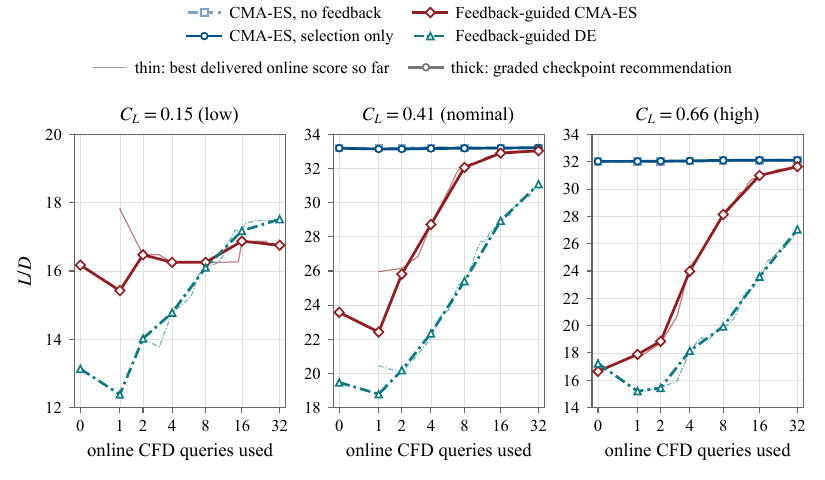}
\caption{Best delivered online score against the score of the recommended
design, for the four surrogate-assisted baselines by task. The thin line is the
best $L/D$ a run had received from its online queries so far, averaged after
each query over the runs that had received at least one score by then, from
the online rows of \texttt{data/evaluations.csv}. The thick line is the mean
$L/D$ of the delivered recommendations at the checkpoints 0, 1, 2, 4, 8, 16 and
32 queries, from \texttt{data/recommendations.csv}. A checkpoint at which no
run's recommendation delivered is left out. Both lines use delivered scores
only. CMA-ES with no feedback and CMA-ES with selection only delivered nothing
at low load, so they do not appear in that panel. The horizontal axis is spaced
by the logarithm of the query count plus one.}
\label{fig:app-discovery-vs-recommendation}
\end{figure}

\begin{figure}[H]
\centering
\includegraphics[width=\textwidth]{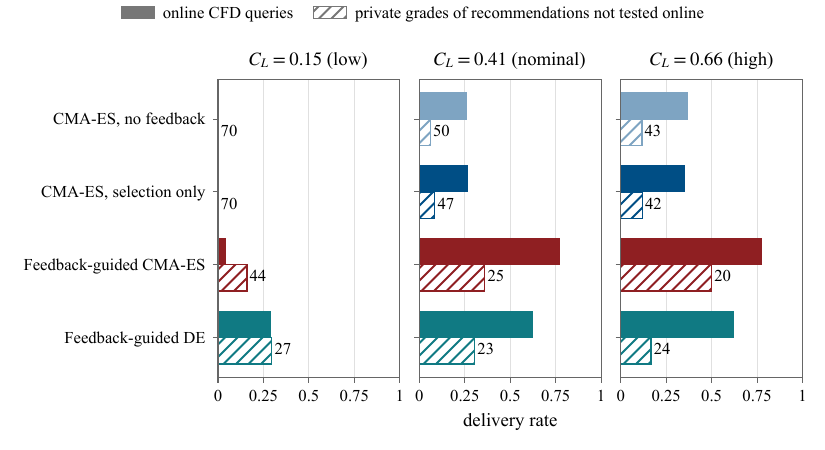}
\caption{Delivery rate of the online CFD queries against the delivery rate of
the private grades, for the four surrogate-assisted baselines by task, from
\texttt{data/evaluations.csv}. The filled bar is the share of the 320 online
queries of one method on one task that returned a score. The hatched bar is
the share of its private grades that returned a score. A private grade is the
CFD evaluation of a recommended wing that the method had not queried online,
so the method never saw the result. The number next to each hatched bar is the
count of these private grades.}
\label{fig:app-fresh-hidden-vs-online-yield}
\end{figure}

%% file: app_prompts.tex
\section{Agent prompts and task documents}
\label{app:prompts}

The complete prompt files remain in \texttt{agents/}; this appendix gives a
compact map instead of duplicating them verbatim. All methods read the same
objective, parameterization, source records, surrogate interface and CFD
feedback. The original controller that launches model sessions and remote
CFD jobs is not distributed.

\begin{table}[htbp]
\centering\small
\caption{Prompt and tool map, relative to \texttt{agents/}.}
\begin{tabular}{p{0.40\linewidth}p{0.52\linewidth}}
\toprule
File or directory & Role \\
\midrule
\texttt{workspace/TASK.md}, \texttt{tasks/task\_documents.json} & Objective, task conditions, budgets and nomination contract \\
\texttt{workspace/source/} & Parameterization, condition records, geometry and visualization helpers \\
\texttt{workspace/campaign.py} & Surrogate calls, budget inspection, handoff and nomination \\
\texttt{prompts/manager\_messages.md} & System prompt and messages resuming each decision \\
\texttt{prompts/reflection/reflect.md} & Lessons from the latest result and bounded cumulative summary \\
\texttt{prompts/planning/learner.md} & Evidence, beliefs, calibration and remaining-budget memo \\
\texttt{prompts/planning/planner.md} & Four independently proposed two-step contingent plans \\
\texttt{prompts/planning/evaluator.md} & Completion probabilities and $L/D$ quantiles for each plan \\
\texttt{prompts/planning/executor.md} & Execute the selected first procedure, allowing debugging and refinement \\
\texttt{prompts/planning/\allowbreak correction.md} & Repair a reply that fails the required JSON schema \\
\bottomrule
\end{tabular}
\end{table}

Open-ended research has no additional decision template. Reflection carries
forward the latest reflection plus a summary of at most 4{,}000 characters.
Planning uses a learner, four planners, a common evaluator and a deterministic
selection rule (Appendix~\ref{app:agent-comparison}); the executor takes one
step and then replans. At most three planning cycles are allowed per decision.
The executor may debug and refine the selected procedure, so its contribution
is substantive. Schema validity alone does not ensure executability:
Appendix~\ref{app:agent-behavior} reports explicit placeholder mentions,
bounded recoveries and observed corrections. These implementation details
limit attribution of outcomes to planning or reflection as general principles.

%% file: app_resource_followup.tex
\section{Focused resource follow-up}
\label{app:resource-followup}

This follow-up measures recommendation outcomes and resource use for selected
configurations, separately from the original 240-episode campaign.
Table~\ref{tab:resource-followup} lists the eight method--load groups, each
with three seeds (2026111201--2026111203), for 24 initiated runs. Each method
retained its 32-query search budget and 4{,}096-surrogate allowance.
Recommendations were recorded at queries 0, 1, 2 and 4, with execution capped
at query 4. Agents were instructed to optimize over the usual 32-query budget.

\paragraph{Initial information and comparison scope.}
The two CMA-ES methods shared 62 prior CFD records and evaluation allowances.
Agents received SuperWing source data without that context. Actual surrogate
use differed, and no common total-compute or wall-clock cap was imposed.
Search budgets match the original campaign; software environments differ,
and the model service's immutable checkpoint has not been confirmed to match
the original deployment. Cross-family results therefore describe the tested
configurations. This focused batch cannot rerank the full campaign under
equal information and computation.

\begin{table}[htbp]
\centering\small
\caption{Follow-up recommendations after four CFD queries. Valid counts use
all three initiated runs per group; mean $L/D$ uses valid recommendations
only. Each time column is a separate median over runs that reached query 4:
two for planning and three for every other row.}
\label{tab:resource-followup}
\setlength{\tabcolsep}{3pt}
\begin{tabular*}{\textwidth}{@{\extracolsep{\fill}}lrrrrr}
\toprule
Method / load & Valid/3 & Mean $L/D$ & \shortstack{Search\\CFD h} &
\shortstack{Grading\\CFD h} & \shortstack{Local\\CPU h} \\
\midrule
Feedback-guided CMA-ES / low & 3/3 & 17.17 & 4.12 & 0.07 & 0.47 \\
Open-ended / low & 3/3 & 17.35 & 10.89 & 2.62 & 1.73 \\
Feedback-guided CMA-ES / nominal & 3/3 & 30.32 & 6.11 & 4.54 & 0.49 \\
Selection-only CMA-ES / nominal & 1/3 & 32.87 & 0.18 & 0.18 & 3.25 \\
Open-ended / nominal & 3/3 & 30.57 & 9.21 & 2.63 & 2.22 \\
Planning / nominal & 2/3 & 30.87 & 10.84 & 2.32 & 2.08 \\
Feedback-guided CMA-ES / high & 3/3 & 24.14 & 13.19 & 2.82 & 0.55 \\
Open-ended / high & 3/3 & 30.97 & 12.74 & 2.94 & 3.06 \\
\bottomrule
\end{tabular*}
\end{table}

\paragraph{Recommendation outcomes.}
Open-ended research and feedback-guided CMA-ES each returned valid wings in
9/9 runs. Their mean $L/D$ was close at low and nominal load. At high load,
open-ended research reached 30.97 versus 24.14 for feedback-guided CMA-ES,
while using more local CPU time. Selection-only CMA-ES had the highest
nominal mean $L/D$ among these methods but returned a valid wing in only
1/3 runs. Its time medians include the two runs without valid recommendations.

\paragraph{Evaluation and model usage.}
The batch used 94 search CFD evaluations ($23\times4+2$) and 43 additional
evaluations to grade recommendations. Private grading does not count toward
the search budget. Recommendations with an existing search result reused
that result. By query 4, feedback-guided CMA-ES used 512 surrogate evaluations
per run and selection-only used all 4{,}096. Open-ended runs used
643--2{,}306, and the two completed planning runs used 1{,}068--1{,}471.
Across all 12 agent runs, we recorded 2{,}653 model calls, 397.61 million
input tokens (including 396.02 million cached) and 2.75 million output tokens.
Search decisions took 25--32 minutes for feedback-guided CMA-ES and
156--248 minutes for selection-only. Agent research and review took
76--252 minutes per run. These are recorded decision-stage durations,
including local work and tool waits, rather than pure model-inference times.

\paragraph{Resource definitions and execution.}
CFD hours sum evaluator start-to-finish execution time, separately for search
and grading; they are not processor-core hours. Local CPU hours sum user and
system CPU time across the baseline controller tree, or agent research/review
and episode-service trees, including surrogates. This is accumulated CPU time,
not elapsed time, and excludes remote CFD and model-provider computation.
No local GPUs were used. Model-provider usage is reported as calls and tokens.
CFD and baselines ran on AWS m7a.4xlarge machines (16 vCPUs); agents and
surrogates shared a CPU-only r7a.12xlarge machine. Each batch allowed up to
12 concurrent CFD jobs.

%% file: main.bbl
\begin{thebibliography}{30}
\providecommand{\natexlab}[1]{#1}
\providecommand{\url}[1]{\texttt{#1}}
\expandafter\ifx\csname urlstyle\endcsname\relax
  \providecommand{\doi}[1]{doi: #1}\else
  \providecommand{\doi}{doi: \begingroup \urlstyle{rm}\Url}\fi

\bibitem[Bachoc et~al.(2020)Bachoc, Helbert, and Picheny]{bachoc2020failures}
Fran{\c c}ois Bachoc, C{\'e}line Helbert, and Victor Picheny.
\newblock Gaussian process optimization with failures: classification and convergence proof.
\newblock \emph{Journal of Global Optimization}, 78\penalty0 (3):\penalty0 483--506, 2020.

\bibitem[Boiko et~al.(2023)Boiko, MacKnight, Kline, and Gomes]{boiko2023coscientist}
Daniil~A. Boiko, Robert MacKnight, Ben Kline, and Gabe Gomes.
\newblock Autonomous chemical research with large language models.
\newblock \emph{Nature}, 624\penalty0 (7992):\penalty0 570--578, 2023.

\bibitem[Bran et~al.(2024)Bran, Cox, Schilter, Baldassari, White, and Schwaller]{bran2024chemcrow}
Andres~M. Bran, Sam Cox, Oliver Schilter, Carlo Baldassari, Andrew~D. White, and Philippe Schwaller.
\newblock Augmenting large language models with chemistry tools.
\newblock \emph{Nature Machine Intelligence}, 6\penalty0 (5):\penalty0 525--535, 2024.

\bibitem[Bussemaker et~al.(2024)Bussemaker, Saves, Bartoli, Lefebvre, and Nagel]{bussemaker2024hidden}
Jasper Bussemaker, Paul Saves, Nathalie Bartoli, Thierry Lefebvre, and Bj{\"o}rn Nagel.
\newblock Surrogate-based optimization of system architectures subject to hidden constraints.
\newblock In \emph{AIAA Aviation Forum and Ascend 2024}, AIAA Paper 2024-4401, 2024.

\bibitem[Chakrabarty et~al.(2021)Chakrabarty, Bortoff, and Laughman]{chakrabarty2021failure}
Ankush Chakrabarty, Scott~A. Bortoff, and Christopher~R. Laughman.
\newblock Simulation failure robust {Bayesian} optimization for estimating black-box model parameters.
\newblock In \emph{IEEE International Conference on Systems, Man, and Cybernetics (SMC)}, pp.\  1533--1538, 2021.

\bibitem[Chan et~al.(2025)Chan, Chowdhury, Jaffe, Aung, Sherburn, Mays, Starace, Liu, Maksin, Patwardhan, Weng, and Madry]{chan2025mlebench}
Jun~Shern Chan, Neil Chowdhury, Oliver Jaffe, James Aung, Dane Sherburn, Evan Mays, Giulio Starace, Kevin Liu, Leon Maksin, Tejal Patwardhan, Lilian Weng, and Aleksander Madry.
\newblock {MLE}-bench: Evaluating machine learning agents on machine learning engineering.
\newblock In \emph{International Conference on Learning Representations}, 2025.

\bibitem[Chen et~al.(2024)Chen, White, Mooney, Payani, Su, and Sun]{chen2024treesearch}
Ziru Chen, Michael White, Ray Mooney, Ali Payani, Yu~Su, and Huan Sun.
\newblock When is tree search useful for {LLM} planning? {I}t depends on the discriminator.
\newblock In \emph{Proceedings of the 62nd Annual Meeting of the Association for Computational Linguistics (Volume 1: Long Papers)}, pp.\  13659--13678, 2024.

\bibitem[Chen et~al.(2025)Chen, Chen, Ning, Zhang, Wang, Yu, Li, Liao, Wei, Lu, Dey, Xue, Baker, Burns, Adu-Ampratwum, Huang, Ning, Gao, Su, and Sun]{chen2025scienceagentbench}
Ziru Chen, Shijie Chen, Yuting Ning, Qianheng Zhang, Boshi Wang, Botao Yu, Yifei Li, Zeyi Liao, Chen Wei, Zitong Lu, Vishal Dey, Mingyi Xue, Frazier~N. Baker, Benjamin Burns, Daniel Adu-Ampratwum, Xuhui Huang, Xia Ning, Song Gao, Yu~Su, and Huan Sun.
\newblock {ScienceAgentBench}: Toward rigorous assessment of language agents for data-driven scientific discovery.
\newblock In \emph{International Conference on Learning Representations}, 2025.

\bibitem[Fazliani et~al.(2026)Fazliani, Chawla, Guo, Shen, Ihme, and Udell]{fazliani2026shapebench}
Shaghayegh Fazliani, Krissh Chawla, Jack Guo, Yiren Shen, Matthias Ihme, and Madeleine Udell.
\newblock {ShapeBench}: A scalable benchmark and diagnostic suite for standardized evaluation in aerodynamic shape optimization.
\newblock \emph{arXiv preprint arXiv:2605.20763}, 2026.

\bibitem[Felten et~al.(2025)Felten, Apaza, Br{\"a}unlich, Diniz, Dong, Drake, Habibi, Hoffman, Keeler, Massoudi, VanGessel, and Fuge]{felten2025engibench}
Florian Felten, Gabriel Apaza, Gerhard Br{\"a}unlich, Cashen Diniz, Xuliang Dong, Arthur Drake, Milad Habibi, Nathaniel Hoffman, Matthew Keeler, Soheyl Massoudi, Francis VanGessel, and Mark Fuge.
\newblock {EngiBench}: A framework for data-driven engineering design research.
\newblock In \emph{Advances in Neural Information Processing Systems}, volume~38, 2025.

\bibitem[Ferreira et~al.(2026)Ferreira, Wobbe, Krishnakumar, Hutter, and Zela]{ferreira2026autoresearch}
Fabio Ferreira, Lucca Wobbe, Arjun Krishnakumar, Frank Hutter, and Arber Zela.
\newblock Can {LLMs} beat classical hyperparameter optimization algorithms? {A} study on autoresearch.
\newblock \emph{arXiv preprint arXiv:2603.24647}, 2026.

\bibitem[Forrester \& Keane(2009)Forrester and Keane]{forrester2009recent}
Alexander I.~J. Forrester and Andy~J. Keane.
\newblock Recent advances in surrogate-based optimization.
\newblock \emph{Progress in Aerospace Sciences}, 45\penalty0 (1--3):\penalty0 50--79, 2009.

\bibitem[Gelbart et~al.(2014)Gelbart, Snoek, and Adams]{gelbart2014unknown}
Michael~A. Gelbart, Jasper Snoek, and Ryan~P. Adams.
\newblock {Bayesian} optimization with unknown constraints.
\newblock In \emph{Proceedings of the Thirtieth Conference on Uncertainty in Artificial Intelligence}, pp.\  250--259, 2014.

\bibitem[Gottweis et~al.(2026)Gottweis, Weng, Daryin, Tu, Sirkovic, Myaskovsky, Glowaty, Weissenberger, Orlandi, Popovici, Palepu, Rong, Tanno, Saab, Zhang, Blum, Carroll, Kulkarni, Toma{\v{s}}ev, Zverinski, Rendulic, Vedadi, Hasler, Rimanic, Boia, Budiselic, Feinstein, Bellaiche, Sheffer, Freyberg, Ratcliff, Bertolli, Chou, Hassidim, Gokturk, Vahdat, Guan, Dhillon, Vaishnav, Lee, Costa, Penad{\'e}s, Peltz, Matias, Manyika, Hassabis, Xu, Kohli, Pawlosky, Karthikesalingam, and Natarajan]{gottweis2026coscientist}
Juraj Gottweis, Wei-Hung Weng, Alexander Daryin, Tao Tu, Petar Sirkovic, Artiom Myaskovsky, Grzegorz Glowaty, Felix Weissenberger, Alessio Orlandi, Dan Popovici, Anil Palepu, Keran Rong, Ryutaro Tanno, Khaled Saab, Fan Zhang, Jacob Blum, Andrew Carroll, Kavita Kulkarni, Nenad Toma{\v{s}}ev, Dina Zverinski, Ivor Rendulic, Elahe Vedadi, Florian Hasler, Luka Rimanic, Marina Boia, Ivan Budiselic, Ben Feinstein, Mathias Bellaiche, Tom Sheffer, Jan Freyberg, Jeremy Ratcliff, Ottavia Bertolli, Katherine Chou, Avinatan Hassidim, Burak Gokturk, Amin Vahdat, Yuan Guan, Vikram Dhillon, Eeshit~Dhaval Vaishnav, Byron Lee, Tiago R.~D. Costa, Jos{\'e}~R. Penad{\'e}s, Gary Peltz, Yossi Matias, James Manyika, Demis Hassabis, Yunhan Xu, Pushmeet Kohli, Annalisa Pawlosky, Alan Karthikesalingam, and Vivek Natarajan.
\newblock Accelerating scientific discovery with {Co-Scientist}.
\newblock \emph{Nature}, 655\penalty0 (8122):\penalty0 487--496, 2026.
\newblock \doi{10.1038/s41586-026-10644-y}.

\bibitem[Gupta et~al.(2025)Gupta, Hartford, and Liu]{gupta2025llmbo}
Rushil Gupta, Jason Hartford, and Bang Liu.
\newblock {LLMs} for {Bayesian} optimization in scientific domains: Are we there yet?
\newblock In \emph{Findings of the Association for Computational Linguistics: EMNLP 2025}, pp.\  15482--15510, 2025.

\bibitem[Hansen(2016)]{hansen2016tutorial}
Nikolaus Hansen.
\newblock The {CMA} evolution strategy: A tutorial.
\newblock \emph{arXiv preprint arXiv:1604.00772}, 2016.

\bibitem[Hansen \& Ostermeier(2001)Hansen and Ostermeier]{hansen2001cmaes}
Nikolaus Hansen and Andreas Ostermeier.
\newblock Completely derandomized self-adaptation in evolution strategies.
\newblock \emph{Evolutionary Computation}, 9\penalty0 (2):\penalty0 159--195, 2001.

\bibitem[He et~al.(2020)He, Sun, Song, and Wang]{he2020dual}
Youwei He, Jinju Sun, Peng Song, and Xuesong Wang.
\newblock Dual {Kriging} assisted efficient global optimization of expensive problems with evaluation failures.
\newblock \emph{Aerospace Science and Technology}, 105:\penalty0 106006, 2020.

\bibitem[He et~al.(2021)He, Sun, Song, and Wang]{he2021variable}
Youwei He, Jinju Sun, Peng Song, and Xuesong Wang.
\newblock Variable-fidelity expected improvement based efficient global optimization of expensive problems in presence of simulation failures and its parallelization.
\newblock \emph{Aerospace Science and Technology}, 111:\penalty0 106572, 2021.

\bibitem[Huang et~al.(2024{\natexlab{a}})Huang, Chen, Mishra, Zheng, Yu, Song, and Zhou]{huang2024selfcorrect}
Jie Huang, Xinyun Chen, Swaroop Mishra, Huaixiu~Steven Zheng, Adams~Wei Yu, Xinying Song, and Denny Zhou.
\newblock Large language models cannot self-correct reasoning yet.
\newblock In \emph{International Conference on Learning Representations}, 2024.

\bibitem[Huang et~al.(2024{\natexlab{b}})Huang, Vora, Liang, and Leskovec]{huang2024mlagentbench}
Qian Huang, Jian Vora, Percy Liang, and Jure Leskovec.
\newblock {MLAgentBench}: Evaluating language agents on machine learning experimentation.
\newblock In \emph{Proceedings of the 41st International Conference on Machine Learning}, volume 235 of \emph{Proceedings of Machine Learning Research}, pp.\  20271--20309, 2024.

\bibitem[Jin(2011)]{jin2011surrogate}
Yaochu Jin.
\newblock Surrogate-assisted evolutionary computation: Recent advances and future challenges.
\newblock \emph{Swarm and Evolutionary Computation}, 1\penalty0 (2):\penalty0 61--70, 2011.

\bibitem[Kamoi et~al.(2024)Kamoi, Zhang, Zhang, Han, and Zhang]{kamoi2024selfcorrection}
Ryo Kamoi, Yusen Zhang, Nan Zhang, Jiawei Han, and Rui Zhang.
\newblock When can {LLMs} actually correct their own mistakes? {A} critical survey of self-correction of {LLMs}.
\newblock \emph{Transactions of the Association for Computational Linguistics}, 12:\penalty0 1417--1440, 2024.

\bibitem[Kandasamy et~al.(2017)Kandasamy, Dasarathy, Schneider, and P{\'o}czos]{kandasamy2017multifidelity}
Kirthevasan Kandasamy, Gautam Dasarathy, Jeff Schneider, and Barnab{\'a}s P{\'o}czos.
\newblock Multi-fidelity {Bayesian} optimisation with continuous approximations.
\newblock In \emph{Proceedings of the 34th International Conference on Machine Learning}, volume~70 of \emph{Proceedings of Machine Learning Research}, pp.\  1799--1808. PMLR, 2017.

\bibitem[Kulfan(2008)]{kulfan2008universal}
Brenda~M. Kulfan.
\newblock Universal parametric geometry representation method.
\newblock \emph{Journal of Aircraft}, 45\penalty0 (1):\penalty0 142--158, 2008.

\bibitem[Le~Digabel \& Wild(2024)Le~Digabel and Wild]{ledigabel2024taxonomy}
S{\'e}bastien Le~Digabel and Stefan~M. Wild.
\newblock A taxonomy of constraints in black-box simulation-based optimization.
\newblock \emph{Optimization and Engineering}, 25\penalty0 (2):\penalty0 1125--1143, 2024.

\bibitem[Lee et~al.(2011)Lee, Gramacy, Linkletter, and Gray]{lee2011hidden}
Herbert K.~H. Lee, Robert~B. Gramacy, Crystal Linkletter, and Genetha~A. Gray.
\newblock Optimization subject to hidden constraints via statistical emulation.
\newblock \emph{Pacific Journal of Optimization}, 7\penalty0 (3):\penalty0 467--478, 2011.

\bibitem[Lindauer et~al.(2022)Lindauer, Eggensperger, Feurer, Biedenkapp, Deng, Benjamins, Ruhkopf, Sass, and Hutter]{lindauer2022smac3}
Marius Lindauer, Katharina Eggensperger, Matthias Feurer, Andr{\'e} Biedenkapp, Difan Deng, Carolin Benjamins, Tim Ruhkopf, Ren{\'e} Sass, and Frank Hutter.
\newblock {SMAC3}: A versatile {Bayesian} optimization package for hyperparameter optimization.
\newblock \emph{Journal of Machine Learning Research}, 23\penalty0 (54):\penalty0 1--9, 2022.

\bibitem[Liu et~al.(2024)Liu, Hu, Zhang, Guo, Ke, Liu, and Wang]{liu2024rafa}
Zhihan Liu, Hao Hu, Shenao Zhang, Hongyi Guo, Shuqi Ke, Boyi Liu, and Zhaoran Wang.
\newblock Reason for future, act for now: A principled architecture for autonomous {LLM} agents.
\newblock In \emph{Proceedings of the 41st International Conference on Machine Learning}, volume~235 of \emph{Proceedings of Machine Learning Research}, pp.\  31186--31261. PMLR, 2024.

\bibitem[Lu et~al.(2026)Lu, Lu, Lange, Yamada, Hu, Foerster, Ha, and Clune]{lu2026aiscientist}
Chris Lu, Cong Lu, Robert~Tjarko Lange, Yutaro Yamada, Shengran Hu, Jakob Foerster, David Ha, and Jeff Clune.
\newblock Towards end-to-end automation of {AI} research.
\newblock \emph{Nature}, 651\penalty0 (8107):\penalty0 914--919, 2026.
\newblock \doi{10.1038/s41586-026-10265-5}.

\bibitem[Lyu et~al.(2015)Lyu, Kenway, and Martins]{lyu2015crm}
Zhoujie Lyu, Gaetan K.~W. Kenway, and Joaquim R.~R.~A. Martins.
\newblock Aerodynamic shape optimization investigations of the {Common Research Model} wing benchmark.
\newblock \emph{AIAA Journal}, 53\penalty0 (4):\penalty0 968--985, 2015.

\bibitem[Mader et~al.(2020)Mader, Kenway, Yildirim, and Martins]{mader2020adflow}
Charles~A. Mader, Gaetan K.~W. Kenway, Anil Yildirim, and Joaquim R.~R.~A. Martins.
\newblock {ADflow}: An open-source computational fluid dynamics solver for aerodynamic and multidisciplinary optimization.
\newblock \emph{Journal of Aerospace Information Systems}, 17\penalty0 (9):\penalty0 508--527, 2020.

\bibitem[Martins(2022)]{martins2022aerodynamic}
Joaquim R.~R.~A. Martins.
\newblock Aerodynamic design optimization: Challenges and perspectives.
\newblock \emph{Computers \& Fluids}, 239:\penalty0 105391, 2022.

\bibitem[MDO Lab(n.d.)]{mdolabpyhyp}
MDO Lab.
\newblock Options.
\newblock \emph{pyHyp documentation}, n.d. Accessed November 13, 2026.
\newblock \url{https://mdolab-pyhyp.readthedocs-hosted.com/en/latest/options.html}.

\bibitem[Meindl et~al.(2026)Meindl, Tian, Cui, Thost, Hong, Chen, Matusik, and Konakovi{\'c} Lukovi{\'c}]{meindl2026semanticopt}
Jamison Meindl, Yunsheng Tian, Tony Cui, Veronika Thost, Zhang-Wei Hong, Jie Chen, Wojciech Matusik, and Mina Konakovi{\'c} Lukovi{\'c}.
\newblock {SemanticOpt}: Towards {LLM}-based semantic black-box optimization.
\newblock \emph{arXiv preprint arXiv:2510.25404v3}, 2026.

\bibitem[Nathani et~al.(2025)Nathani, Madaan, Roberts, Bashlykov, Menon, Moens, Plekhanov, Budhiraja, Magka, Vorotilov, Chaurasia, Hupkes, Cabral, Shavrina, Foerster, Bachrach, Wang, and Raileanu]{nathani2025mlgym}
Deepak Nathani, Lovish Madaan, Nicholas Roberts, Nikolay Bashlykov, Ajay Menon, Vincent Moens, Mikhail Plekhanov, Amar Budhiraja, Despoina Magka, Vladislav Vorotilov, Gaurav Chaurasia, Dieuwke Hupkes, Ricardo~Silveira Cabral, Tatiana Shavrina, Jakob Foerster, Yoram Bachrach, William~Yang Wang, and Roberta Raileanu.
\newblock {MLGym}: A new framework and benchmark for advancing {AI} research agents.
\newblock In \emph{Second Conference on Language Modeling}, 2025.

\bibitem[Novikov et~al.(2025)Novikov, V\~{u}, Eisenberger, Dupont, Huang, Wagner, Shirobokov, Kozlovskii, Ruiz, Mehrabian, Kumar, See, Chaudhuri, Holland, Davies, Nowozin, Kohli, and Balog]{novikov2025alphaevolve}
Alexander Novikov, Ng\^{a}n V\~{u}, Marvin Eisenberger, Emilien Dupont, Po-Sen Huang, Adam~Zsolt Wagner, Sergey Shirobokov, Borislav Kozlovskii, Francisco J.~R. Ruiz, Abbas Mehrabian, M.~Pawan Kumar, Abigail See, Swarat Chaudhuri, George Holland, Alex Davies, Sebastian Nowozin, Pushmeet Kohli, and Matej Balog.
\newblock {AlphaEvolve}: A coding agent for scientific and algorithmic discovery.
\newblock \emph{arXiv preprint arXiv:2506.13131}, 2025.

\bibitem[Plaban et~al.(2026)Plaban, Bachman, and Renganathan]{plaban2026derivativefree}
Punya Plaban, Peter Bachman, and Ashwin Renganathan.
\newblock Derivative-free optimization is competitive for aerodynamic design optimization in moderate dimensions.
\newblock \emph{Structural and Multidisciplinary Optimization}, 69\penalty0 (6):\penalty0 154, 2026.

\bibitem[Romera-Paredes et~al.(2024)Romera-Paredes, Barekatain, Novikov, Balog, Kumar, Dupont, Ruiz, Ellenberg, Wang, Fawzi, Kohli, and Fawzi]{romeraparedes2024funsearch}
Bernardino Romera-Paredes, Mohammadamin Barekatain, Alexander Novikov, Matej Balog, M.~Pawan Kumar, Emilien Dupont, Francisco J.~R. Ruiz, Jordan~S. Ellenberg, Pengming Wang, Omar Fawzi, Pushmeet Kohli, and Alhussein Fawzi.
\newblock Mathematical discoveries from program search with large language models.
\newblock \emph{Nature}, 625\penalty0 (7995):\penalty0 468--475, 2024.

\bibitem[Sacher et~al.(2018)Sacher, Duvigneau, Le~Ma{\^i}tre, Durand, Berrini, Hauville, and Astolfi]{sacher2018classification}
Matthieu Sacher, R{\'e}gis Duvigneau, Olivier Le~Ma{\^i}tre, Mathieu Durand, {\'E}lisa Berrini, Fr{\'e}d{\'e}ric Hauville, and Jacques-Andr{\'e} Astolfi.
\newblock A classification approach to efficient global optimization in presence of non-computable domains.
\newblock \emph{Structural and Multidisciplinary Optimization}, 58\penalty0 (4):\penalty0 1537--1557, 2018.

\bibitem[Schmidgall et~al.(2025)Schmidgall, Su, Wang, Sun, Wu, Yu, Liu, Moor, Liu, and Barsoum]{schmidgall2025agentlab}
Samuel Schmidgall, Yusheng Su, Ze Wang, Ximeng Sun, Jialian Wu, Xiaodong Yu, Jiang Liu, Michael Moor, Zicheng Liu, and Emad Barsoum.
\newblock Agent {Laboratory}: Using {LLM} agents as research assistants.
\newblock In \emph{Findings of the Association for Computational Linguistics: EMNLP 2025}, pp.\  5977--6043, 2025.

\bibitem[Shinn et~al.(2023)Shinn, Cassano, Gopinath, Narasimhan, and Yao]{shinn2023reflexion}
Noah Shinn, Federico Cassano, Ashwin Gopinath, Karthik Narasimhan, and Shunyu Yao.
\newblock {Reflexion}: Language agents with verbal reinforcement learning.
\newblock In \emph{Advances in Neural Information Processing Systems}, volume~36, pp.\  8634--8652. Curran Associates, Inc., 2023.

\bibitem[Smith \& Winkler(2006)Smith and Winkler]{smith2006curse}
James~E. Smith and Robert~L. Winkler.
\newblock The optimizer's curse: Skepticism and postdecision surprise in decision analysis.
\newblock \emph{Management Science}, 52\penalty0 (3):\penalty0 311--322, 2006.

\bibitem[Somasekharan et~al.(2026)Somasekharan, Yue, Cao, Li, Emami, Bhargav, Acharya, Xie, and Pan]{somasekharan2026cfdllmbench}
Nithin Somasekharan, Ling Yue, Yadi Cao, Weichao Li, Patrick Emami, Pochinapeddi~Sai Bhargav, Anurag Acharya, Xingyu Xie, and Shaowu Pan.
\newblock {CFDLLMBench}: A benchmark suite for evaluating large language models in computational fluid dynamics.
\newblock \emph{Journal of Data-centric Machine Learning Research}, 3\penalty0 (13):\penalty0 1--40, 2026.

\bibitem[Stenger \& Abel(2022)Stenger and Abel]{stenger2022benchmark}
David Stenger and Dirk Abel.
\newblock Benchmark of {Bayesian} optimization and metaheuristics for control engineering tuning problems with crash constraints.
\newblock \emph{arXiv preprint arXiv:2211.02571}, 2022.

\bibitem[Storn \& Price(1997)Storn and Price]{storn1997differential}
Rainer Storn and Kenneth Price.
\newblock Differential evolution -- a simple and efficient heuristic for global optimization over continuous spaces.
\newblock \emph{Journal of Global Optimization}, 11\penalty0 (4):\penalty0 341--359, 1997.

\bibitem[Swanson et~al.(2025)Swanson, Wu, Bulaong, Pak, and Zou]{swanson2025virtuallab}
Kyle Swanson, Wesley Wu, Nash~L. Bulaong, John~E. Pak, and James Zou.
\newblock The {Virtual Lab} of {AI} agents designs new {SARS-CoV-2} nanobodies.
\newblock \emph{Nature}, 646\penalty0 (8085):\penalty0 716--723, 2025.
\newblock \doi{10.1038/s41586-025-09442-9}.

\bibitem[Takeno et~al.(2020)Takeno, Fukuoka, Tsukada, Koyama, Shiga, Takeuchi, and Karasuyama]{takeno2020multifidelity}
Shion Takeno, Hitoshi Fukuoka, Yuhki Tsukada, Toshiyuki Koyama, Motoki Shiga, Ichiro Takeuchi, and Masayuki Karasuyama.
\newblock Multi-fidelity {Bayesian} optimization with max-value entropy search and its parallelization.
\newblock In \emph{Proceedings of the 37th International Conference on Machine Learning}, volume~119 of \emph{Proceedings of Machine Learning Research}, pp.\  9334--9345. PMLR, 2020.

\bibitem[Tfaily et~al.(2024)Tfaily, Diouane, Bartoli, and Kokkolaras]{tfaily2024hidden}
Ali Tfaily, Youssef Diouane, Nathalie Bartoli, and Michael Kokkolaras.
\newblock {Bayesian} optimization with hidden constraints for aircraft design.
\newblock \emph{Structural and Multidisciplinary Optimization}, 67\penalty0 (7):\penalty0 123, 2024.
\newblock \doi{10.1007/s00158-024-03833-8}.

\bibitem[Tran et~al.(2019)Tran, Sun, Furlan, Pagalthivarthi, Visintainer, and Wang]{tran2019pbo}
Anh Tran, Jing Sun, John~M. Furlan, Krishnan~V. Pagalthivarthi, Robert~J. Visintainer, and Yan Wang.
\newblock {pBO-2GP-3B}: A batch parallel known/unknown constrained {Bayesian} optimization with feasibility classification and its applications in computational fluid dynamics.
\newblock \emph{Computer Methods in Applied Mechanics and Engineering}, 347:\penalty0 827--852, 2019.

\bibitem[Vassberg et~al.(2008)Vassberg, DeHaan, Rivers, and Wahls]{vassberg2008crm}
John~C. Vassberg, Mark DeHaan, Melissa Rivers, and Richard~A. Wahls.
\newblock Development of a {Common Research Model} for applied {CFD} validation studies.
\newblock In \emph{26th AIAA Applied Aerodynamics Conference}, AIAA Paper 2008-6919, 2008.

\bibitem[Yamada et~al.(2025)Yamada, Lange, Lu, Hu, Lu, Foerster, Clune, and Ha]{yamada2025aiscientistv2}
Yutaro Yamada, Robert~Tjarko Lange, Cong Lu, Shengran Hu, Chris Lu, Jakob Foerster, Jeff Clune, and David Ha.
\newblock The {AI} {Scientist}-v2: Workshop-level automated scientific discovery via agentic tree search.
\newblock \emph{arXiv preprint arXiv:2504.08066}, 2025.

\bibitem[Yang et~al.(2026{\natexlab{a}})Yang, Gholami, Guerbuez, Rashed, and Thuerey]{yang2026aerotransformer}
Yunjia Yang, Babak Gholami, Caglar Guerbuez, Mohammad Rashed, and Nils Thuerey.
\newblock Toward a foundation-model paradigm for aerodynamic prediction in three-dimensional design.
\newblock \emph{AIAA Journal}, advance online publication, pp. 1--21, 2026{\natexlab{a}}.
\newblock \doi{10.2514/1.J066861}.

\bibitem[Yang et~al.(2026{\natexlab{b}})Yang, Tang, Liu, Thuerey, Zhang, and Chen]{yang2026superwing}
Yunjia Yang, Weishao Tang, Mengxin Liu, Nils Thuerey, Yufei Zhang, and Haixin Chen.
\newblock {SuperWing}: A comprehensive transonic wing dataset for data-driven aerodynamic design.
\newblock \emph{Scientific Data}, 13:\penalty0 1113, 2026{\natexlab{b}}.
\newblock \doi{10.1038/s41597-026-07769-0}.

\bibitem[Yue et~al.(2026)Yue, Somasekharan, Zhang, Cao, Chen, Di, and Pan]{yue2026foamagent}
Ling Yue, Nithin Somasekharan, Tingwen Zhang, Yadi Cao, Zhangze Chen, Shimin Di, and Shaowu Pan.
\newblock {Foam-Agent}: A large language model-based multi-agent framework for automating computational fluid dynamics workflows.
\newblock \emph{Computer Methods in Applied Mechanics and Engineering}, 461:\penalty0 119271, 2026.

\end{thebibliography}
